\documentclass[sigconf]{acmart}
\usepackage{xspace}
\usepackage{graphicx} % DO NOT CHANGE THIS
\usepackage{algorithmic}
\usepackage{mathrsfs}
\usepackage{bm}
\usepackage{amsmath}

\usepackage{amssymb}
\usepackage{booktabs}
\usepackage{pifont}
\usepackage{multirow}
\usepackage{hyperref}
\usepackage{cancel}
\usepackage{dsfont}
\usepackage{subfig}
\usepackage{bbm}
\usepackage{marvosym}
\usepackage{balance}
\usepackage[ruled,vlined]{algorithm2e}
\newcommand{\para}[1]{{\vspace{2pt} \bf \noindent #1 \hspace{1pt}}}

\newcommand{\etc}{\textit{etc}.}
\newcommand{\ie}{\textit{i}.\textit{e}.}
\newcommand{\eg}{\textit{e}.\textit{g}.}
\newcommand{\ourmethod}{\textsc{CHAIN}\xspace}
\newcommand{\cmark}{\ding{51}}
\newcommand{\xmark}{\ding{55}}
\definecolor{linkcolor}{RGB}{218, 50, 138}
\AtBeginDocument{%
  }

\copyrightyear{2023}
\acmYear{2023}
\setcopyright{acmlicensed}\acmConference[MM '23]{Proceedings of the 31st
ACM International Conference on Multimedia}{October 29-November 3,
2023}{Ottawa, ON, Canada}
\acmBooktitle{Proceedings of the 31st ACM International Conference on
Multimedia (MM '23), October 29-November 3, 2023, Ottawa, ON, Canada}
\acmPrice{15.00}
\acmDOI{10.1145/3581783.3613440}
\acmISBN{979-8-4007-0108-5/23/10}

\acmSubmissionID{217}

\begin{document}

\title{CHAIN: Exploring Global-Local Spatio-Temporal Information\\ for Improved Self-Supervised Video Hashing}



\author{Rukai Wei}
\affiliation{%
  \institution{Huazhong University of Science and Technology}
  \streetaddress{}
  \city{Wuhan}
  \country{China}}
\email{weirukai@hust.edu.cn}

\author{Yu Liu}
\authornote{Corresponding author.}
\affiliation{%
  \institution{Huazhong University of Science and Technology}
  \streetaddress{}
  \city{Wuhan}
  \country{China}}
\email{liu_yu@hust.edu.cn}

\author{Jingkuan Song}
\affiliation{%
  \institution{University of Electronic Science and Technology of China}
  \streetaddress{}
  \city{Chengdu}
  \country{China}}
\email{jingkuan.song@gmail.com}

\author{Heng Cui}
\affiliation{%
  \institution{Huazhong University of Science and Technology}
  \city{Wuhan}
  \country{China}
}
\email{hengcui@hust.edu.cn}

\author{Yanzhao Xie}
\affiliation{%
 \institution{Huazhong University of Science and Technology}
    \city{Wuhan}
  \country{China}
  }
\email{yzxie@hust.edu.cn}

\author{Ke Zhou}
\affiliation{%
  \institution{Huazhong University of Science and Technology}
  \city{Wuhan}
  \country{China}}
\email{zhke@hust.edu.cn}

\renewcommand{\shortauthors}{Rukai Wei et al.}

\begin{abstract}

Compressing videos into binary codes can improve retrieval speed and reduce storage overhead. However, learning accurate hash codes for video retrieval can be challenging due to high local redundancy and complex global dependencies between video frames, especially in the absence of labels. Existing self-supervised video hashing methods have been effective in designing expressive temporal encoders, but have not fully utilized the temporal dynamics and spatial appearance of videos due to less challenging and unreliable learning tasks. To address these challenges, we begin by utilizing the contrastive learning task to capture global spatio-temporal information of videos for hashing. With the aid of our designed augmentation strategies, which focus on spatial and temporal variations to create positive pairs, the learning framework can generate hash codes that are invariant to motion, scale, and viewpoint. Furthermore, we incorporate two collaborative learning tasks, \ie, frame order verification and scene change regularization, to capture local spatio-temporal details within video frames, thereby enhancing the perception of temporal structure and the modeling of spatio-temporal relationships. Our proposed  \underline{\textbf{C}}ontrastive \underline{\textbf{H}}ashing with Global-Local Sp\underline{\textbf{a}}tio-temporal \underline{\textbf{I}}nformatio\underline{\textbf{n}} (\ourmethod) outperforms state-of-the-art self-supervised video hashing methods on four video benchmark datasets. Our codes will be released.
 
\end{abstract}




\maketitle

\begin{figure}[tb!]
    \centering
    \includegraphics[width=\linewidth]{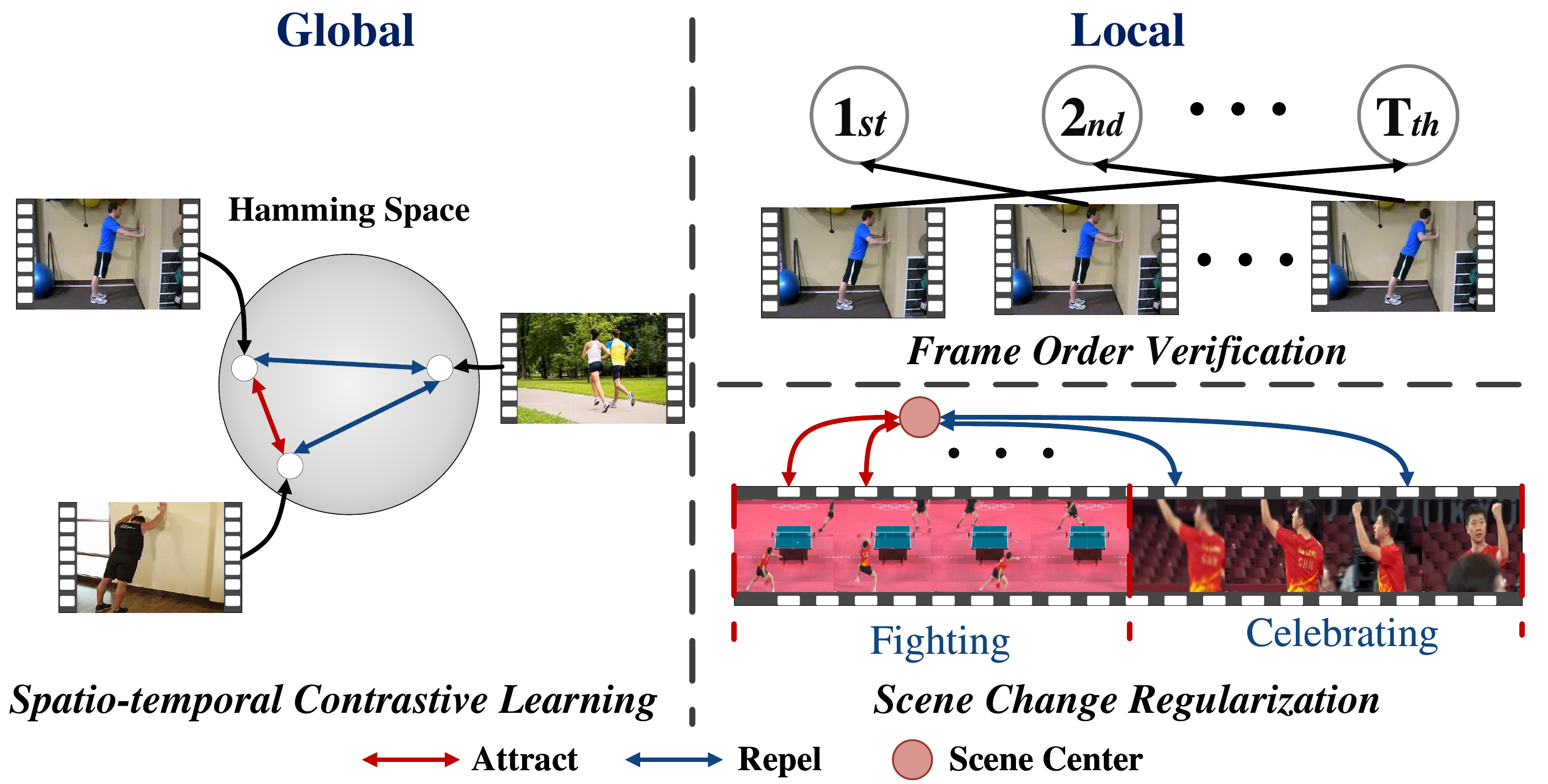}
    \caption{Illustration of the proposed three complementary learning tasks. Spatio-temporal contrastive learning concentrates on the global inter-video relationships, while frame order verification and scene change regularization focus on the local intra-video spatio-temporal details.}
    \label{fig:tasks}
    \vspace{-1em}
\end{figure}

\section{Introduction}

Videos, as common multimedia data, have recently received significant attention from researchers. As the demand for large-scale video retrieval increases in various multimedia applications, efficient retrieval and low storage costs have become urgent needs. To cope with these requirements, hashing-based methods~\cite{CSQ2020CVPR,DHN2016AAAI,WSIH2022MM,ASSPH2022MM,FSVCH2022MM} have emerged as the dominant solution for scalable video retrieval. They compress high-dimensional features into binary codes and preserve the similarities between samples in Hamming space, offering an efficient way to retrieve videos at scale. Since manual annotation for large-scale video data is expensive and biased, self-supervised video hashing methods~\cite{NPH2019ICCV,SSVH2018TIP,BTH2021CVPR,MCMSH2022MM} are in the spotlight.

Compared to 2D image retrieval, retrieving videos by self-supervised hashing is a challenging task due to the high local redundancy and complex global dependencies between video frames~\cite{dong2022partially,CVPR2023Video}. To address this issue, existing self-supervised video hashing methods~\cite{SSVH2018TIP,NPH2019ICCV,BTH2021CVPR,MCMSH2022MM} primarily focus on designing expressive learning models that capture temporal dependence. For instance, SSVH~\cite{SSVH2018TIP}, BTH~\cite{BTH2021CVPR}, and MCMSH~\cite{MCMSH2022MM} employ the LSTM model~\cite{LSTM1997NECO}, the BERT model~\cite{BERT2019NAACL}, and the MC-MLP model based on MLP-Mixer~\cite{MLP-Mixer2021NIPS}, respectively. Nevertheless, their learning tasks, such as \textit{structure or neighborhood preserving}, focus on static spatial relationships, while neglecting valuable temporal cues, such as sequential order and changes over time, that are crucial for motion recognition. Therefore, it is imperative to develop more challenging learning tasks that can effectively capture both temporal and spatial information in videos, with the aim of enhancing retrieval performance.

To this end, we first design a contrastive learning task for self-supervised video hashing. To create high-quality positive pairs, we use spatial augmentations based on standard image contrastive learning methods~\cite{SimCLR2020ICML,MoCo2020CVPR,BYOL2020NIPS}, as well as temporal augmentations through segment sampling~\cite{VCLR2021ICCVW}. This spatio-temporal contrastive learning framework allows us to perceive both the spatial and temporal context of videos from unlabeled samples, and use the context as a supervisory signal to learn motion, scale, and viewpoint invariant hash codes. (See $\S$\ref{sec:contrastive_learning}). 

%our proposed \textbf{S}patio\textbf{t}emporal \textbf{C}ontrastive \textbf{H}ashing (\ourmethod) introduces three collaborative learning tasks, including spatio-temporal contrastive learning, frame order verification, and scene change regularization. First, \ourmethod employs a contrastive learning framework for self-supervised video hash learning, where we define not only spatial augmentations following typical image contrastive learning methods~\cite{SimCLR2020ICML,MoCo2020CVPR,BYOL2020NIPS} but also segment sampling-based temporal augmentations~\cite{VCLR2021ICCVW} to form high-quality positive pairs. The spatio-temporal contrastive learning enables us to use both spatial and temporal context as a supervisory signal (from unlabelled videos) to learn motion, scale, and viewpoint invariant hash codes (See $\S$\ref{sec:contrastive_learning}).

 % different human actions or scenes are the key information to describe a video

Note that the conventional contrastive learning task is primarily concerned with the global spatio-temporal representations of videos~\cite{VCLR2021ICCVW,TimeMatter2022ICML,SeCo2021AAAI,dong2022dual}. As shown in the left segment of Figure~\ref{fig:tasks}, global information focuses on the correlation between video instances. However, local spatio-temporal details in the video frames are also critical to understanding the content of videos since they often provide key cues for distinguishing between different video instances. As shown in the right segment of Figure~\ref{fig:tasks}, the evolution of actions as well as variations in viewpoint within a video also provide crucial information for describing the video. To address this issue through learning tasks, we propose the frame order verification task and the scene change regularization task. For the former, we follow classic methods~\cite{VideoOrder2017ICCV,VideoOrder2016ECCV} for predicting the absolute temporal position of frames in a sampled clip one by one (See $\S$\ref{sec:frame_order}), achieving full exploitation of the inherent sequential structure of a video. For the latter, we use intra-video prototypical contrastive learning~\cite{PCL2021ICLR}, which enables the model to distinguish between various scenes within a video (See $\S$\ref{sec:scene_change}). This task can capture the necessary variation, which is crucial for robust spatio-temporal modeling. It can compensate for the weak perceptibility of transformer-based contrastive learning to frequent scene changes~\cite{TimeMatter2022ICML}.

 %while it can be challenging to capture meaningful local temporal details, which often provide essential cues for discriminating between different video instances, such as human actions. To address this limitation, frame order verification and scene change regularization tasks are proposed to compensate it from local temporal and spatial perspectives, respectively. On the one hand, we follow~\cite{VideoOrder2017ICCV,VideoOrder2016ECCV} and propose a frame order verification task, which infers the absolute temporal position of frames in a sampled clip one-by-one (See $\S$\ref{sec:frame_order}), to exploit the inherent sequential structure of a video. On the other hand, contrastive learning with transformers tends to overlook frequent scene changes in videos, despite the fact that they can bring large spatio-temporal shifts and reduce the discriminatory power of the model. Therefore, we introduce a scene change regularization task, which performs intra-video prototypical contrastive learning~\cite{PCL2021ICLR} with frames from all augmented clips, to capture the variation for robust spatio-temporal modeling (See $\S$\ref{sec:scene_change}). Frame order verification captures the temporal connections between frames, while scene change regulation focuses on spatial relationships between frames. Contrastive learning emphasizes the global relationships between videos. By combining these three complementary tasks, our method can achieve excellent spatio-temporal modeling with multi-granularity information.
Finally, we integrate the above learning tasks into the contrastive learning framework and propose a novel method, namely, \underline{\textbf{C}}ontrastive \underline{\textbf{H}}ashing with Global-Local Sp\underline{\textbf{a}}tio-Temporal \underline{\textbf{I}}nformatio\underline{\textbf{n}} (\ourmethod). The spatio-temporal contrastive hashing task emphasizes the global relationships between video instances. Meanwhile, frame order verification captures the temporal connections between video frames, while scene change regulation focuses on spatial relationships between video frames. To accomplish these tasks effectively, the model is required to explore the multi-granularity context for robust spatio-temporal modeling. We conduct extensive experiments on four video benchmark datasets, including UCF-101~\cite{UCF-1012012arxiv}, HMDB-51~\cite{HMDB2011ICCV}, FCVID~\cite{FCVID2018TPAMI}, and ActivityNet~\cite{ActivityNet2015CVPR}. Experimental results demonstrate that our proposed \ourmethod outperforms state-of-the-art self-supervised video hashing methods by a large margin. Our main contributions can be outlined as follows:
\begin{itemize}
    % \item We propose a novel self-supervised video hashing paradigm based on the contrastive learning framework, where we define both spatial and temporal augmentations for positive pair sampling, enhancing the model to learn motion, scale, and viewpoint invariant hash codes for videos.
    % \item We introduce frame order verification and scene change regularization as complementary learning tasks for a superior understanding of local temporal dynamics and spatial appearance. Frame order verification helps exploit the inherent sequential structure in videos by predicting the absolute temporal positions frame-by-frame. Scene change regularization captures the spatio-temporal variations for robust spatio-temporal modeling.
\item For global spatio-temporal relationships between video instances, we propose a novel spatio-temporal contrastive learning framework, where we define both spatial and temporal augmentations to create positive pairs, enhancing the model to learn motion, scale, and viewpoint invariant hash codes for videos.
\item For local temporal connections between video frames, we introduce a frame order verification task to predict the absolute temporal positions of video frames, achieving full exploitation of the inherent sequential structure.
\item For local spatial relationships between video frames, we incorporate a scene change regularization task to distinguish various scenes within a video, allowing the model to capture the spatio-temporal variations for robust spatio-temporal modeling.
    % \item We introduce a frame order verification task to predict the absolute temporal positions, achieving full exploitation of the inherent sequential structure.
    % \item We incorporate a scene change regularization task to capture the spatio-temporal variations for robust spatio-temporal modeling.
    % \item Extensive experimental results on four benchmark datasets demonstrate the superiority of our proposed \ourmethod.

\end{itemize}

\begin{figure*}[thb]
    \centering
    \includegraphics[width=0.95\linewidth]{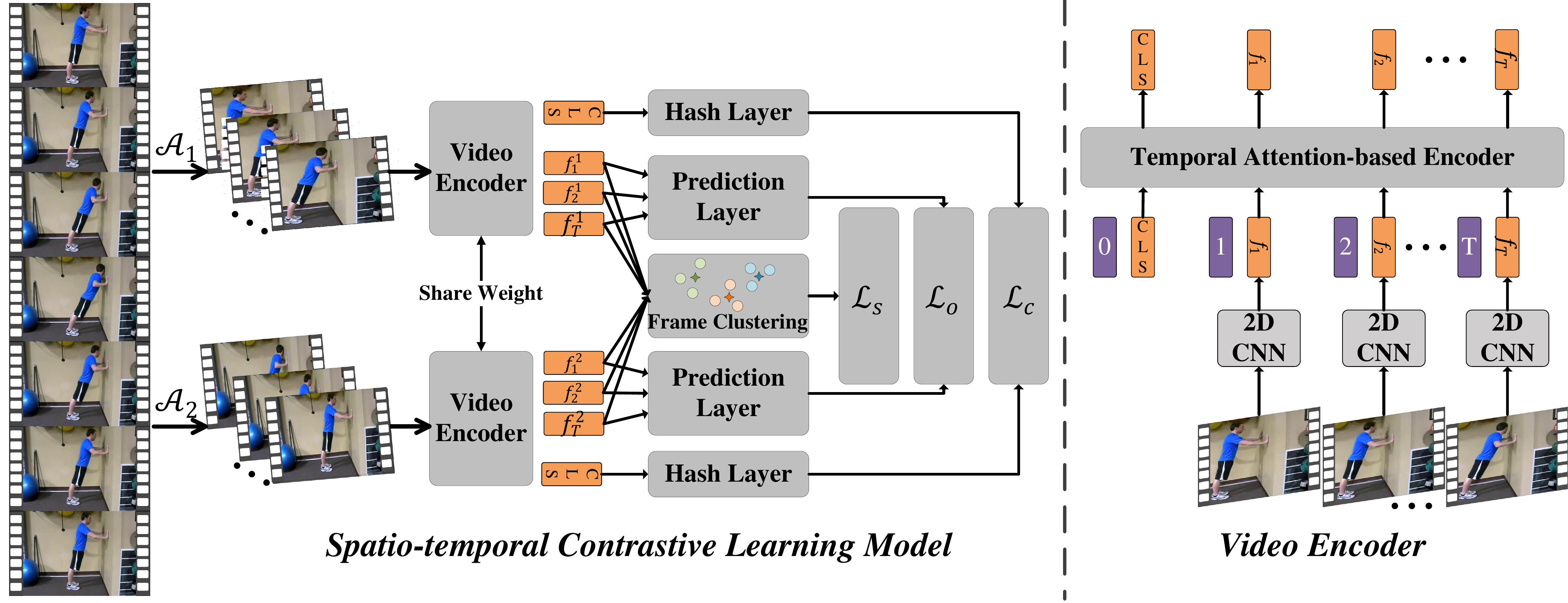}
    \caption{The overall framework of the proposed \ourmethod. First, the proposed spatio-temporal augmentation strategies ($\mathcal{A}_1$ and $\mathcal{A}_2$) are applied to the input video, resulting in two augmented clips. Next, frame-level features are extracted using a 2D CNN, while long-range dependencies are modeled using a temporal attention-based encoder. Finally, a spatio-temporal contrastive learning framework is employed to learn high-quality hash codes. To better utilize spatio-temporal cues, two collaborative tasks are introduced in addition to the contrastive learning task. $\mathcal{L}_{c}$, $\mathcal{L}_{o}$, and $\mathcal{L}_{s}$ symbolize the objective functions of three learning tasks, respectively.}
    \label{fig:framework}
    \vspace{-1.0em}
\end{figure*}

\section{Related Work}

In this paper, we focus on extending contrastive representation learning to self-supervised video hashing tasks. Therefore, our work is mainly related to the following two research areas:

\para{Self-supervised video hashing.} Previous research~\cite{ITQ2011CVPR,SpectralHashing2008NIPS,MPH2011MM} considers frames within a video in isolation and learns hash codes without exploiting the temporal cues that naturally exist in videos, which fails to achieve satisfying retrieval performance. Subsequently, to model temporal relations between video frames, most recent works~\cite{SSTH2016MM,SSTH2016MM} adopt RNN~\cite{RNN2014arxiv} and its variants LSTM~\cite{LSTM1997NECO} networks which are inherently suitable for capturing the sequence data, \eg, SSVH~\cite{SSVH2018TIP}, JATE~\cite{JTAE2017CIKM}, and NPH~\cite{NPH2019ICCV}. Most recently, BTH~\cite{BTH2021CVPR} uses a bidirectional transformer to mine more reliable temporal dependencies. MCMSH~\cite{MCMSH2022MM} proposes MC-MLP to separately model the three granularities of dependence. In addition, most methods define multi-tasks to learn hash codes in label-free scenarios. For example, SSVH, BTH, and MCMSH use similarity-preserving algorithms to reconstruct pairwise similarities between high-dimensional features in Hamming space. Meanwhile, they perform auxiliary tasks (\ie, masked frames prediction, cluster alignment, inter/intra-class variation \etc) for a better understanding of the temporal structure. 

\para{Contrastive self-supervised learning.} Contrastive learning is a popular technique for learning view-invariant representations by attracting positive samples and repelling negative ones. This technique has shown promising performance in image representation learning, with frameworks such as MoCo~\cite{MoCo2020CVPR}, SimCLR~\cite{SimCLR2020ICML}, BYOL~\cite{BYOL2020NIPS}, and SimSiam~\cite{SimSiam2021CVPR} being widely adopted in many downstream tasks. Recently, contrastive learning has been extended to video understanding. For example, VideoMoco~\cite{VideoMoCo2021CVPR} extends the image-based MoCo framework to video pre-training by modeling the temporal decay of the keys. In addition, SVT~\cite{SVT2022CVPR}, CVRL~\cite{CVRL2021CVPR}, and VCLR~\cite{VCLR2021ICCVW} are established on the contrastive learning framework by defining various positive and negative sampling strategies.

\para{\textit{Improvements over existing jobs.}} Our proposed \ourmethod leverages state-of-the-art video hashing methods for its model design, while utilizing spatio-temporal cues to learn video hash codes within a contrastive learning framework, resulting in an enhanced understanding of global video content. In addition, our frame order verification and scene change regulation tasks improve the model's ability to capture local spatio-temporal details. With the ability to fully exploit multi-granularity spatio-temporal information, our method achieves superior video hash learning.

%benefits from state-of-the-art video hashing methods in terms of the model design, while \ourmethod learns video hash codes with the spatio-temporal cues under the contrastive learning framework, enabling a superior understanding of video semantics. Moreover, the frame order verification and the scene change regulation tasks enhance the model's ability to capture local temporal details. Our method can fully exploit multi-grained spatio-temporal information for superior video hash learning.

\section{Method}

\subsection{Problem Define and Overview}
Given a training set $\mathcal{X}=\left\{x_{i}\right\}_{i=1}^{N}$ with $N$ videos, we represent a video by its clip consisting of $T$ frames, \ie, $x_{i} \in \mathcal{R}^{T\times C\times H \times W}$, where $C$ is the channel dimension and $H\times W$ is the spatial size. \ourmethod aims to learn a nonlinear hash function that encodes each input $x_{i}$ into a compact $K$-bit binary hash code $b_{i}$ in the Hamming space, denoted as $x_{i}\rightarrow \left\{-1,1\right\}^{K}$. Note that \ourmethod does not require any supervision information from hand-crafted labels. The framework of \ourmethod is shown in Figure~\ref{fig:framework}.

\para{Hash feature encoder.} To process a sampled video clip $x=\left\{v_{j}\right\}_{j=1}^{T}$, we adopt the video encoder presented in Figure~\ref{fig:framework}, which consists of a 2D CNN network and a attention-based transformer encoder. We first apply a 2D spatial encoder (\eg, VGG~\cite{vgg2014ICLR} or ResNet50~\cite{ResNet2016CVPR} for fair comparisons with SOTA methods~\cite{BTH2021CVPR,SSVH2018TIP,MCMSH2022MM}) to extract the feature of each frame. Then, following BERT~\cite{BERT2019NAACL} and ViT~\cite{ViT2021ICLR}, we add a special classification token (\ie, $[CLS]$) in the front of the frame-level feature sequence. All these tokens will be added with position embeddings and then fed to the temporal attention-based encoder~\cite{Transformer2017NIPS}, where the attention mechanism will help to model the long-range dependencies. After that, we can obtain temporal-aware representations $\left\{f_{j}\right\}_{j=1}^{T}\cup\left\{z\right\}$, where $z$ denotes the representation of $x$ captured by the $[CLS]$ token, and $f_{j}$ is the output embedding of frame $v_{j}$. For the global representation $z$, \ourmethod will further project it into a real-valued hash code $h\in \left(-1,1\right)^{K}$ with the hash layer, which contains two fully-connected layers. Finally, we discretize it into a binary code $b\in \left\{-1,1\right\}^{K}$ by $sgn$. 

% As for the frame-level representation $\left\{f_{j}\right\}_{j=1}^{T}$, they will be sent to the order prediction layer to infer the absolute position in the video clip frame-by-frame. Additionally, all these representations will also be clustered for scene change regularization.

\para{Hash learning process.} \ourmethod learns high-quality hash codes under the contrastive learning framework. Given $x_{i}$, we first perform both spatial and temporal augmentations (See $\S$~\ref{sec:contrastive_learning}) to generate two views, \ie, $x_{i}^{1}=\left\{v_{j}^{1} \mid j=1, ..., T\right\}$ and $x_{i}^{2}=\left\{v_{j}^{2}\mid j=1, ..., T\right\}$. Note that these video clips may contain different video frames due to the randomness of temporal augmentation. Using the video encoder aforementioned, we can obtain the hash codes $z_{i}^{1}$ and $z_{i}^{2}$ as well as frame-level representations $f_{i,j}^{1}$ and $f_{i,j}^{2}$, where $j=1, ..., T$. For the hash codes learning through global information, we adopt a standard contrastive learning objective to maximize the consistency between different views from the same video (See $\S$~\ref{sec:contrastive_learning}). Regarding the frame-level representations, on one hand, we send them to the order prediction layer to infer their absolute positions in the video clip frame by frame (See $\S$~\ref{sec:frame_order}). On the other hand, all $f_{i,j}^{1}$ and $f_{i,j}^{2}$ will be clustered into several classes with the Affinity-Propagation algorithm~\cite{APClustering} for scene change regularization (See $\S$~\ref{sec:scene_change}). By introducing three complementary learning tasks, \ourmethod allows for a better comprehension of spatial and temporal cues, resulting in more robust hash codes for videos.

\subsection{Spatio-temporal Contrastive Learning}
% Contrastive learning, which draws positive pairs and repels negative pairs, endows the model with exceptional discriminative ability and has proven to be effective in a number of tasks.
\label{sec:contrastive_learning}
 In contrast to SOTA methods~\cite{MCMSH2022MM,NPH2019ICCV,SSVH2018TIP} that define a similarity preserving task to learn hash codes without the perception of temporal dynamics, we design a spatio-temporal contrastive learning framework for superior video understanding. The instance-wise contrastive learning frameworks~\cite{SimCLR2020ICML,MoCo2020CVPR,BYOL2020NIPS} aim to maximize the agreement between views that are augmented from the same instance. These frameworks rely on augmentation strategies to provide variance while preserving consistency, especially when applied to video learning. To this end, \ourmethod applies both spatial and temporal augmentation to form high-quality positive pairs for videos.

\para{Segment sampling based temporal augmentation.} Temporal augmentation encourages the introduction of variance along the temporal dimension without altering video content. However, videos are more complicated than images with 2D spatial features due to the additional temporal dimension, which requires more effective augmentation strategies. Videos have two prominent characteristics: 1) Near frames are often redundant. 2) Videos depend on global context to tell the evolution of events. As a result, we follow the classic method~\cite{VCLR2021ICCVW} to perform segment-based frame sampling to learn from the global context. Specifically, for a video containing $L$ frames, we divide it into $T$ segments with equal duration. We randomly select one frame within each segment, and this allows us to obtain $T$ frames in total that represent the video. By performing the random segment-level sampling twice, we can obtain two clips, \ie, $\left\{v_{1}^{1}, v_{2}^{1}, ..., v_{T}^{1}\right\}$ and $\left\{v_{1}^{2}, v_{2}^{2}, ..., v_{T}^{2}\right\}$. They form a positive pair. Segment-level sampling has the potential to reduce local redundancy and provide a global view of the video for a better understanding of the entire event. Moreover, it allows the model to sample more scenes in the video, which is the foundation for implementing scene change regulation.

\para{Temporally consistent spatial augmentation.} Intuitively, we can apply existing image-based spatial augmentation methods, \eg, RandomResizedCrop, RandomGrayScale, and RandomColorJitter, to videos frame-by-frame. However, since the randomness in cropping and blurring will break the continuity between frames, we follow the classic method~\cite{CVRL2021CVPR} to fix the randomness across frames to make the spatial augmentation consistent along the temporal dimension. Note that spatial augmentation should differ across clips as our goal is to introduce variance between views. For example, we set the same random seed for all frames within a clip, while the seed for the other clip will be different.

% It is important to note that, while we ensure consistent spatial augmentation within each clip, the spatial augmentation applied to different clips can still differ. This is intentional, as our objective is to introduce variance between different views.

\para{Contrastive learning objective.} We define the clips sampled from the same video as \textit{positive pairs}, while the clips from other videos within a mini-batch are \textit{negative pairs}. By attracting positive pairs and repelling negative pairs, contrastive learning enables the model to generate hash codes without annotations. Following classic methods~\cite{SimCLR2020ICML,CIBHash2021IJCAI,DATE2021MM}, we can define the contrastive loss as follows:
\begin{equation}
    \begin{aligned}
    \hat{\ell}_{i}^{1}=-\log \frac{exp\left(sim(b_{i}^{1},b_{i}^{2})/\tau\right)}{exp\left(sim(b_{i}^{1},b_{i}^{2})/\tau\right)+\sum_{n,j\neq i} exp\left(sim(b_{i}^{1},b_{j}^{n})/\tau\right)},
    \end{aligned}
\end{equation}
where $\ell_{i}^{n}$ is the loss of the $n$-th ($n =1\text{ or }2$) view of $b_{i}$, $sim(\cdot)$ can be defined as the cosine similarity, \ie, $sim(b_{i},b_{j})=\frac{b_{i}^{T}b_{j}}{\left \| b_{i} \right \| \left \| b_{j} \right \|  }$, and $\tau$ is the temperature parameter~\cite{SimCLR2020ICML}. Furthermore, the spatio-temporal contrastive loss of all samples within a batch is

\begin{equation}
    \begin{aligned}
    \mathcal{L}_{c}=\frac{1}{2\cdot B}\sum_{i=1}^{B}(\ell_{i}^{1}+\ell_{i}^{2}).
    \end{aligned}
\end{equation}

% Instead of adopting the continuous relaxation with a quantization loss function as SOTA methods~\cite{MCMSH2022MM,NPH2019ICCV,BTH2021CVPR,SSVH2018TIP}, which causes considerable information loss due to the large accumulated quantization errors, we follow~\cite{CIBHash2021IJCAI} to introduce a probabilistic binary representation layer into the model for 

\para{Binary constraint.} The $sgn$ function has a derivative of 0 almost everywhere, which makes it seemingly incompatible with back-propagation. This, in turn, renders the optimization problem above NP-hard and intractable. Following the commonly adopted method~\cite{BTH2021CVPR,NPH2019ICCV}, we use the straight-through estimator~\cite{STE2015NIPS} to solve the binary optimization problem. 

% To reduce the accumulated quantization error, we define the following quantization loss following~\cite{DHN2016AAAI}.
% \begin{equation}
%     \begin{aligned}
%      \mathcal{L}_{q}=\frac{1}{2}\sum_{i=1}^{B}\sum_{k=1}^{K}\sum_{n=1}^{2}\left(\log cosh(|h_{i,k}^{n}|-\mathbf{1}) \right),
%     \end{aligned}
% \end{equation}

\subsection{Frame Order Verification}
\label{sec:frame_order}
Our proposed spatio-temporal contrastive learning with transformer leverages the global information for hash learning but is easily biased to neglect local temporal details~\cite{TimeMatter2022ICML}. To compensate for this deficiency, we follow the classic method~\cite{TimeMatter2022ICML} to define a frame order prediction task to use the temporal coherence as a supplementary supervisory signal.

Different from predicting a relative order between frames or clips~\cite{VideoOrder2016ECCV,VideoOrder2017ICCV}, the transformer's ability to capture long-term dependencies allows us to maintain temporal information within the video, frame-by-frame. This means we can assign the correct frame order as a self-supervised label, predicting the absolute temporal order of video frames. Specifically, we first obtain the representations of two self-augmented clips yielded by the video encoder, \ie, $\left\{f_{i,1}^{1},f_{i,2}^{1}, ..., f_{i,T}^{1}\right\}$ and $\left\{f_{i,1}^{2},f_{i,2}^{2}, ..., f_{i,T}^{2}\right\}$, where $f_{i,j}^{n}$ is the representation of the $j$-th frame of the $n$-th view of $x_{i}$. Then, we train an auxiliary order prediction layer $g_{\xi}$ (\ie, a fully-connected layer followed by a softmax classifier parameterized by $\xi$) to reason the absolute position of each $f_{i,j}^{n}$ in the clip. We define the temporal order $y_{j}^{order}=j$ as the ground-truth label of the $j$-th frame. Therefore, the order verification loss can be described as a classification loss, \ie, 

\begin{equation}
    \begin{aligned}
    \Tilde{\ell}_{i}^{n}=\frac{1}{T}\sum_{j=1}^{T}CE\left(g_{\xi}(f_{i,j}^{n}),y_{j}^{order}\right),
    \end{aligned}
\end{equation}

\begin{equation}
    \begin{aligned}
    \mathcal{L}_{o}=\frac{1}{2\cdot B}\sum_{i=1}^{ B}(\Tilde{\ell}_{i}^{1}+\Tilde{\ell}_{i}^{2}),
    \end{aligned}
\end{equation}

where $\Tilde{l}_{i}^{n}$ is the loss of the $n$-th clip with $T$ frames augmented from $x_{i}$, and $CE(x,y)$ denotes the standard cross-entropy loss between an input $x$ and a given label $y$, respectively.

\subsection{Scene Change Regularization}
\label{sec:scene_change}

Scene changes are ubiquitous in videos, especially in lengthy ones. As shown in Figure~\ref{fig:tasks}, scene changes bring a large shift in both the spatial background and the temporal motion. However, the attention-based transformer encoder tends to smooth temporal differences, weakening the discriminatory power of the contrastive model. Therefore, we propose a task for regulating scene changes that attracts similar frames while repelling frames that differ significantly from each other. We achieve this by leveraging prototypical contrastive learning~\cite{PCL2021ICLR} to regulate the model's ability to perceive spatio-temporal changes.

We collect all the frame-level representations  $\left\{f_{i,1}^{1},f_{i,2}^{1}, ..., f_{i,T}^{1}\right\}$ and $\left\{f_{i,1}^{2},f_{i,2}^{2}, ..., f_{i,T}^{2}\right\}$ of $x_{i}$ from the output of the transformer-based temporal encoder. Then, all these representations within a video will be clustered by the Affinity Propagation algorithm~\cite{APClustering}, where each video can be adaptively divided into multiple scenes without the need to specify the number of scenes in advance. Intuitively, every cluster can be described as a common scene, where the prototype $c_{i}$ represents the semantic center of this scene. Therefore, \ourmethod contrasts between frame-level representations and scene prototypes with prototypical contrastive learning~\cite{PCL2021ICLR} to encourage frames within the same scene to be similarly represented while frames from different scenes are exactly the opposite. The prototypical contrastive loss is described below.
% \begin{figure}[t]
%     \centering
%     \includegraphics[width=\linewidth]{fig/scene_change.pdf}
%     \caption{An illustration of a scene change in a video. The table tennis player is fighting the game at the beginning, and then he is celebrating his victory.}
%     \label{fig:scence_change}
%     \vspace{-1.0em}
% \end{figure}

\begin{equation}
\footnotesize
   \begin{aligned}
    \bar{\ell}_{i,j}^{n}\!=-\log \frac{exp\left(sim(f_{i,j}^{n},\mathcal{C}(f_{i,j}^{n}))/\tau\right)}{exp\left(sim(f_{i,j}^{n},\mathcal{C}(f_{i,j}^{n}))/\tau\right)\!+\!\sum\limits_{c_{k}\neq \mathcal{C}(f_{i,j}^{n}) }\!exp\left(sim(f_{i,j}^{n},c_{k})/\tau\right)},
\end{aligned}
\end{equation}
where $\ell_{i}^{n}$ is the prototypical contrastive loss of the $n$-th ($n =1\text{ or }2$) view of $f_{i}$, $\mathcal{C}(f_{i,j}^{n})$ is the prototype that corresponds to $f_{i,j}^{n}$. The scene change regularization loss is defined below.
\begin{equation} 
    \begin{aligned}
     \mathcal{L}_{s}=\frac{1}{2\cdot B\cdot T}\sum_{i=1}^{B}\sum_{j=1}^{T}\left(\bar{\ell}_{i,j}^{1}+\bar{\ell}_{i,j}^{2}\right).
    \end{aligned}
\end{equation}

\para{Overall learning objective.} Combining the three tasks, the overall learning objective can be derived as follows:

\begin{equation}
    \begin{aligned}
     \mathcal{L}= \mathcal{L}_{c}+  \mathcal{L}_{o} +  \mathcal{L}_{s},
    \end{aligned}
\end{equation}
which will jointly optimize the model by fully exploiting the multi-granularity Spatio-temporal context.

\begin{figure*}[thb]
    \includegraphics[width=0.9\textwidth]{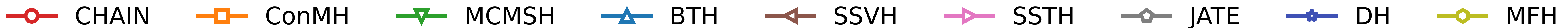}\\
    \subfloat[ActivityNet 16 bits]{\includegraphics[width=0.24\textwidth]{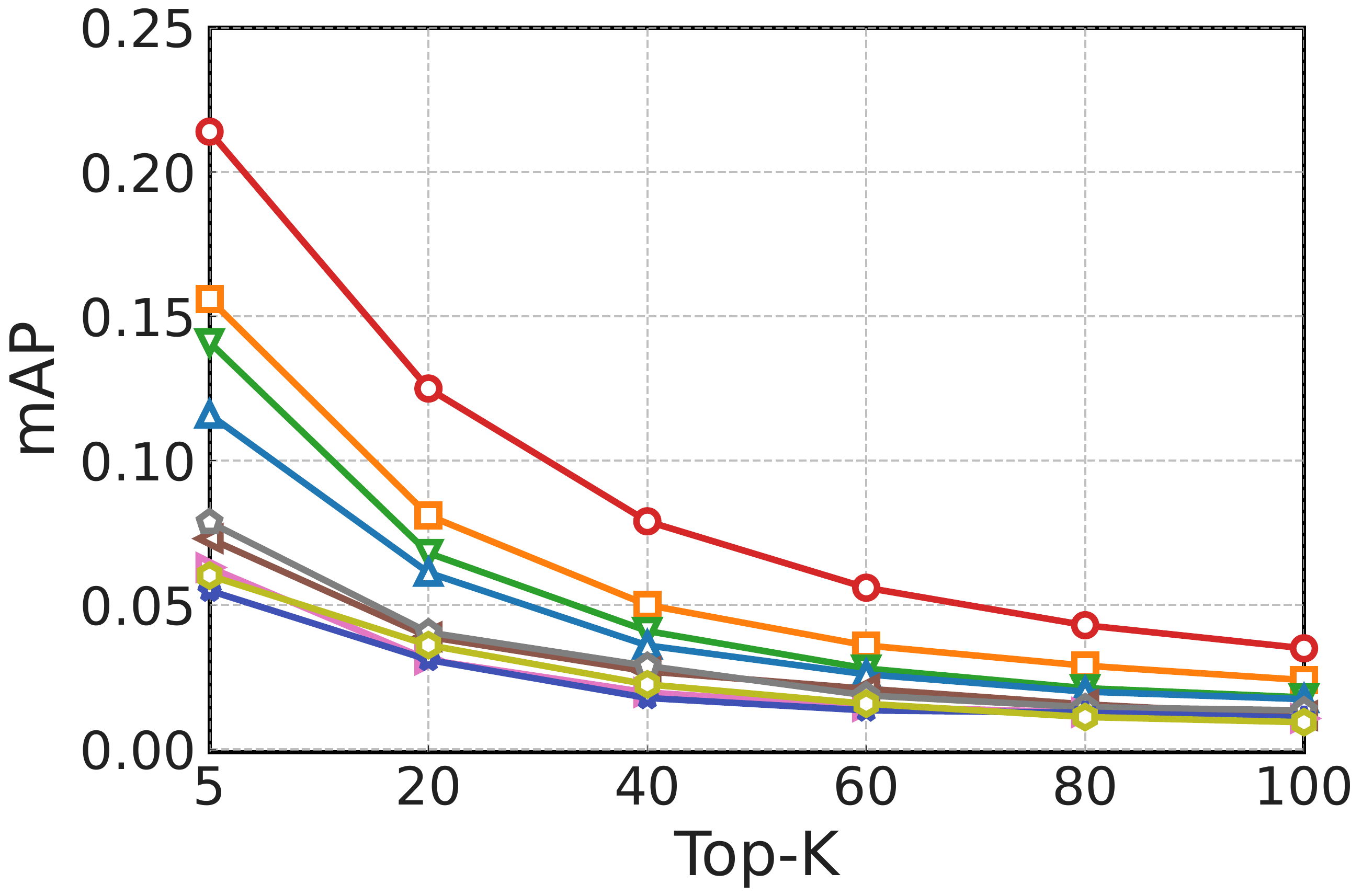}}
    \hfill
     \subfloat[FCVID 16 bits]{\includegraphics[width=0.24\textwidth]{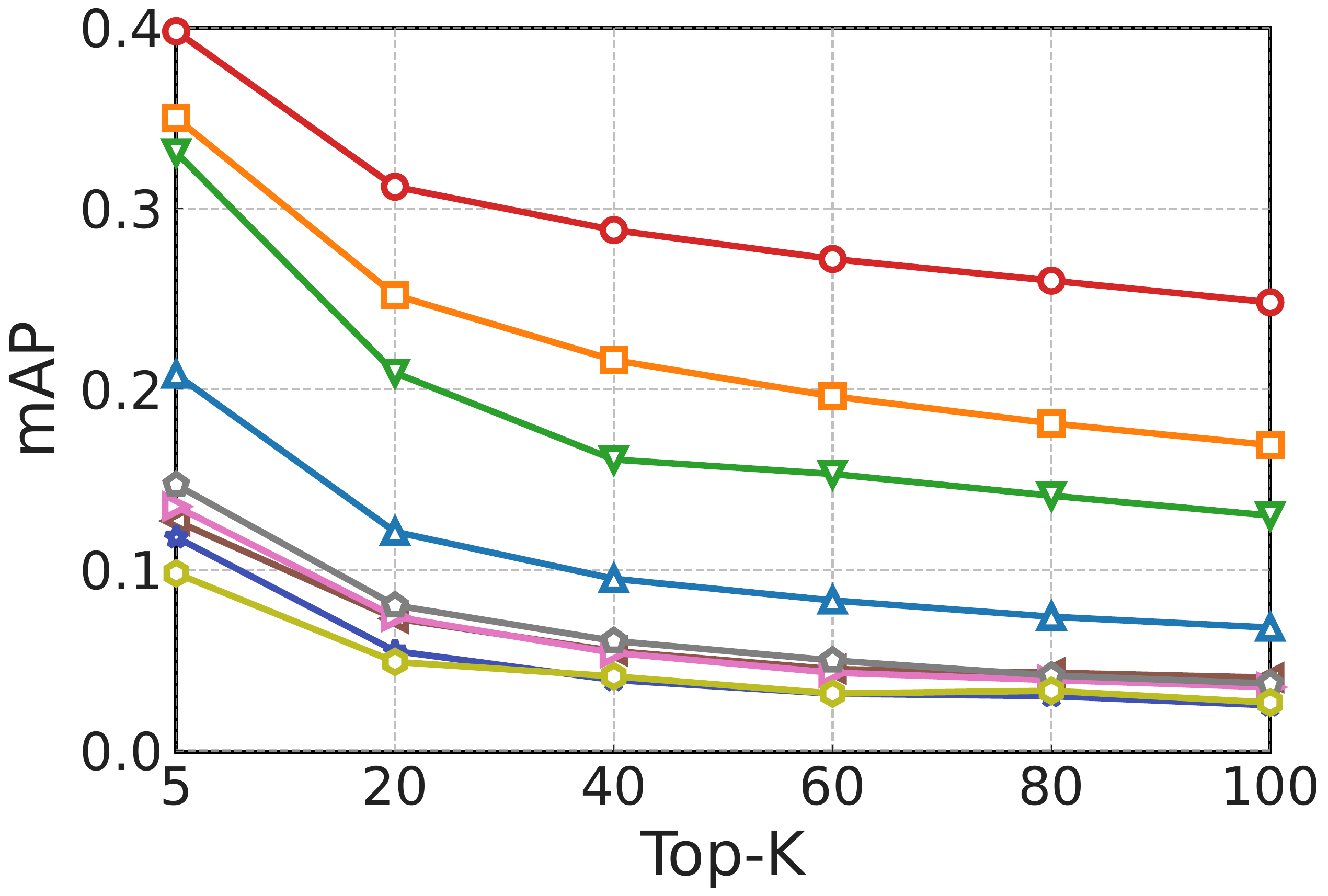}}
     \hfill
      \subfloat[UCF101 16 bits]{\includegraphics[width=0.24\textwidth]{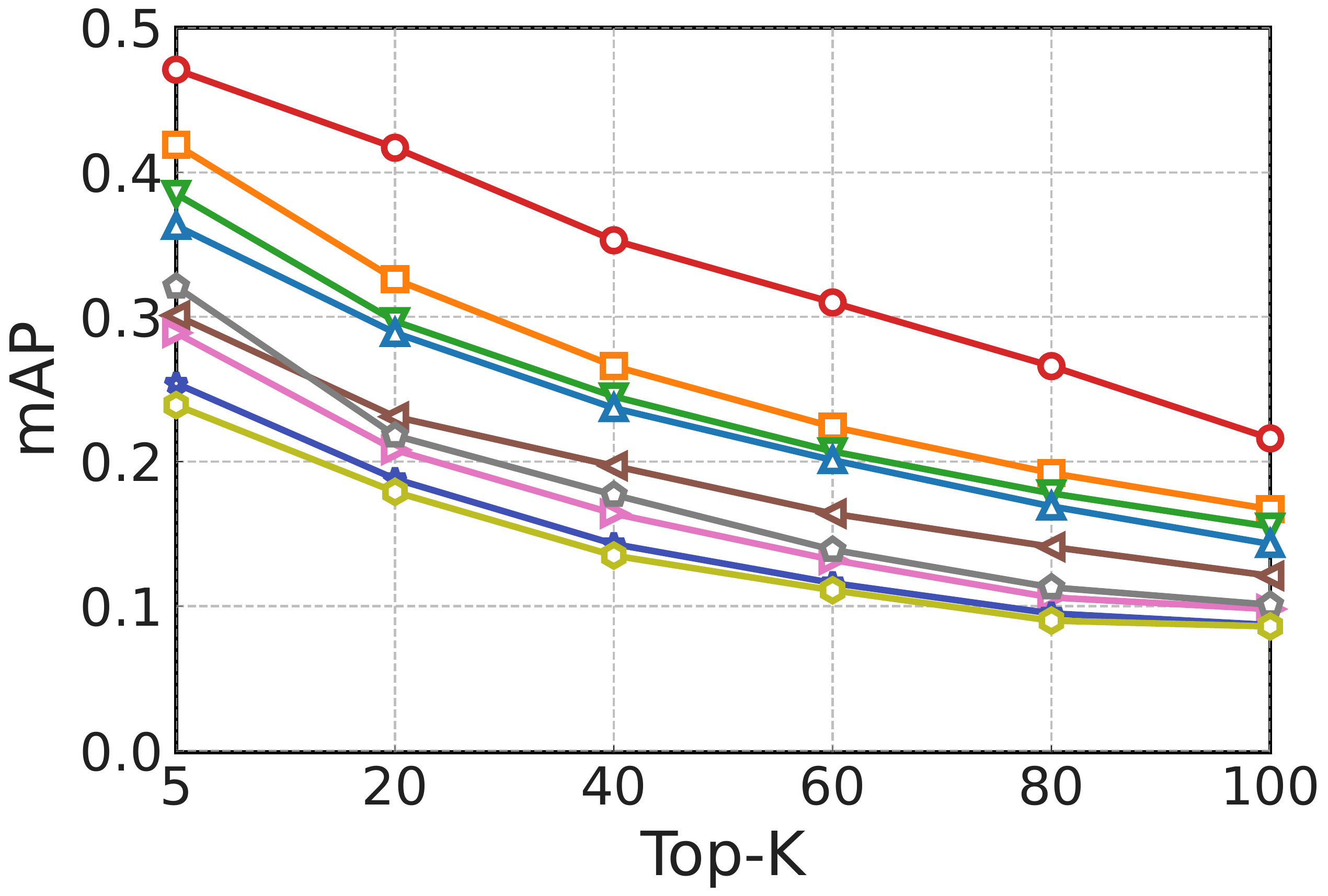}}
      \hfill
       \subfloat[HMDB51 16 bits]{\includegraphics[width=0.24\textwidth]{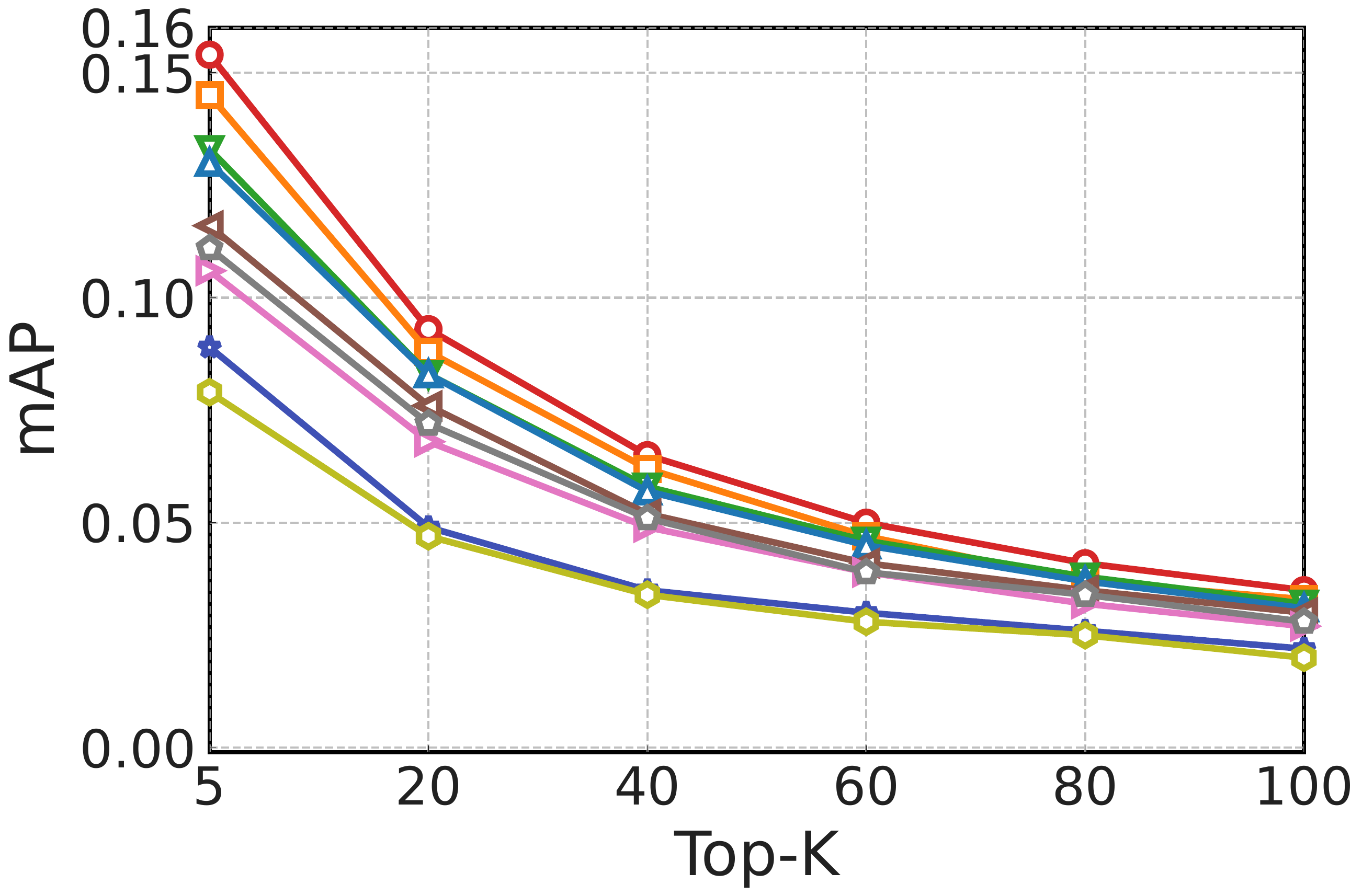}}
       \\
    \subfloat[ActivityNet 32 bits]{\includegraphics[width=0.24\textwidth]{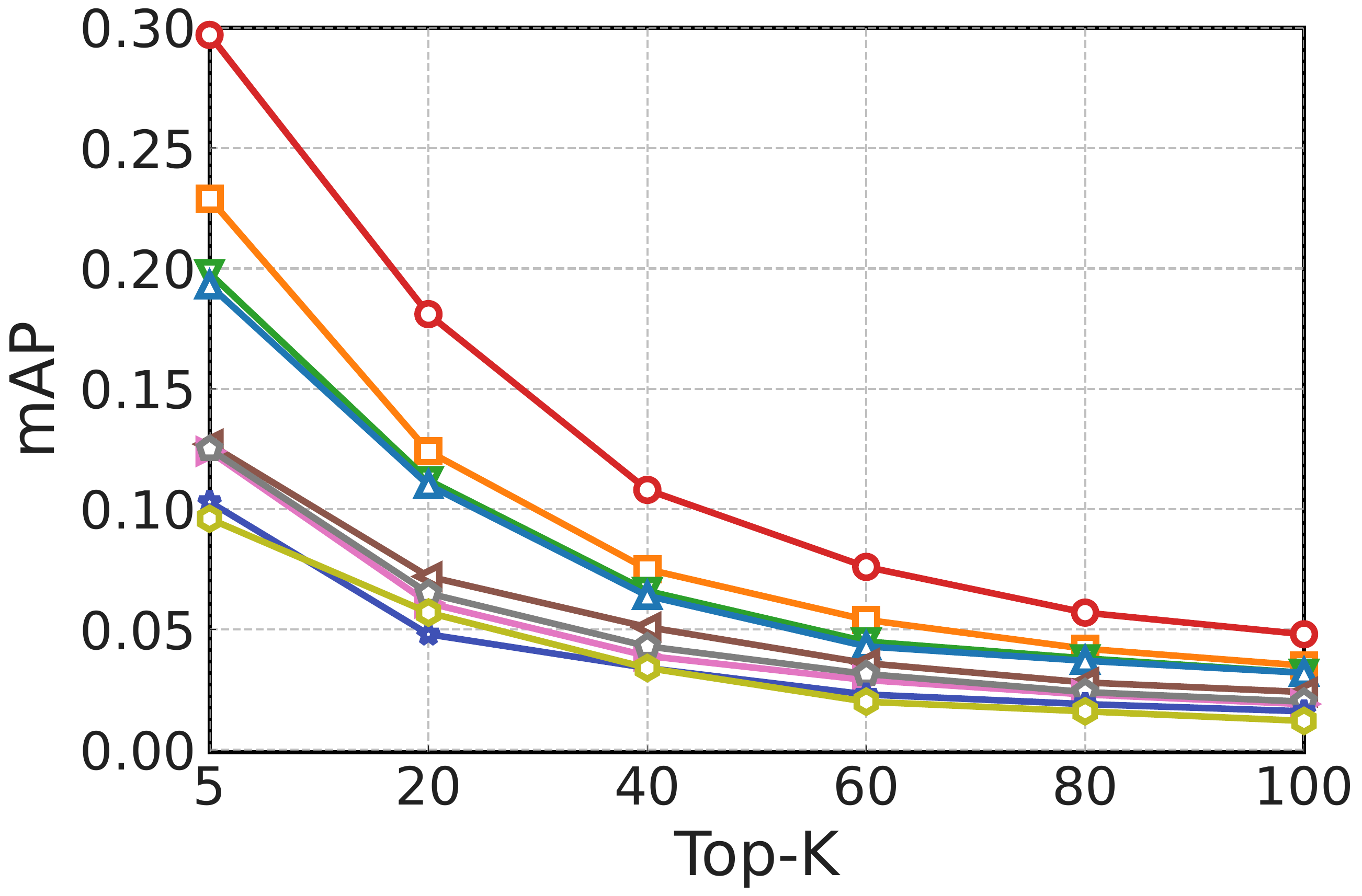}}
    \hfill
    \subfloat[FCVID 32 bits]{\includegraphics[width=0.24\textwidth]{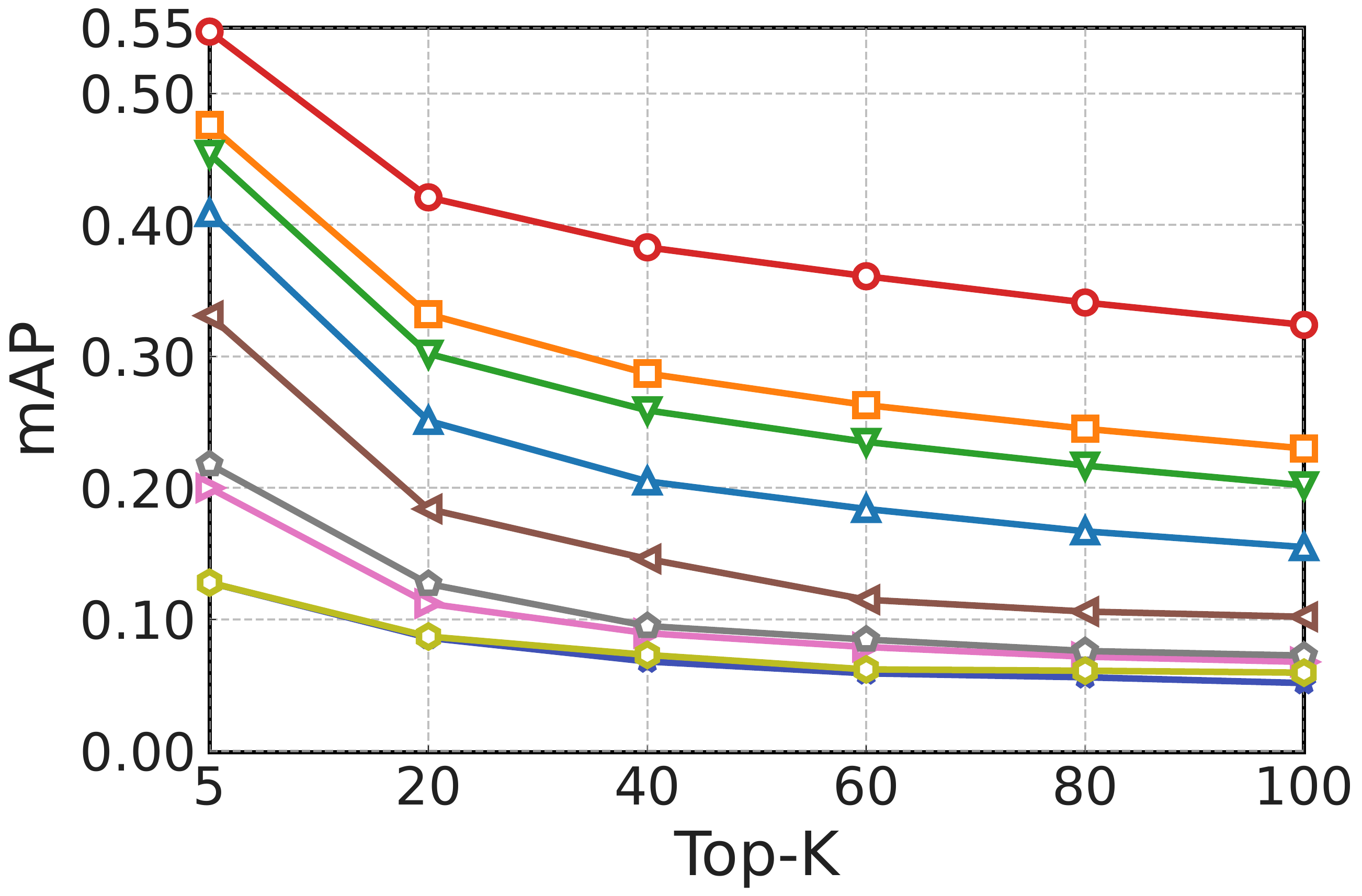}}
    \hfill
    \subfloat[UCF101 32 bits]{\includegraphics[width=0.24\textwidth]{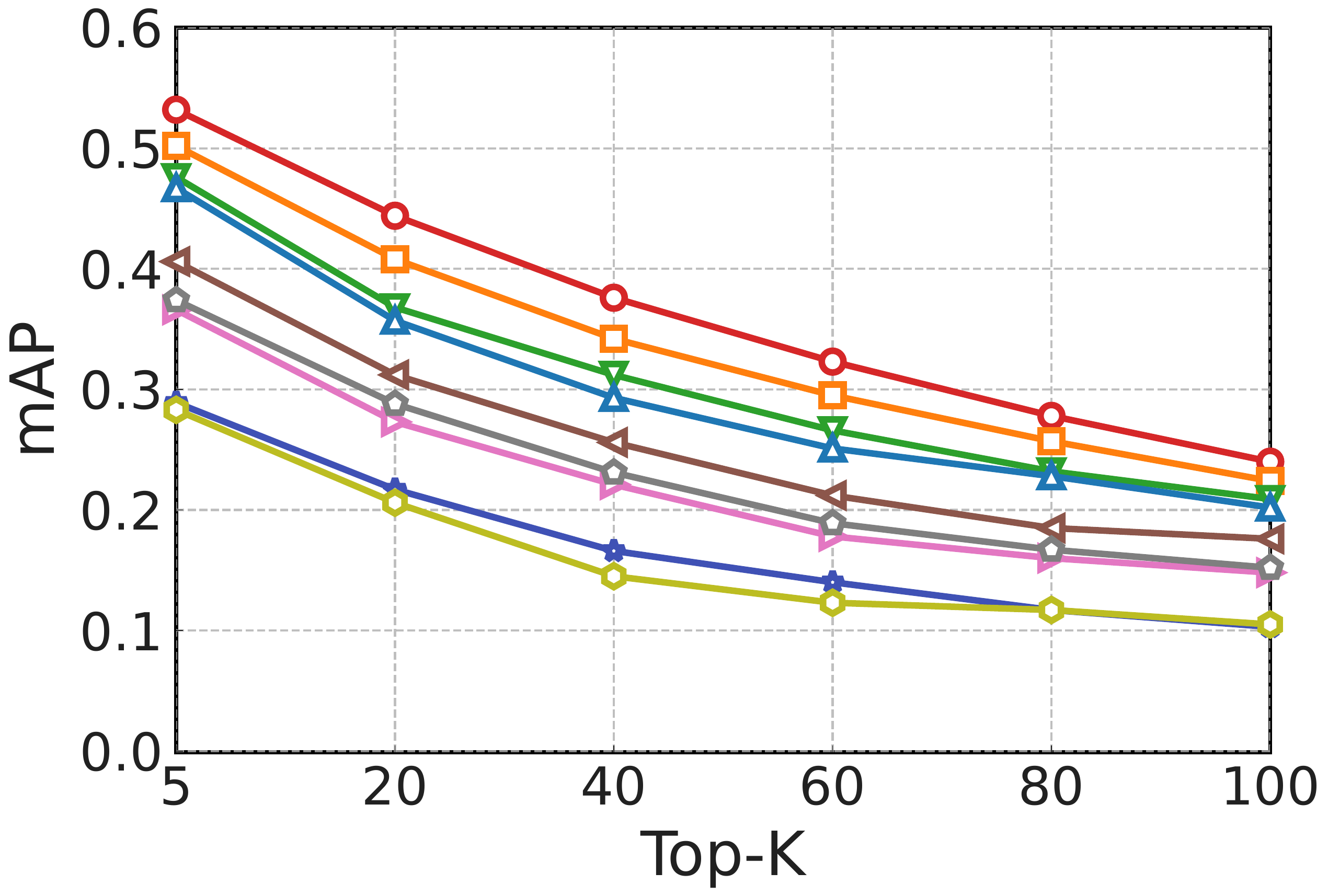}}
    \hfill
    \subfloat[HMDB51 32 bits]{\includegraphics[width=0.24\textwidth]{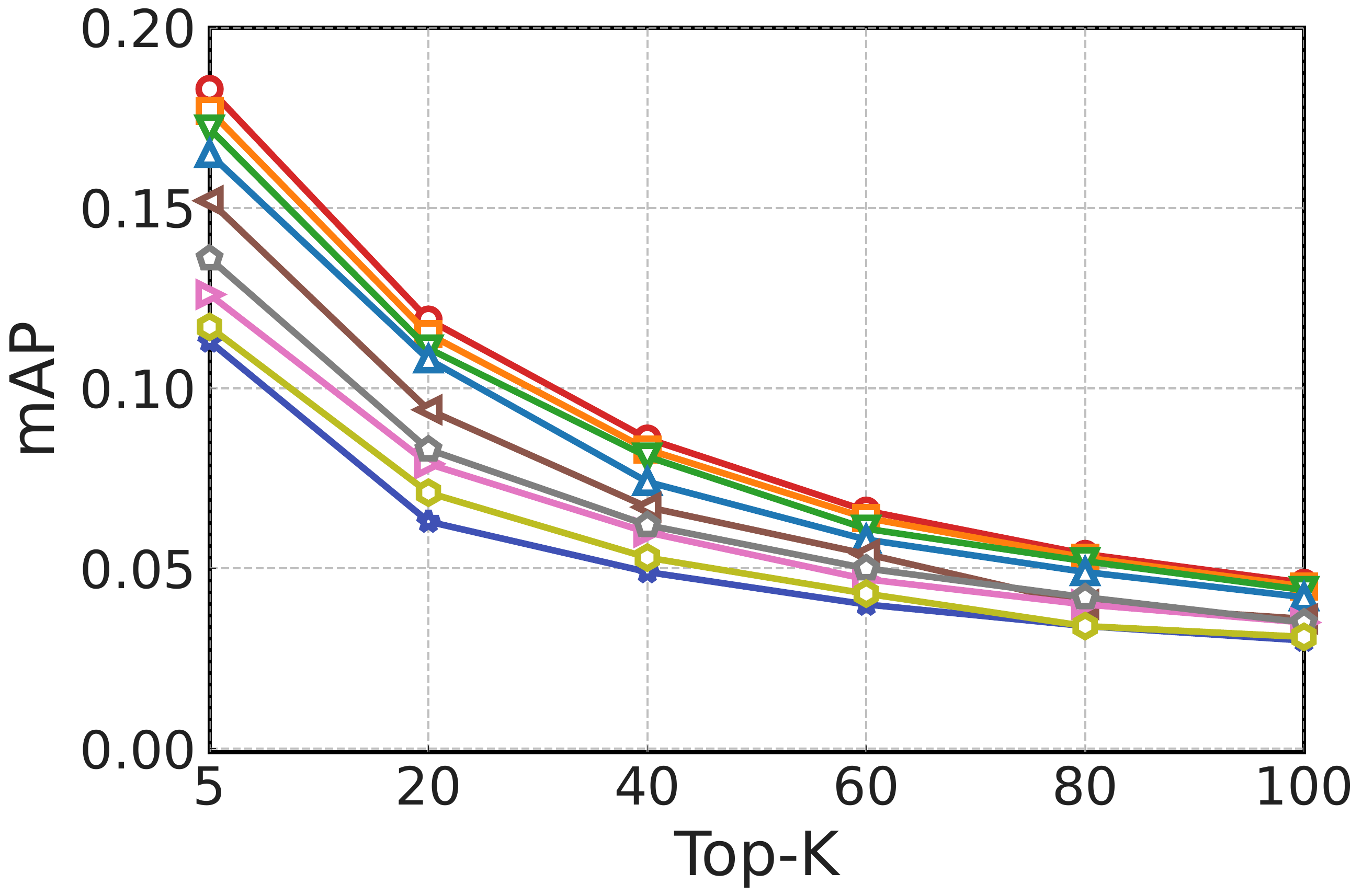}}\\
    \subfloat[ActivityNet 64 bits]{\includegraphics[width=0.24\textwidth]{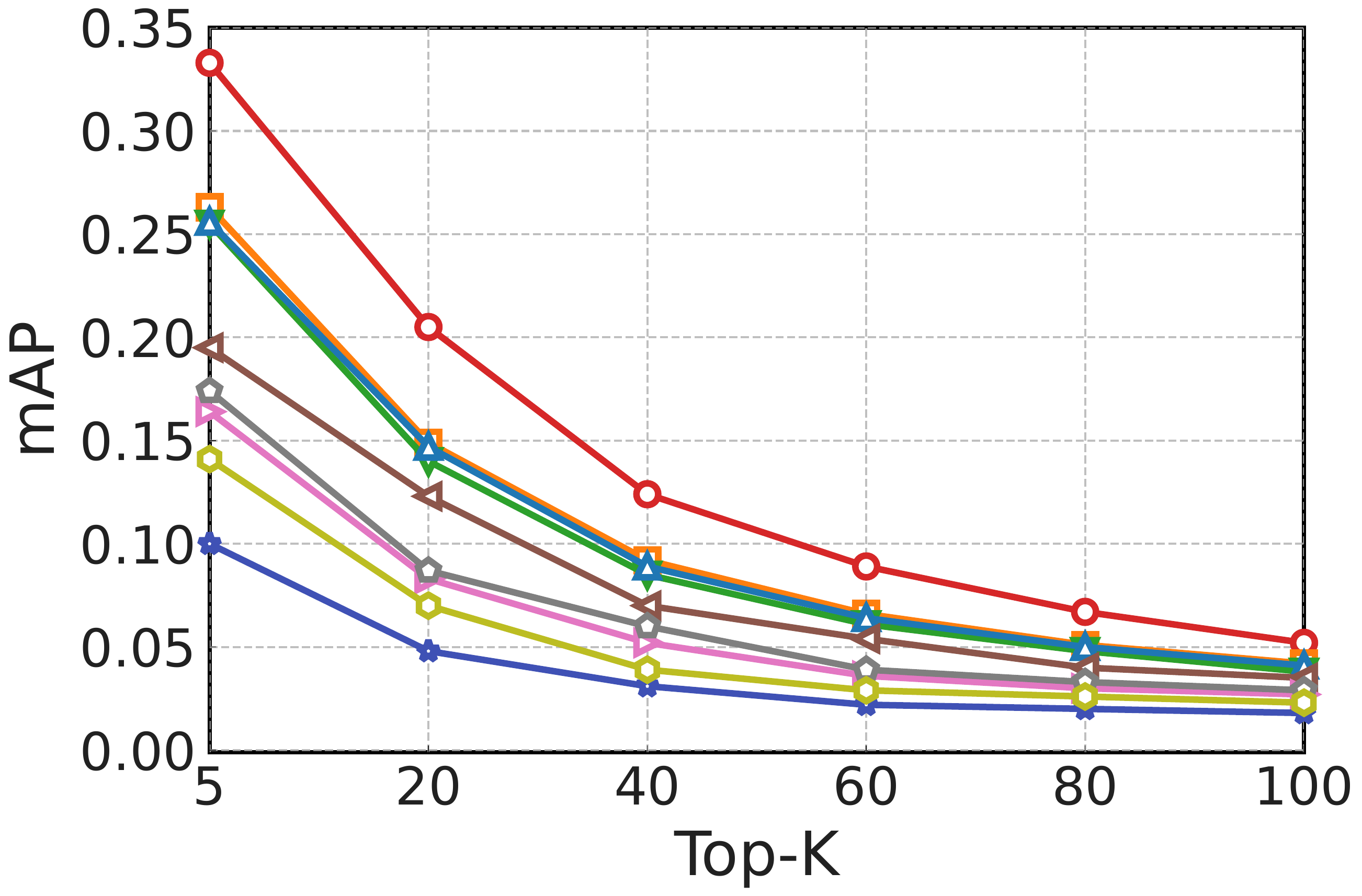}}
    \hfill
    \subfloat[FCVID 64 bits]{\includegraphics[width=0.24\textwidth]{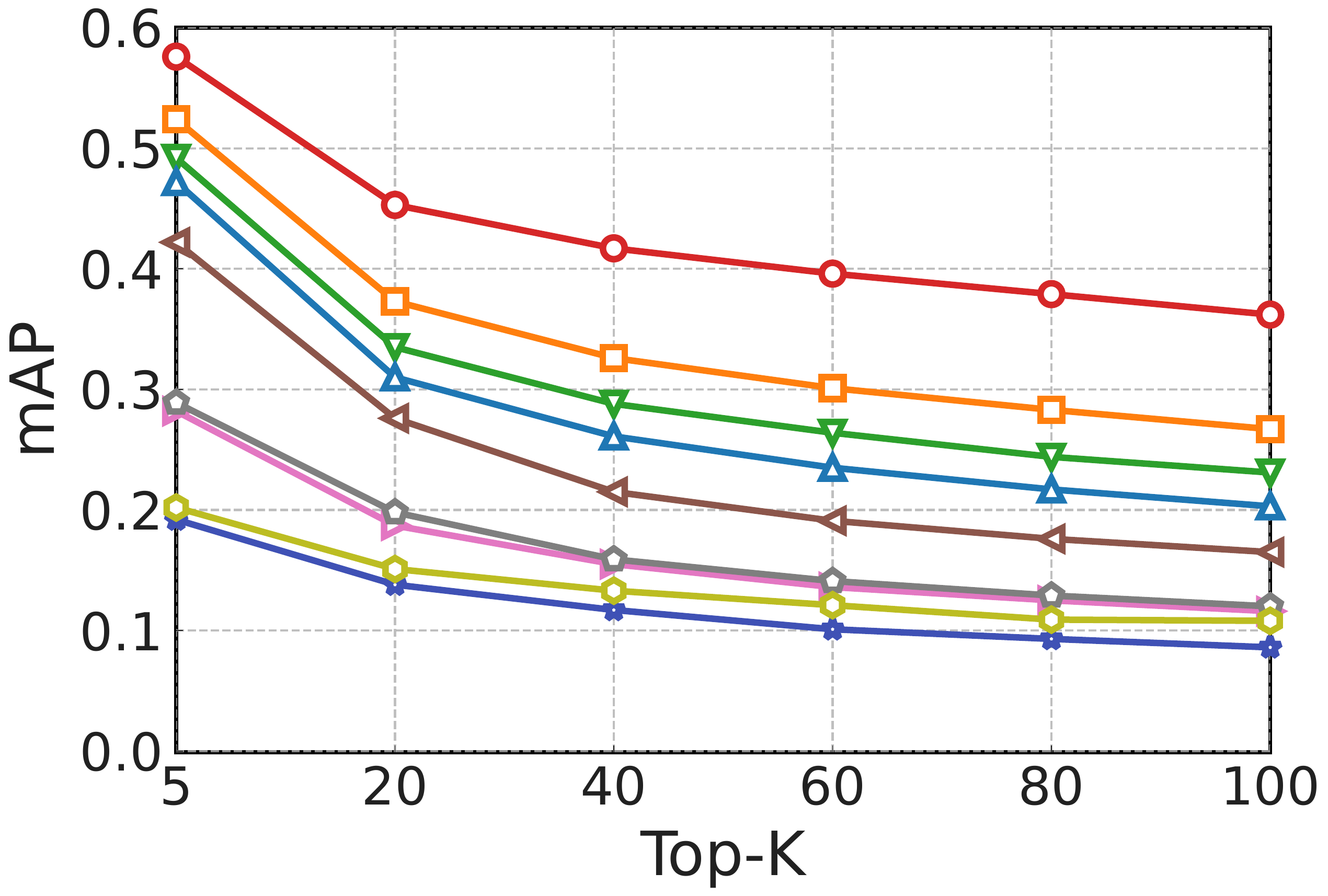}}
    \hfill
    \subfloat[UCF101 64 bits]{\includegraphics[width=0.24\textwidth]{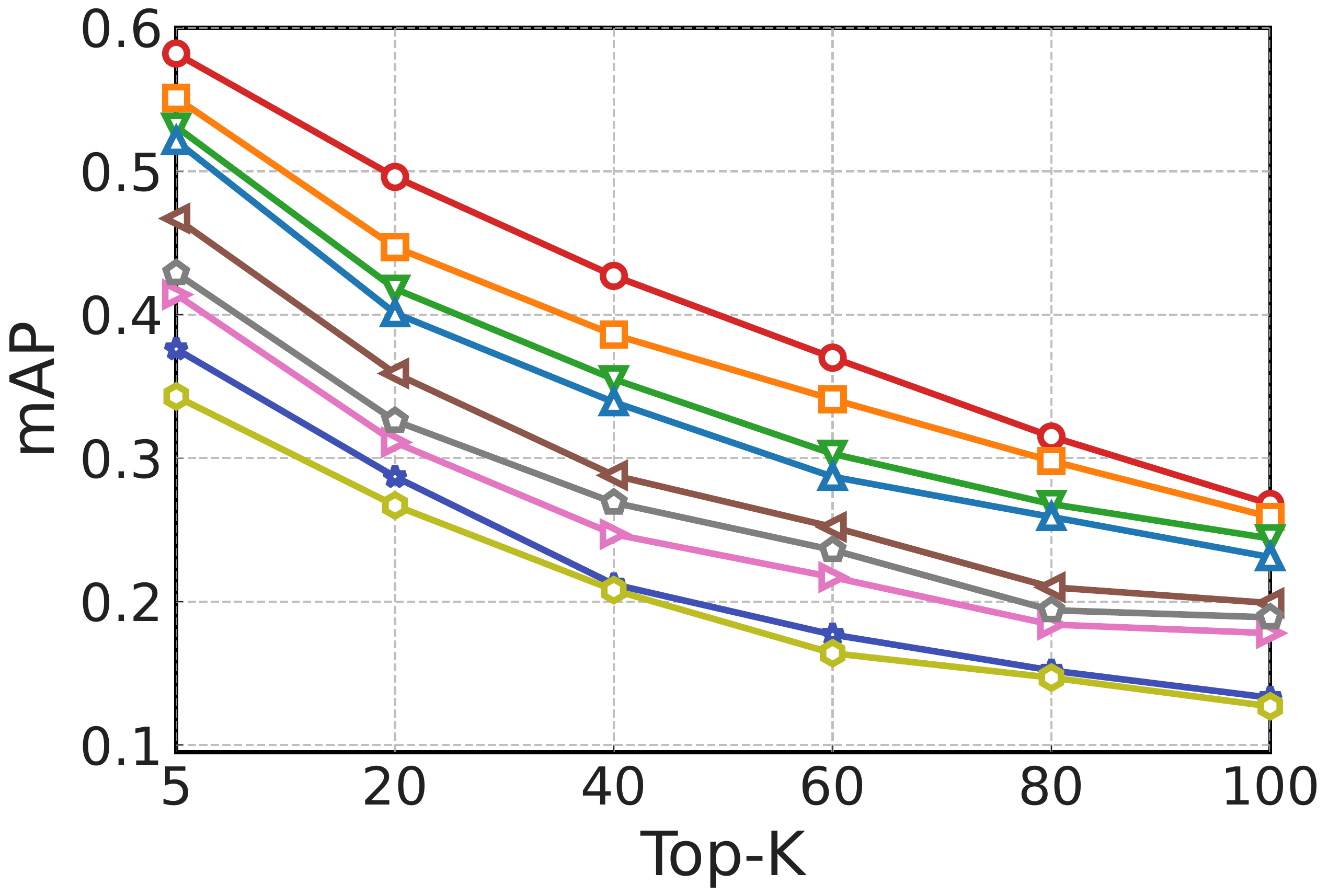}}
    \hfill
    \subfloat[HMDB51 64 bits]{\includegraphics[width=0.24\textwidth]{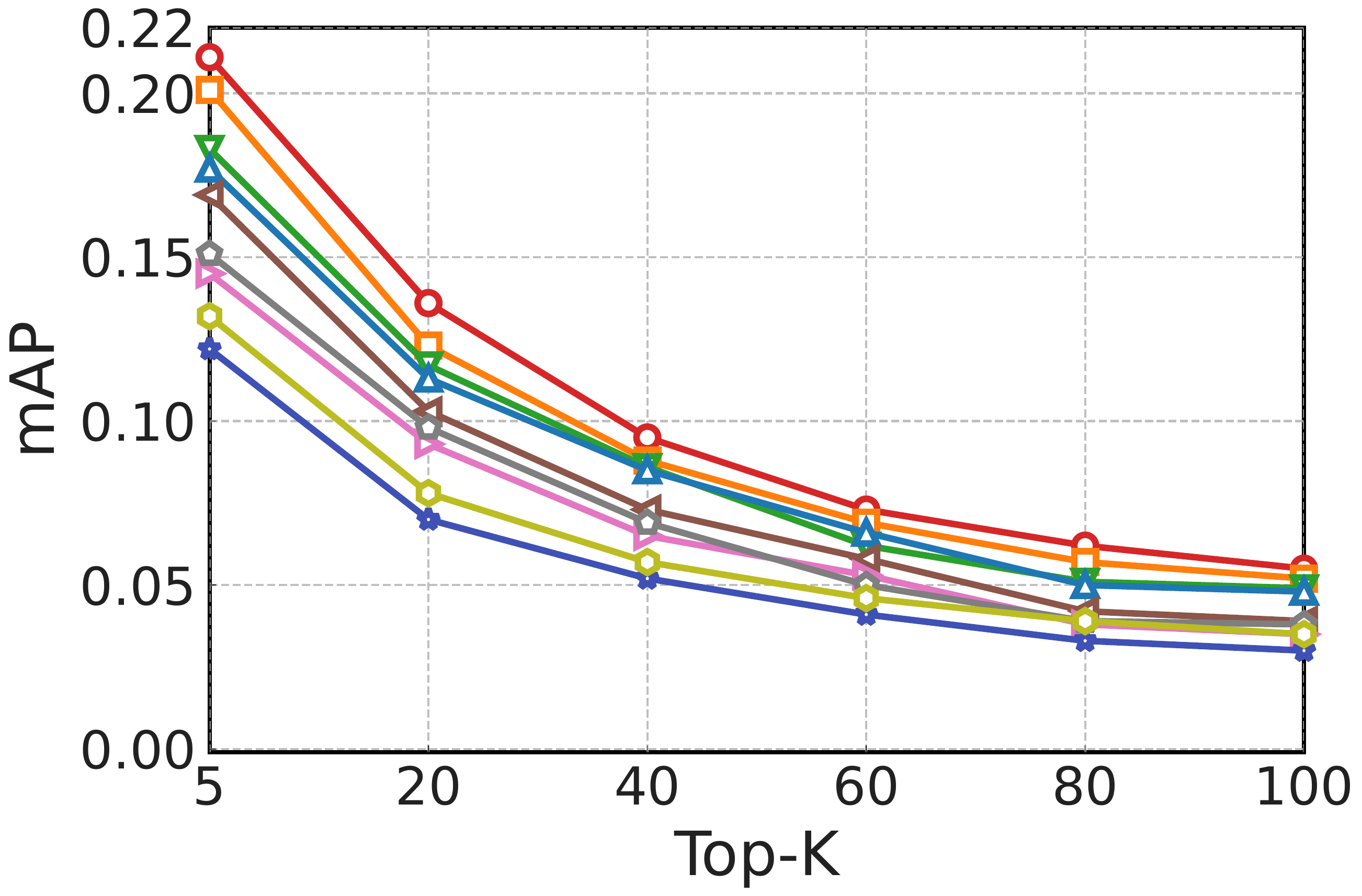}}
    \caption{Retrieval performance compared with state-of-the-art methods in terms of mAP@K over four datasets.}
    \label{fig:map_all}
\end{figure*}

\section{Experiments}
In this section, we conduct extensive experiments on four public benchmarks to demonstrate the superiority of our proposed \ourmethod compared to state-of-the-art self-supervised video hashing methods. We also provide ablation studies and visualizations to verify the rationality of the design. More interesting  experimental exploration can be found in the \textbf{supplementary material}.

\subsection{Dataset and Evaluation Protocols}

The datasets include UCF101~\cite{UCF-1012012arxiv}, HMDB51~\cite{HMDB2011ICCV}, FCVID~\cite{FCVID2018TPAMI}, ActivityNet~\cite{ActivityNet2015CVPR}. In addition, we advocate the settings of ~\cite{SSTH2016MM,SSVH2018TIP,BTH2021CVPR,MCMSH2022MM} to split the datasets for a fair comparison.

\noindent\textbf{FCVID} consists of 91,223 videos from 239 annotated categories. We follow~\cite{NPH2019ICCV,BTH2021CVPR,SSVH2018TIP} to employ 91,185 videos, of which 45,585 are used for training and the other 45,600 for evaluation.

\noindent\textbf{UCF101} contains 13,320 videos spread across 101 categories of human actions. Since both the UCF101 and HMDB51 datasets provide official train/test splits, following classic methods~\cite{CSQ2020CVPR,BitEntropy}, we adopt the default setting of using 9,537 videos for training and retrieval and 3,783 videos from the test split as the query set.

\noindent\textbf{HMDB51} contains 6,849 videos spread across 51 action categories. Referring to~\cite{CSQ2020CVPR,BitEntropy}, we use 3,570 videos for training and retrieval, and 1,530 videos from the test split as the query set.

\noindent\textbf{ActivityNet v1.3} covers a broad range of human activities with 200 categories and a total of 14,950 videos, including 10,024 training videos and 4,926 validation videos. Due to some missing and damaged videos, we follow~\cite{BTH2021CVPR,MCMSH2022MM} to use 9,992 videos as the training set and exploit the validation set as our test set. In addition, we randomly select 1,000 videos for queries and 3,919 videos as the retrieval set.

\noindent\textbf{Evaluation protocols} Following the same evaluation protocol as~\cite{NPH2019ICCV,BTH2021CVPR,ConMH2023AAAI,MCMSH2022MM}, we utilize the mean Average Precision at top-K retrieved results (mAP@K) to evaluate the retrieval performance, where videos are sorted according to their Hamming distance and similar ones are required to be ranked higher in the retrieved result. Besides, we use Precision-Recall (PR) curves as an additional evaluation metric to provide a more comprehensive evaluation of the retrieval performance by considering the trade-off between precision and recall. We evaluate the retrieval results with code lengths of 16, 32, and 64 bits, following most baselines.
 
\subsection{Implementation Details}
\para{Feature preparation} To ensure fair comparisons, we follow~\cite{MCMSH2022MM,BTH2021CVPR,NPH2019ICCV} to sample 25 (\ie, $T=25$) frames for each video from UCF101, HMDB51, and FCVID datasets. We use VGG-16~\cite{vgg2014ICLR} as the 2D CNN encoder to extract 4096-D frame-level features. For the ActivityNet dataset, we follow~\cite{BTH2021CVPR,MCMSH2022MM} to sample 30 (\ie, $T=30$) frames per video and use ResNet-50~\cite{ResNet2016CVPR} to extract 2048-D frame-level features. Both VGG-16 and ResNet-50 are pre-trained on ImageNet~\cite{ImageNet2009CVPR}, and features are pre-extracted before the training phase.

\para{Model architecture.} We use a single-layer transformer with a single attention head as the temporal attention-based encoder in our base model, and we also provide experimental results using different encoders, \eg, MC-MLP~\cite{MCMSH2022MM}. Besides, we use a two-layer MLP, whose output dimension is $K$, as the hash layer. For the frame order prediction layer, we adopt single-layer MLP to predict the absolute position.

\para{Training details.} We set the batch size $B=128$ and the initial learning rate as 0.0001, which will be decayed to $90\%$ every 20 epochs with a minimal learning rate of 0.00001, following~\cite{ConMH2023AAAI}. We optimize our model by using the Adam optimizer algorithm~\cite{Adam2015ICLR} with momentum 0.9. We implement our method in Pytorch with a single NVIDIA RTX 3090 GPU.

\begin{figure}
    \centering
    \subfloat[ActivityNet 64 bits]{\includegraphics[width=0.48\linewidth]{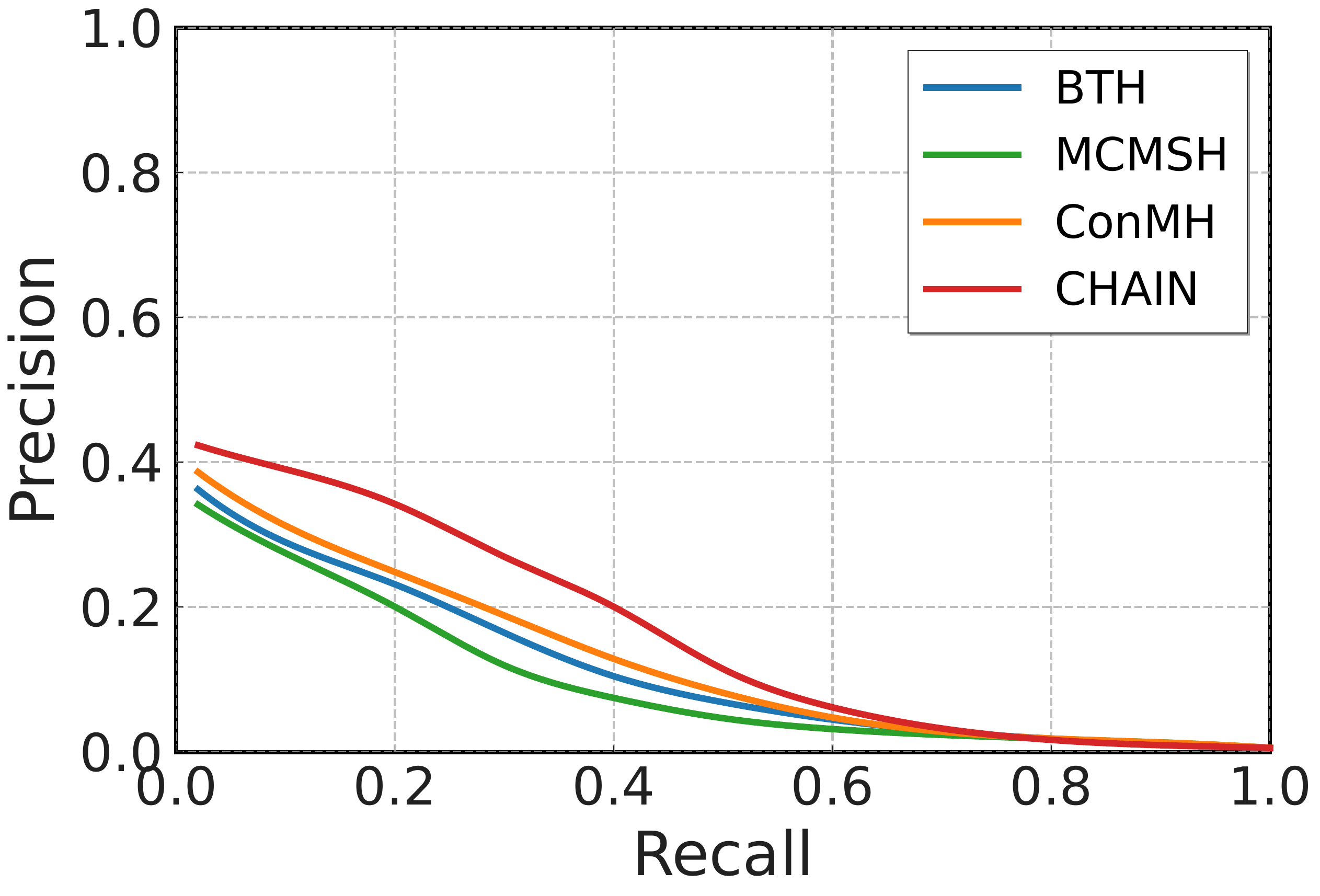}}
    \hfill
     \subfloat[FCVID 64 bits]{\includegraphics[width=0.48\linewidth]{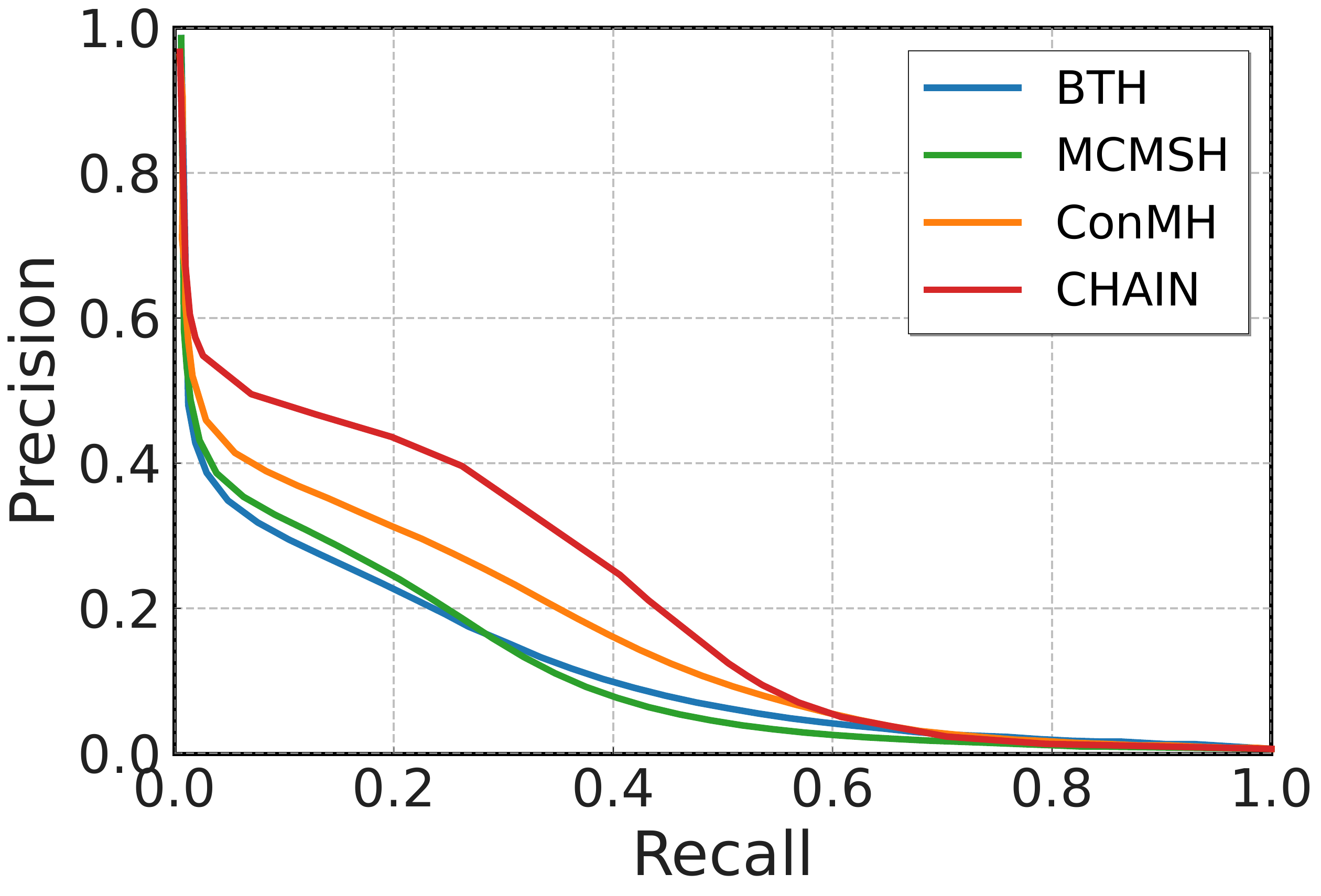}}
     \\
      \subfloat[ActivityNet 32 bits]{\includegraphics[width=0.48\linewidth]{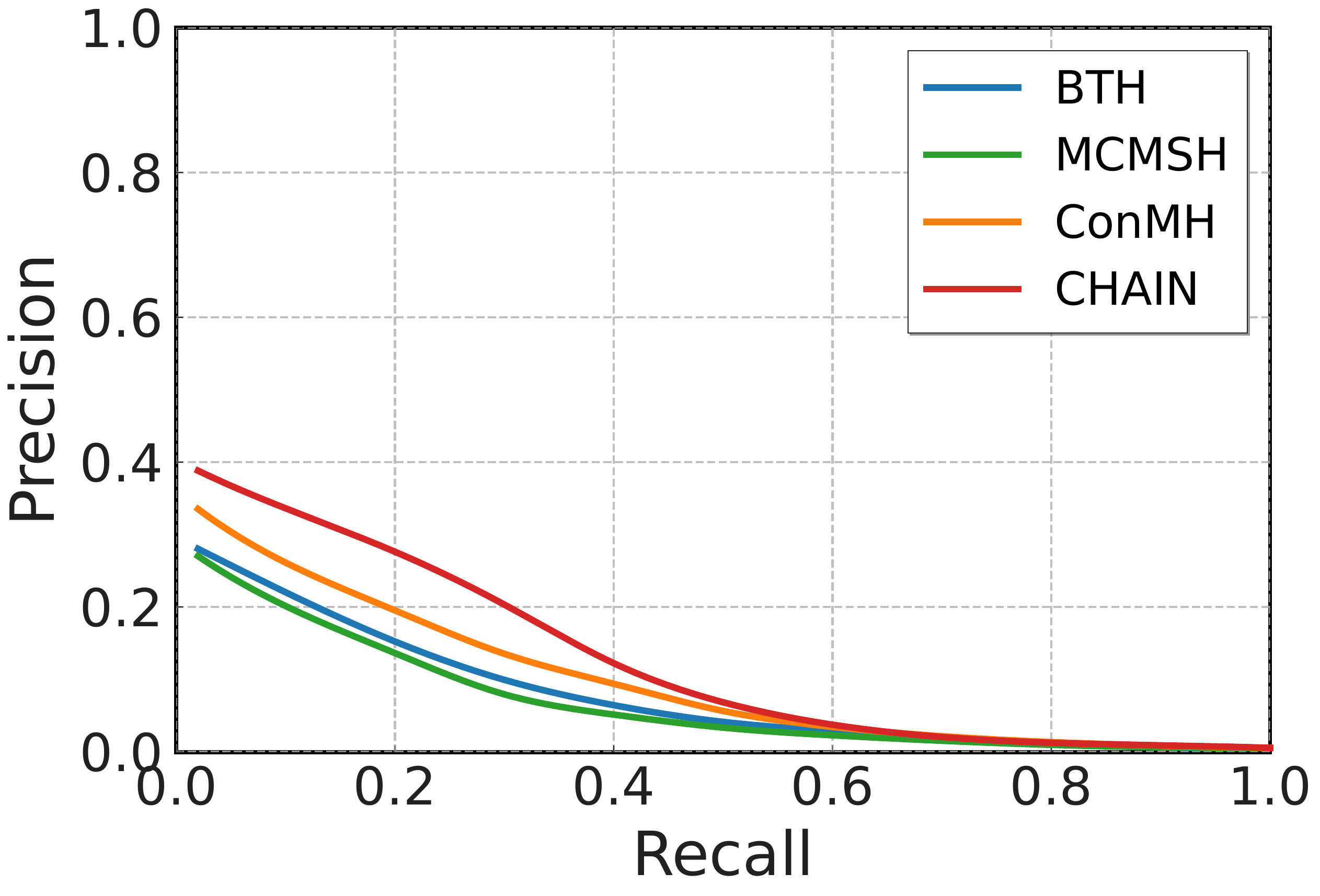}}
      \hfill
       \subfloat[FCVID 32 bits]{\includegraphics[width=0.48\linewidth]{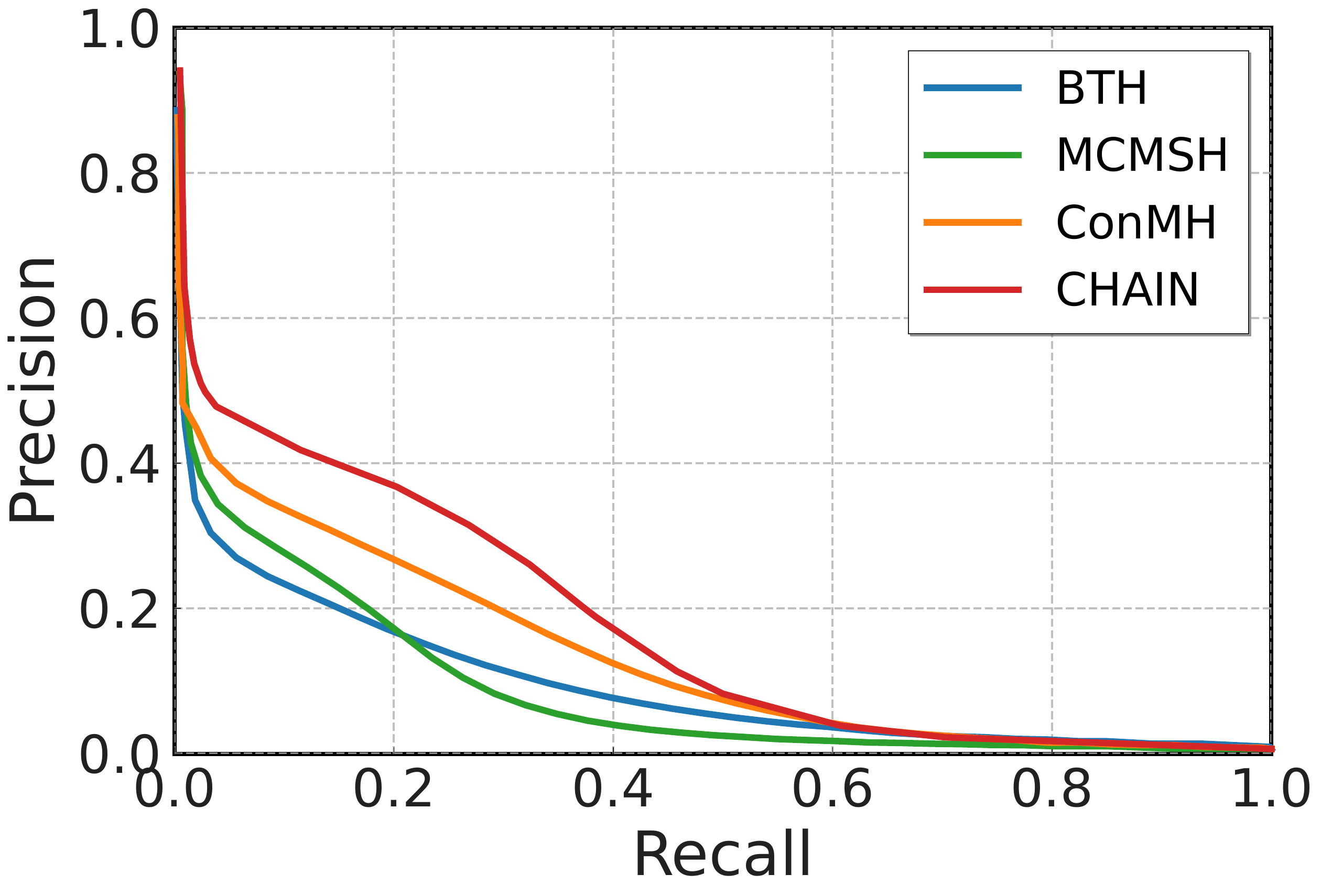}}
       \\
        \subfloat[ActivityNet 16 bits]{\includegraphics[width=0.48\linewidth]{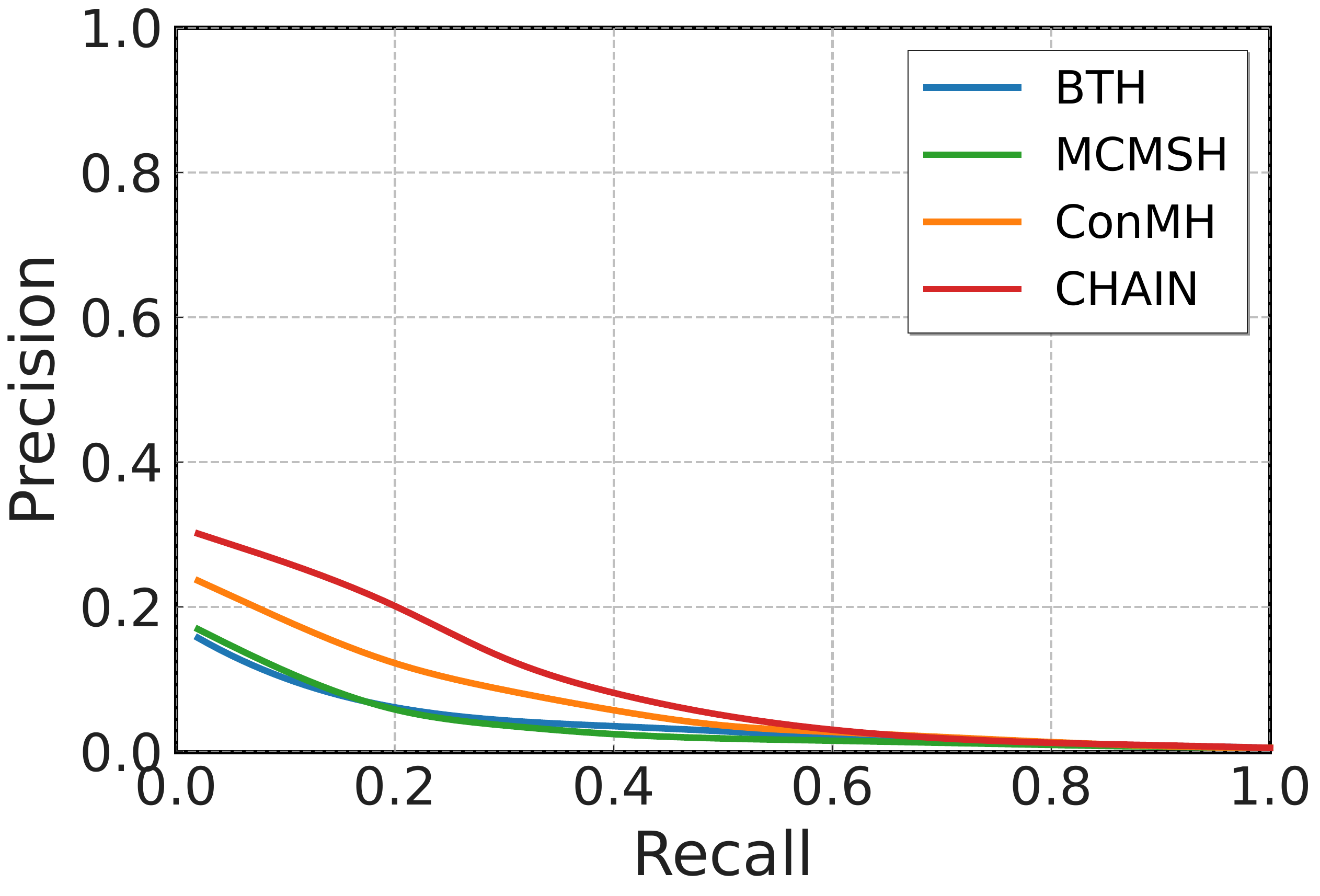}}
        \hfill
         \subfloat[FCVID 16 bits]{\includegraphics[width=0.48\linewidth]{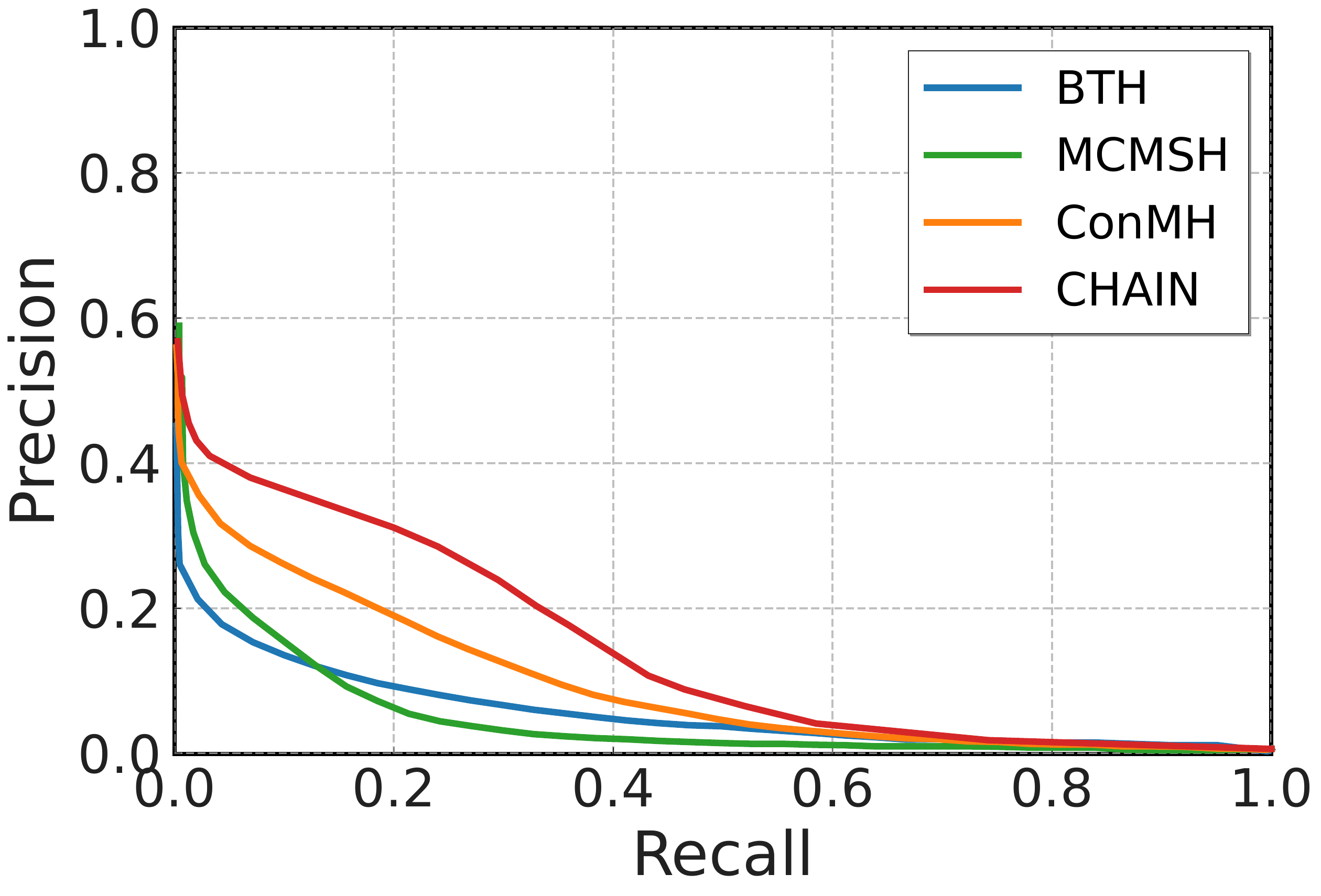}}
    \caption{PR curves of the proposed \ourmethod and the state-of-the-art baselines, including ConMH, MCMSH, and BTH on FCVID and ActivityNet.}
    \label{fig:pr_curve}
    \vspace{-1em}
\end{figure}

\subsection{Comparison with State-of-the-arts}
We compare \ourmethod with several state-of-the-art self-supervised video hashing methods, including MFH~\cite{MFH2011MM}, DH~\cite{DH2015CVPR}, JTAE~\cite{JTAE2017CIKM}, SSTH~\cite{SSTH2016MM}, SSVH~\cite{SSVH2018TIP}, BTH~\cite{BTH2021CVPR}, MCMSH~\cite{MCMSH2022MM}, and ConMH~\cite{ConMH2023AAAI}. For the results of baseline methods, we refer to ConMH~\cite{ConMH2023AAAI}.

\para{mAP comparisons.}
As shown in Figure~\ref{fig:map_all}, \ourmethod outperforms all the compared hashing methods, yielding remarkable results at all code lengths. Specifically, on the FCVID, UCF101, ActivityNet, and HMDB51 datasets, the mAP@20 values of \ourmethod are $21.4\%$, $10.9\%$, $37.5\%$, and $10.5\%$ higher than ConMH, respectively, with 64-bit hash codes. Note that \ourmethod significantly outperforms ConMH, even with low-bit hash codes, \eg, up to $23.8\%$, $27.9\%$, $54.3\%$, and $5.6\%$ performance improvements on the four datasets at 16 bits, demonstrating its superiority in scenarios requiring high real-time performance and low storage demand. We attribute this advantage to the collaborative learning tasks, in which the contrastive learning task assists in capturing the video semantics by combining motion and stillness from a global perspective, and the order verification task and the scene change regularization task effectively mine the local temporal details.

We also observe that the improvement on HMDB51 is smaller than that on the other three datasets. We attribute this to two reasons. Firstly, the limited number of samples per class in HMDB51 is insufficient to fully stimulate the potential of the model. Secondly, inherent difficulties in motion recognition, such as shorter video duration, greater variability in camera viewpoints, and more complex interactions, make HMDB51 challenging. These challenges motivate us to explore more effective solutions in the future.

\para{PR curves comparison.}
The precision-recall curves of BTH, MCMSH, ConMH, and \ourmethod on FCVID and ActivityNet are shown in Figure~\ref{fig:pr_curve}. It is clear that \ourmethod achieves higher precision at the same recall rate than other methods. These results demonstrate that \ourmethod can achieve stable and superior retrieval performance. In particular, when a relatively low recall rate is acceptable, the precision advantage of \ourmethod becomes even more prominent.

\begin{table}[t]
    \centering
    \caption{ Cross-dataset mAP@20 retrieval results and corresponding performance degrades of various methods, which are
trained on FCVID and tested on UCF101 and  HMDB51.}
\scalebox{0.88}{
    \begin{tabular}{ccccc}
    \toprule
         Method&BTH&MCMSH&ConMH&\ourmethod\\
         \midrule
         UCF101& 0.244  ($\downarrow 39\%$) &0.252 ($\downarrow 40\%$) &0.278 ($\downarrow 42\%$) &\textbf{0.309} ($\downarrow 37\%$)\\
         HMDB51&0.090  ($\downarrow 17\%$) &0.091  ($\downarrow 18\%$) &0.099  ($\downarrow 19\%$) &\textbf{0.117} ($\downarrow 14\%$)\\
    \bottomrule
    \end{tabular}}
    \label{tab:cross_dataset}
    \vspace{-1em}
\end{table}

\para{Cross-dataset performance comparison.} To investigate the generalization of our \ourmethod, we evaluate cross-dataset retrieval performance among different methods in this subsection. In detail, we train various methods on FCVID and test them on UCF101 and HMDB51. We compare the retrieval results with a single-dataset case, \ie, training and testing on the same dataset. The mAP@20 results under the cross-dataset setting are reported in Table~\ref{tab:cross_dataset}. All methods degrade in retrieval performance due to the domain gap between training and test data. Note that \ourmethod still surpasses all baseline methods, and the drop in mAP ($37\%\downarrow$ and $14\% \downarrow$) is relatively smaller than others. We credit this excellent generalization ability to the spatio-temporal contrastive learning framework. It ensures the alignment of positive samples, the divergence of class centers, and concentration of augmented data, which have been identified as three key factors contributing to the generalization ability according to recent research~\cite{Generalization2023ICLR}.

 % CHAIN&0.208&0.280&0.309&0.077&0.098&0.117\\
 
\subsection{Ablation Study}
\para{Effect of spatio-temporal augmentations.}
In \ourmethod, we introduce both spatial and temporal augmentations to collaboratively generate high-quality positive pairs for the proposed contrastive learning framework. We experiment on the augmentations to verify their effectiveness, and the results are reported in Table~\ref{tab:ablation}. We observed that contrastive learning with only spatial or temporal augmentation will result in sub-optimal performance (the first two rows vs. the third row). As a result, we conclude that both spatial and temporal information is essential for robust hash learning, and the proposed spatio-temporal augmentation strategy assists the model in focusing on the relevant spatio-temporal features.

\para{Effect of frame order verification.} We study the effect of the frame order verification (FOV) task in Table~\ref{tab:ablation}. Based on the comparison between the fourth row and the third row, we see an average improvement of $5.3\%$ and $11.7\%$ in UCF101 and ActivityNet, respectively. We believe that the FOV task provides a complementary signal to the contrastive learning task, enabling the model to better capture the temporal relationship between frames. These findings suggest that the FOV task is a valuable complement to contrastive learning in the context of spatio-temporal hashing.

\begin{table}[tbp]
    \centering
    \caption{The mAP@20 under different learning tasks. ``CL'' means contrastive learning, and ``SA'' and ``TA'' are spatial augmentation and temporal augmentation, respectively. Similarly,  ``FOV'' and ``SCR'' stand for frame order verification and scene change regularization, respectively.}
    \scalebox{0.85}{\begin{tabular}{cccccccccc}
    \toprule
    \multicolumn{2}{c}{CL}&
    \multirow{2}{*}{FOV}&
    \multirow{2}{*}{SCR}&
    \multicolumn{3}{c}{UCF01}&
    \multicolumn{3}{c}{ActivityNet}\\
    \cmidrule(lr){1-2} \cmidrule(lr){5-7} \cmidrule(lr){8-10}
          SA&TA& &&16bits&32bits&64bits&16bits&32bits&64bits\\
    \midrule
    \cmark&\xmark&\xmark&\xmark&0.351&0.389&0.440&0.079&0.119&0.142\\
    \xmark&\cmark&\xmark&\xmark&0.324&0.370&0.414&0.067&0.104&0.131\\
    \cmark&\cmark&\xmark&\xmark&0.374&0.412&0.456&0.102&0.155&0.168\\
     \cmark&\cmark&\cmark&\xmark&0.392&0.434&0.482&0.110&0.163&0.189\\
     \cmark&\cmark&\xmark&\cmark&0.389&0.432&0.483&0.115&0.169&0.195\\
    \midrule
     \cmark&\cmark&\cmark&\cmark&\textbf{0.417}&\textbf{0.444}&\textbf{0.496}&\textbf{0.125}&\textbf{0.181}&\textbf{0.205}\\
    \bottomrule
    \end{tabular}}
    \label{tab:ablation}
    \vspace{-1em}
\end{table}

\begin{table}[tbp]
    \centering
    \caption{The mAP@20 with different spatio-temporal augmentation strategies. Note that TCA denotes temporally consistent spatial augmentation, while w/o means without TCA, \ie, the randomness in  spatial augmentations are not fixed.}
    \scalebox{0.93}{
    \begin{tabular}{ccccccc}
    \toprule
    \multirow{2}{*}{Augmentations}
    &\multicolumn{3}{c}{UCF101}&
    \multicolumn{3}{c}{ActivityNet}\\
    \cmidrule(lr){2-4} \cmidrule(lr){5-7}
    &16bits&32bits&64bits&16bits&32bits&64bits\\
    \midrule
         w/o TCA& 0.414&0.438&0.491&0.122&0.174&0.197\\
         w TCA& \textbf{0.417}&\textbf{0.444}&\textbf{0.496}&\textbf{0.125}&\textbf{0.181}&\textbf{0.205}\\
         \midrule
         \midrule
         ramdom& 0.368&0.423&0.467&0.108&0.157&0.178 \\
         
         consecutive& 0.305&0.391&0.435&0.084&0.123&0.151 \\
        
         segment& \textbf{0.417}&\textbf{0.444}&\textbf{0.496}&\textbf{0.125}&\textbf{0.181}&\textbf{0.205}\\
         
    \bottomrule
    \end{tabular}
    }
    
    \label{tab:augmentation}
    \vspace{-0.5em}
    
\end{table}

\para{Effect of scene change regularization.} We also test the effect of scene change regulation (SCR) in Table~\ref{tab:ablation}. As seen a comparison between the fifth and third rows, scene change regulation focuses on frame-to-frame relationships and is an excellent complement to contrastive learning that focuses on video-to-video relationships. In addition, we find that the average improvement yielded by SCR on ActivityNet is larger than that on UCF101. We attribute it to the fact that the ActivityNet dataset contains videos captured from the internet. These videos include a diverse and rich set of real-world events and have longer duration compared to the videos in UCF101, resulting in a more prominent scene change problem.

Note that \ourmethod cannot learn hash codes by FOV or SCR tasks solely, and we do not report corresponding results. They can only assist in model learning and do not directly impose constraints on the hash codes. If used alone, the hash layer lacks gradient information and cannot be updated, resulting in hash codes that lack semantic meaning and exhibit high randomness. Although these tasks do not directly optimize the hash layer, the gradients of these tasks can be back-propagated to the video encoder, capturing rich local details and enhancing spatio-temporal modeling for better input representations to the hash layer. As a result, our spatio-temporal contrastive learning framework incorporating both FOV and SCR tasks demonstrates a substantial improvement in retrieval performance, as shown in the last row of Table~\ref{tab:ablation}.

\para{Investigation on different spatio-temporal augmentations.} To determine the fit spatio-temporal augmentation strategy, we follow~\cite{CVRL2021CVPR} to employ a temporally consistent spatial augmentation (TCA), \ie, fix the randomness of spatial augmentation (\ie, random color jitter) along temporal dimension. We evaluate hashing performance when configuring TCA in our approach. Based on a comparison between the first and second rows in Table~\ref{tab:augmentation}, we decided to use TCA in the spatial augmentation process due to the slight performance improvements. In addition, we also test the influence of different temporal augmentation strategies in Table~\ref{tab:augmentation}, including random sampling (\ie, random sample $T$ frames), consecutive sampling (\ie, sample $T$ consecutive frames), and our proposed segment-based sampling. In general, we find that the segment-based sampling strategy outperforms the others on all datasets. The use of segment-based sampling may be beneficial because it 1) covers a long range of a video, capturing more accurate global dependencies, and 2) acquires more scenes for scene change regulation, enhancing the model's discriminatory ability.

\subsection{Encoding Time} The time it takes to generate binary codes is a crucial factor in practical retrieval systems~\cite{BTH2021CVPR,MCMSH2022MM}. We count the encode time from frame-level features to the binary hash code, and record the average time for 100 videos in Table~\ref{tab:encode_time}. Note that we run BTH, MCMSH, ConMH, and \ourmethod on the same platform to ensure fairness. It is clear that \ourmethod achieves significantly better results compared to BTH while requiring only a slightly longer encoding time (0.93ms vs 0.92ms). Furthermore, \ourmethod outperforms the state-of-the-art method, ConMH, in terms of both retrieval performance and time complexity, where the time complexity of ConMH is measured using its small model consisting of 12 layers.

\begin{table}[tb]
    \centering
    \caption{Encoding time of various deep hash methods.}
    \scalebox{1.1}{
    \begin{tabular}{ccccc}
    \toprule
         Methods& BTH & MCMSH & ConMH & \ourmethod  \\
    \midrule
         Encoding time & 0.92ms&3.8ms&4.6ms&0.93ms\\ 
    \bottomrule
    \end{tabular}
    }
    \label{tab:encode_time}
    \vspace{-1.5em}
\end{table}

\subsection{Qualitative Results}
\begin{figure}[thb]
    \centering
    \includegraphics[width=\linewidth]{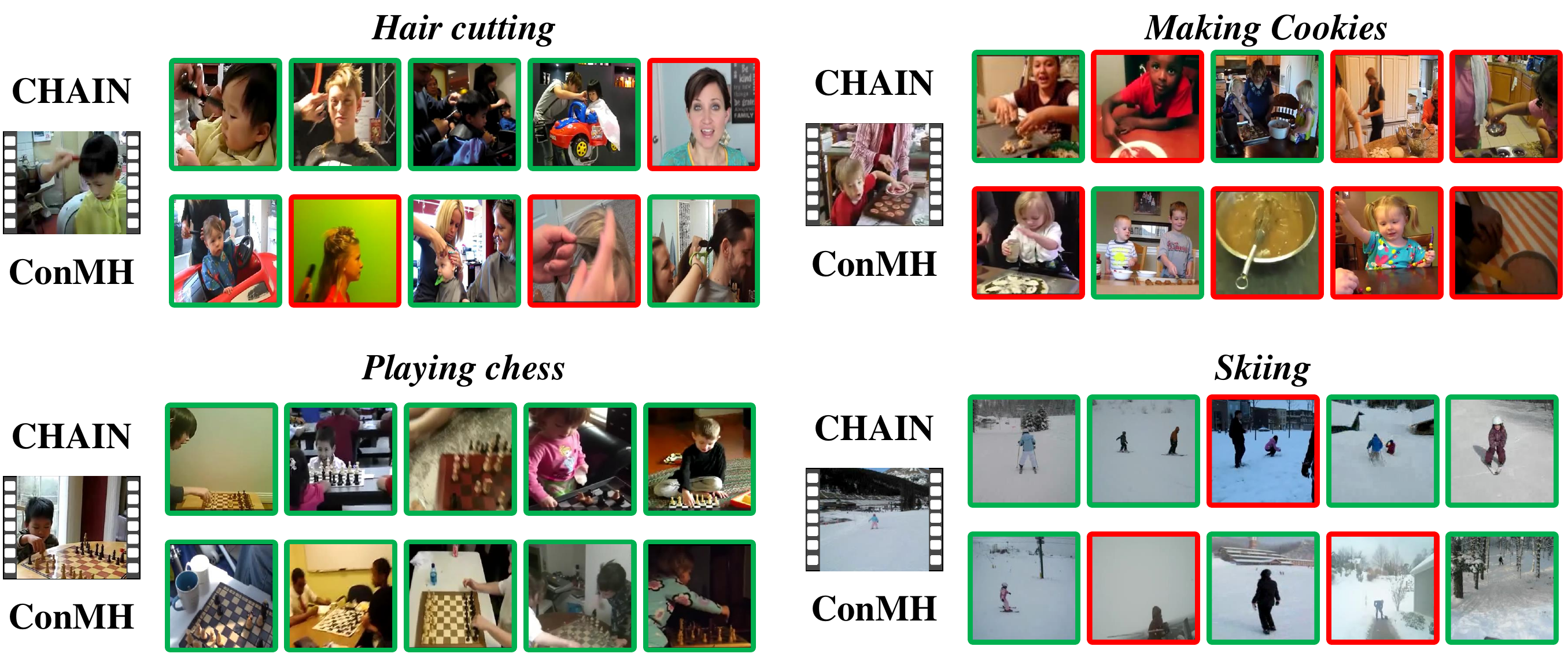}
    \caption{Top-5 retrieved results of \ourmethod and ConMH on FCVID dataset. The video inside the green square is correctly retrieved, while the video inside the red square is incorrect.}
    \label{fig:query}
\end{figure}

\para{Top-5 retrieved results visualization.} 
Figure~\ref{fig:query} illustrates the top-5 retrieved results at 64 bits on FCVID, visually comparing the performance of \ourmethod and ConMH. Following~\cite{MCMSH2022MM}, we select four video classes, \ie, ``Hair cutting'', ``Making cookies'', ``Playing chess'', and ``Skiing'', that exhibit complex human-object interactions and thus demand robust spatio-temporal modeling. Although both methods can provide relevant candidate videos, \ourmethod exhibits a more stable performance in retrieving more relevant videos. For example, when given query videos from categories ``Hair cutting'' and ``Making cookies'', \ourmethod achieves a top-5 precision of $80\%$ and $40\%$, respectively, compared to ConMH's precision of $60\%$ and $20\%$. This indicates that \ourmethod is able to more effectively leverage the rich spatial and temporal information for hash learning.

\begin{figure}[thb]
    \centering
   \subfloat[ConMH 64 bits]{\includegraphics[width=0.31\linewidth]{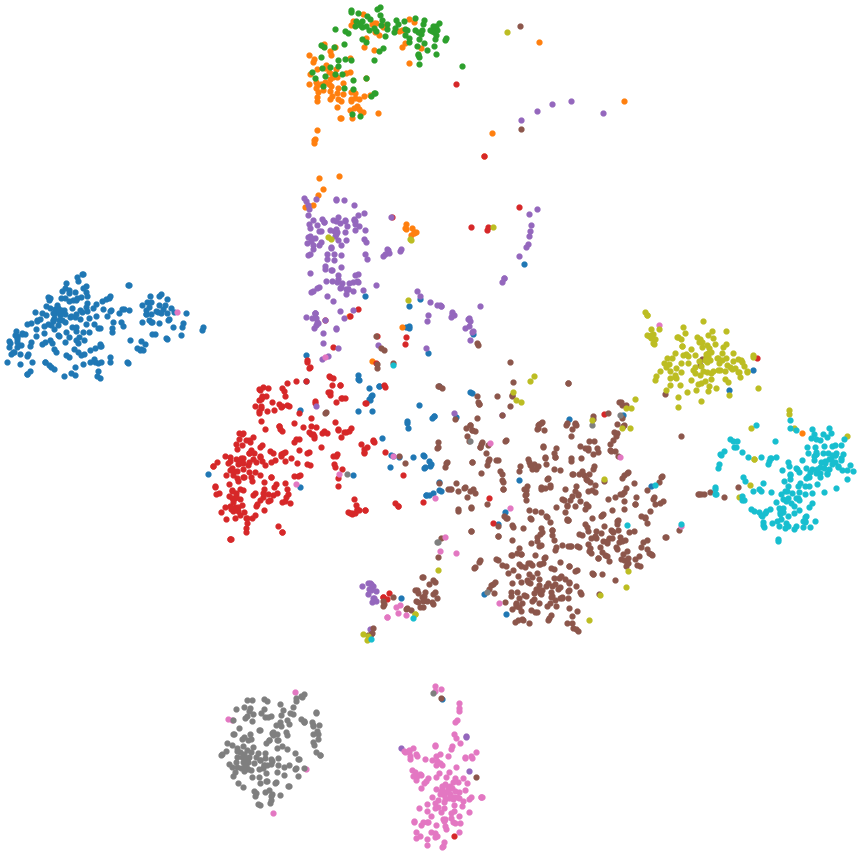}}
   \hfill
  \subfloat[ConMH 32 bits] {\includegraphics[width=0.31\linewidth]{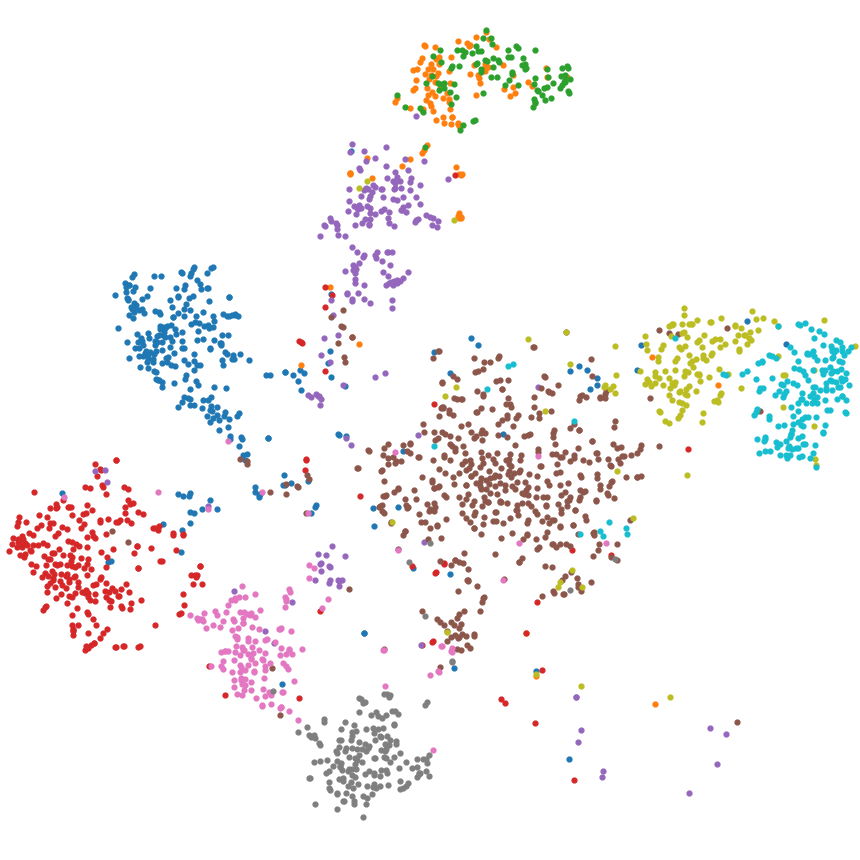}}
  \hfill
  \subfloat[ConMH 16 bits] {\includegraphics[width=0.31\linewidth]{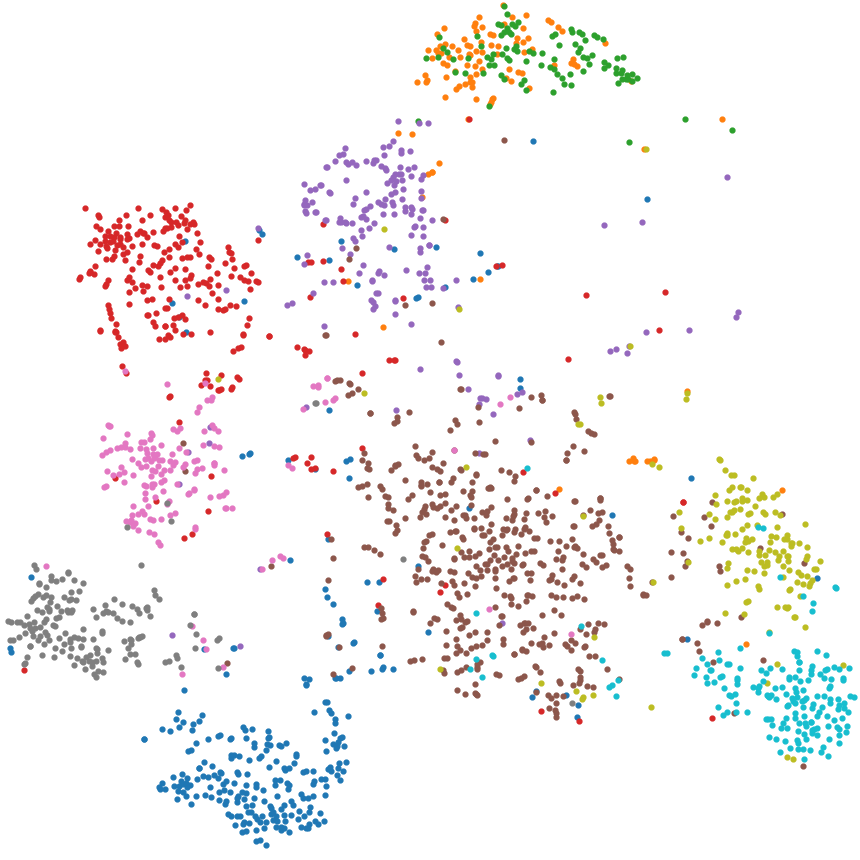}}
  \\
   \subfloat[\ourmethod 64 bits]{\includegraphics[width=0.31\linewidth]{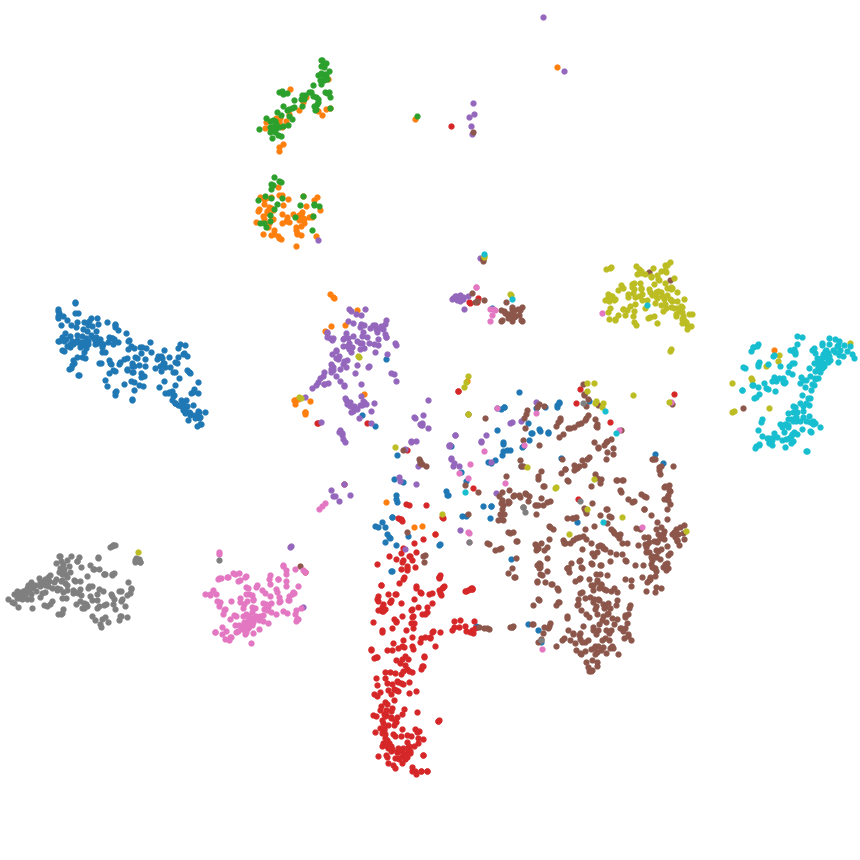}}
   \hfill
\subfloat[\ourmethod 32 bits] {\includegraphics[width=0.31\linewidth]{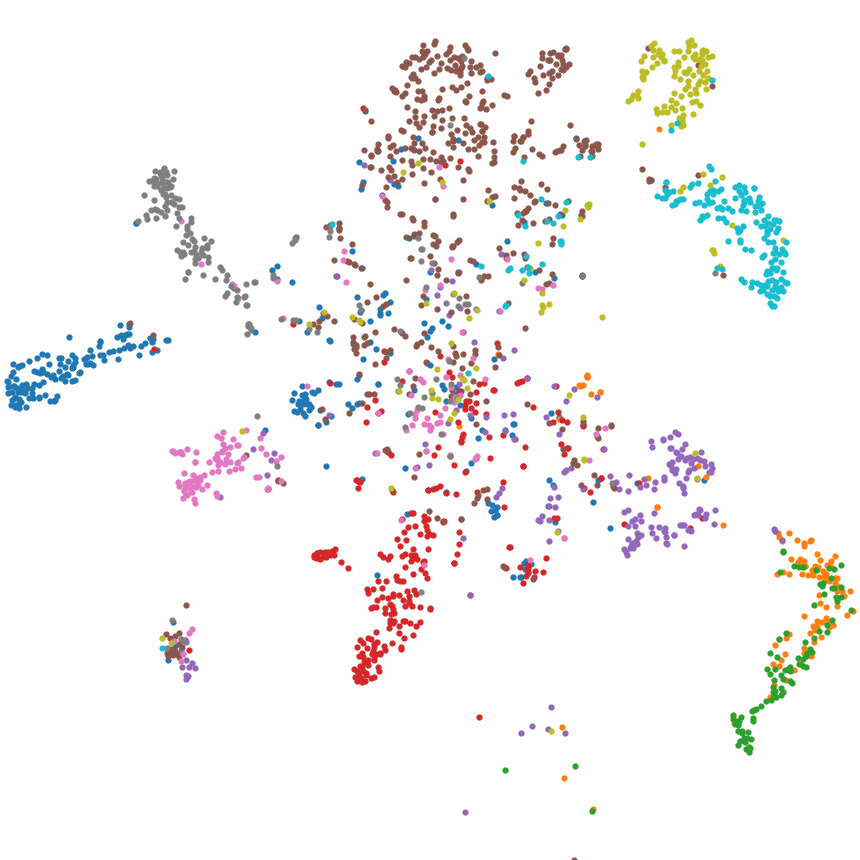}}
\hfill
  \subfloat[\ourmethod 16 bits] {\includegraphics[width=0.31\linewidth]{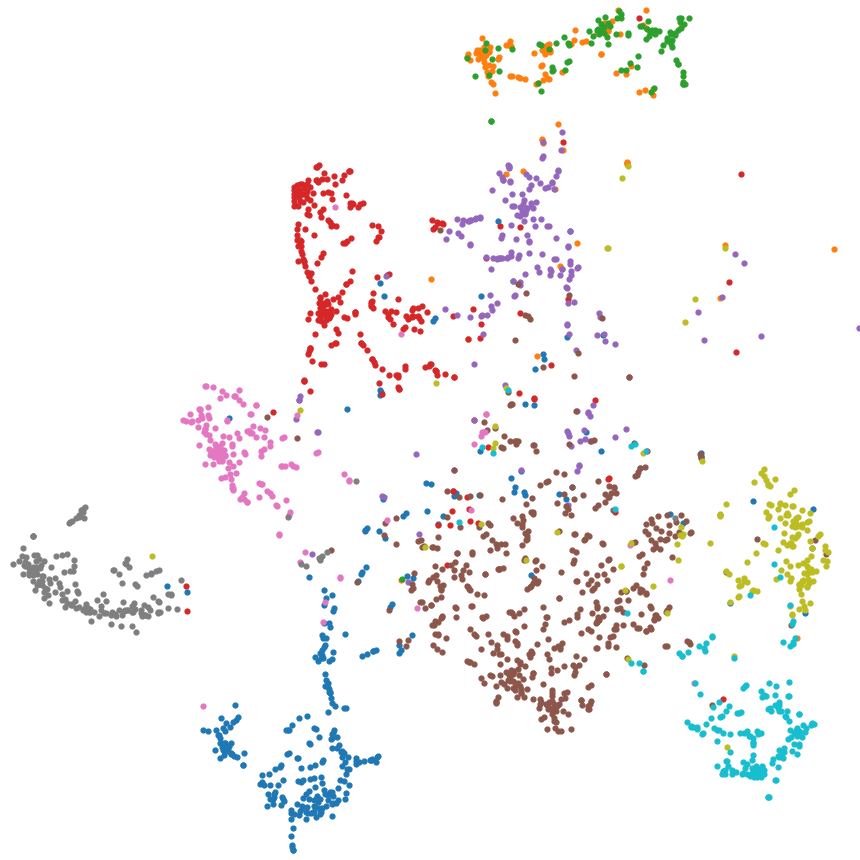}}
  
    \caption{ The t-SNE visualization of the learned 64-bit hash codes from the test set of FCVID. The scattered points of the same color indicate the same category. Note that we only visualize the first 10 classes.}
    \label{fig:t_sne}
    \vspace{-0.5em}
\end{figure}

\para{The t-SNE visualization of hash codes.}
To better understand the results, we use t-SNE~\cite{tsne2008JMLR} to visualize the learned hash codes on the evaluation set of FCVID. As illustrated in Figure~\ref{fig:t_sne}, we observed that the hash codes generated by \ourmethod exhibit more distinct compactness for the same category and dispersion for dissimilar ones, in comparison to ConMH. This suggests that \ourmethod can produce more discriminative binary codes, thereby offering remarkable retrieval performance.

% \section{Limitations and Future Work}
% Although our proposed \ourmethod achieves remarkable retrieval performance, there are still some limitations that need to be solved in our future work.
% \begin{itemize}
%     \item \textbf{Under-exploration of more advanced backbones.} Although we use VGG/ResNet as the spatial encoder and a transformer encoder to model temporal dynamics for fair comparisons, it falls short in dealing with real-world long videos. We plan to extend our method to more recent backbones such as TimeSformer~\cite{TimesFormer2021ICML} and TubeViT~\cite{TubeViT2023arxiv}.

%     \item \textbf{High-cost in model deployment.} Although the time complexity of \ourmethod is in line with the baselines, its deployment in real-time applications remains challenging. We aim to explore computationally efficient models and compress the current model to better meet the needs of scenarios with high real-time requirements.
    
%     \item \textbf{Expensive model updating.} The over-time changes in the distribution of real-world video data will degrade the retrieval performance. To address these issues, we aim to explore more effective updating strategies that can reduce the cost of re-training the model and refreshing hashing indexes.
% \end{itemize}

\section{Conclusion}

In this paper, we propose \ourmethod, a novel self-supervised video hashing algorithm that fully exploits both global and local spatio-temporal cues to achieve robust hash learning. On the one hand, we adopt a novel spatio-temporal contrastive learning framework to learn view-invariant hash codes. On the other hand, we incorporate two collaborative tasks, namely frame order verification and scene change regulation, to further enhance the model's spatio-temporal modeling capability. Experiments on four benchmarks demonstrate that our proposed \ourmethod offers remarkable improvements compared with SOTA methods in terms of retrieval performance. \ourmethod can be productively integrated into multimedia similarity search engines to support computation-efficient and storage-friendly video retrieval.

\begin{acks}
This work was achieved in Key Laboratory of Information Storage System and Ministry of Education of China. It was supported by the National Natural Science Foundation of China No.61902135, the National Natural Science Foundation of China (key program) No.62232007, 
and the Natural Science Foundation of Hubei Province No.2022CFB060.
\end{acks}


\newpage

\bibliographystyle{ACM-Reference-Format}
\balance
\bibliography{reference}


\begin{thebibliography}{64}

%%% the Web address.

\ifx \showCODEN    \undefined \def \showCODEN     #1{\unskip}     \fi
\ifx \showDOI      \undefined \def \showDOI       #1{#1}\fi
\ifx \showISBNx    \undefined \def \showISBNx     #1{\unskip}     \fi
\ifx \showISBNxiii \undefined \def \showISBNxiii  #1{\unskip}     \fi
\ifx \showISSN     \undefined \def \showISSN      #1{\unskip}     \fi
\ifx \showLCCN     \undefined \def \showLCCN      #1{\unskip}     \fi
\ifx \shownote     \undefined \def \shownote      #1{#1}          \fi
\ifx \showarticletitle \undefined \def \showarticletitle #1{#1}   \fi
\ifx \showURL      \undefined \def \showURL       {\relax}        \fi
% invisible to TeX
\providecommand\bibfield[2]{#2}
\providecommand\bibinfo[2]{#2}
\providecommand\natexlab[1]{#1}
\providecommand\showeprint[2][]{arXiv:#2}

\bibitem[\protect\citeauthoryear{Baldrati, Bertini, Uricchio, and
  Bimbo}{Baldrati et~al\mbox{.}}{2022}]%
        {CLIP_retrieval2022CVPR}
\bibfield{author}{\bibinfo{person}{Alberto Baldrati}, \bibinfo{person}{Marco
  Bertini}, \bibinfo{person}{Tiberio Uricchio}, {and}
  \bibinfo{person}{Alberto~Del Bimbo}.} \bibinfo{year}{2022}\natexlab{}.
\newblock \showarticletitle{Effective conditioned and composed image retrieval
  combining CLIP-based features}. In \bibinfo{booktitle}{\emph{{IEEE/CVF}
  Conference on Computer Vision and Pattern Recognition, {CVPR} 2022, New
  Orleans, LA, USA, June 18-24, 2022}}. \bibinfo{publisher}{{IEEE}},
  \bibinfo{pages}{21434--21442}.
\newblock


\bibitem[\protect\citeauthoryear{Bertasius, Wang, and Torresani}{Bertasius
  et~al\mbox{.}}{2021}]%
        {TimesFormer2021ICML}
\bibfield{author}{\bibinfo{person}{Gedas Bertasius}, \bibinfo{person}{Heng
  Wang}, {and} \bibinfo{person}{Lorenzo Torresani}.}
  \bibinfo{year}{2021}\natexlab{}.
\newblock \showarticletitle{Is Space-Time Attention All You Need for Video
  Understanding?}. In \bibinfo{booktitle}{\emph{Proceedings of the 38th
  International Conference on Machine Learning, {ICML} 2021, 18-24 July 2021,
  Virtual Event}} \emph{(\bibinfo{series}{Proceedings of Machine Learning
  Research}, Vol.~\bibinfo{volume}{139})},
  \bibfield{editor}{\bibinfo{person}{Marina Meila} {and} \bibinfo{person}{Tong
  Zhang}} (Eds.). \bibinfo{publisher}{{PMLR}}, \bibinfo{pages}{813--824}.
\newblock


\bibitem[\protect\citeauthoryear{Chen, Kornblith, Norouzi, and Hinton}{Chen
  et~al\mbox{.}}{2020}]%
        {SimCLR2020ICML}
\bibfield{author}{\bibinfo{person}{Ting Chen}, \bibinfo{person}{Simon
  Kornblith}, \bibinfo{person}{Mohammad Norouzi}, {and}
  \bibinfo{person}{Geoffrey~E. Hinton}.} \bibinfo{year}{2020}\natexlab{}.
\newblock \showarticletitle{A Simple Framework for Contrastive Learning of
  Visual Representations}. In \bibinfo{booktitle}{\emph{Proceedings of the 37th
  International Conference on Machine Learning, {ICML} 2020, 13-18 July 2020,
  Virtual Event}} \emph{(\bibinfo{series}{Proceedings of Machine Learning
  Research}, Vol.~\bibinfo{volume}{119})}. \bibinfo{publisher}{{PMLR}},
  \bibinfo{pages}{1597--1607}.
\newblock


\bibitem[\protect\citeauthoryear{Chen and He}{Chen and He}{2021}]%
        {SimSiam2021CVPR}
\bibfield{author}{\bibinfo{person}{Xinlei Chen} {and} \bibinfo{person}{Kaiming
  He}.} \bibinfo{year}{2021}\natexlab{}.
\newblock \showarticletitle{Exploring Simple Siamese Representation Learning}.
  In \bibinfo{booktitle}{\emph{{IEEE} Conference on Computer Vision and Pattern
  Recognition, {CVPR} 2021, virtual, June 19-25, 2021}}.
  \bibinfo{publisher}{Computer Vision Foundation / {IEEE}},
  \bibinfo{pages}{15750--15758}.
\newblock


\bibitem[\protect\citeauthoryear{Courbariaux, Bengio, and David}{Courbariaux
  et~al\mbox{.}}{2015}]%
        {STE2015NIPS}
\bibfield{author}{\bibinfo{person}{Matthieu Courbariaux},
  \bibinfo{person}{Yoshua Bengio}, {and} \bibinfo{person}{Jean{-}Pierre
  David}.} \bibinfo{year}{2015}\natexlab{}.
\newblock \showarticletitle{BinaryConnect: Training Deep Neural Networks with
  binary weights during propagations}. In \bibinfo{booktitle}{\emph{Advances in
  Neural Information Processing Systems 28: Annual Conference on Neural
  Information Processing Systems 2015, December 7-12, 2015, Montreal, Quebec,
  Canada}}, \bibfield{editor}{\bibinfo{person}{Corinna Cortes},
  \bibinfo{person}{Neil~D. Lawrence}, \bibinfo{person}{Daniel~D. Lee},
  \bibinfo{person}{Masashi Sugiyama}, {and} \bibinfo{person}{Roman Garnett}}
  (Eds.). \bibinfo{pages}{3123--3131}.
\newblock


\bibitem[\protect\citeauthoryear{Cui, Zhu, Li, Zhang, and Guan}{Cui
  et~al\mbox{.}}{2022}]%
        {WSIH2022MM}
\bibfield{author}{\bibinfo{person}{Hui Cui}, \bibinfo{person}{Lei Zhu},
  \bibinfo{person}{Jingjing Li}, \bibinfo{person}{Zheng Zhang}, {and}
  \bibinfo{person}{Weili Guan}.} \bibinfo{year}{2022}\natexlab{}.
\newblock \showarticletitle{Webly Supervised Image Hashing with Lightweight
  Semantic Transfer Network}. In \bibinfo{booktitle}{\emph{Proceedings of the
  30th ACM International Conference on Multimedia}} (Lisboa, Portugal)
  \emph{(\bibinfo{series}{MM '22})}. \bibinfo{publisher}{Association for
  Computing Machinery}, \bibinfo{address}{New York, NY, USA},
  \bibinfo{pages}{3451–3460}.
\newblock
\showISBNx{9781450392037}


\bibitem[\protect\citeauthoryear{Deng, Dong, Socher, Li, Li, and
  Fei{-}Fei}{Deng et~al\mbox{.}}{2009}]%
        {ImageNet2009CVPR}
\bibfield{author}{\bibinfo{person}{Jia Deng}, \bibinfo{person}{Wei Dong},
  \bibinfo{person}{Richard Socher}, \bibinfo{person}{Li{-}Jia Li},
  \bibinfo{person}{Kai Li}, {and} \bibinfo{person}{Li Fei{-}Fei}.}
  \bibinfo{year}{2009}\natexlab{}.
\newblock \showarticletitle{ImageNet: {A} large-scale hierarchical image
  database}. In \bibinfo{booktitle}{\emph{2009 {IEEE} Computer Society
  Conference on Computer Vision and Pattern Recognition {(CVPR} 2009), 20-25
  June 2009, Miami, Florida, {USA}}}. \bibinfo{publisher}{{IEEE} Computer
  Society}, \bibinfo{pages}{248--255}.
\newblock


\bibitem[\protect\citeauthoryear{Devlin, Chang, Lee, and Toutanova}{Devlin
  et~al\mbox{.}}{2019}]%
        {BERT2019NAACL}
\bibfield{author}{\bibinfo{person}{Jacob Devlin}, \bibinfo{person}{Ming{-}Wei
  Chang}, \bibinfo{person}{Kenton Lee}, {and} \bibinfo{person}{Kristina
  Toutanova}.} \bibinfo{year}{2019}\natexlab{}.
\newblock \showarticletitle{{BERT:} Pre-training of Deep Bidirectional
  Transformers for Language Understanding}. In
  \bibinfo{booktitle}{\emph{Proceedings of the 2019 Conference of the North
  American Chapter of the Association for Computational Linguistics: Human
  Language Technologies, {NAACL-HLT} 2019, Minneapolis, MN, USA, June 2-7,
  2019, Volume 1 (Long and Short Papers)}}. \bibinfo{publisher}{Association for
  Computational Linguistics}, \bibinfo{pages}{4171--4186}.
\newblock


\bibitem[\protect\citeauthoryear{Dong, Chen, Zhang, Yang, Chen, Li, and
  Wang}{Dong et~al\mbox{.}}{2022a}]%
        {dong2022partially}
\bibfield{author}{\bibinfo{person}{Jianfeng Dong}, \bibinfo{person}{Xianke
  Chen}, \bibinfo{person}{Minsong Zhang}, \bibinfo{person}{Xun Yang},
  \bibinfo{person}{Shujie Chen}, \bibinfo{person}{Xirong Li}, {and}
  \bibinfo{person}{Xun Wang}.} \bibinfo{year}{2022}\natexlab{a}.
\newblock \showarticletitle{Partially Relevant Video Retrieval}. In
  \bibinfo{booktitle}{\emph{Proceedings of the 30th ACM International
  Conference on Multimedia}}. \bibinfo{pages}{246--257}.
\newblock


\bibitem[\protect\citeauthoryear{Dong, Li, Xu, Yang, Yang, Wang, and Wang}{Dong
  et~al\mbox{.}}{2022b}]%
        {dong2022dual}
\bibfield{author}{\bibinfo{person}{Jianfeng Dong}, \bibinfo{person}{Xirong Li},
  \bibinfo{person}{Chaoxi Xu}, \bibinfo{person}{Xun Yang},
  \bibinfo{person}{Gang Yang}, \bibinfo{person}{Xun Wang}, {and}
  \bibinfo{person}{Meng Wang}.} \bibinfo{year}{2022}\natexlab{b}.
\newblock \showarticletitle{Dual encoding for video retrieval by text}.
\newblock \bibinfo{journal}{\emph{IEEE Transactions on Pattern Analysis and
  Machine Intelligence}} \bibinfo{volume}{44}, \bibinfo{number}{8}
  (\bibinfo{year}{2022}), \bibinfo{pages}{4065--4080}.
\newblock


\bibitem[\protect\citeauthoryear{Dosovitskiy, Beyer, Kolesnikov, Weissenborn,
  Zhai, Unterthiner, Dehghani, Minderer, Heigold, Gelly, Uszkoreit, and
  Houlsby}{Dosovitskiy et~al\mbox{.}}{2021}]%
        {ViT2021ICLR}
\bibfield{author}{\bibinfo{person}{Alexey Dosovitskiy}, \bibinfo{person}{Lucas
  Beyer}, \bibinfo{person}{Alexander Kolesnikov}, \bibinfo{person}{Dirk
  Weissenborn}, \bibinfo{person}{Xiaohua Zhai}, \bibinfo{person}{Thomas
  Unterthiner}, \bibinfo{person}{Mostafa Dehghani}, \bibinfo{person}{Matthias
  Minderer}, \bibinfo{person}{Georg Heigold}, \bibinfo{person}{Sylvain Gelly},
  \bibinfo{person}{Jakob Uszkoreit}, {and} \bibinfo{person}{Neil Houlsby}.}
  \bibinfo{year}{2021}\natexlab{}.
\newblock \showarticletitle{An Image is Worth 16x16 Words: Transformers for
  Image Recognition at Scale}. In \bibinfo{booktitle}{\emph{9th International
  Conference on Learning Representations, {ICLR} 2021, Virtual Event, Austria,
  May 3-7, 2021}}. \bibinfo{publisher}{OpenReview.net}.
\newblock


\bibitem[\protect\citeauthoryear{Frey and Dueck}{Frey and Dueck}{2007}]%
        {APClustering}
\bibfield{author}{\bibinfo{person}{Brendan~J Frey} {and}
  \bibinfo{person}{Delbert Dueck}.} \bibinfo{year}{2007}\natexlab{}.
\newblock \showarticletitle{Clustering by passing messages between data
  points}.
\newblock \bibinfo{journal}{\emph{science}} \bibinfo{volume}{315},
  \bibinfo{number}{5814} (\bibinfo{year}{2007}), \bibinfo{pages}{972--976}.
\newblock


\bibitem[\protect\citeauthoryear{Gong and Lazebnik}{Gong and Lazebnik}{2011}]%
        {ITQ2011CVPR}
\bibfield{author}{\bibinfo{person}{Yunchao Gong} {and}
  \bibinfo{person}{Svetlana Lazebnik}.} \bibinfo{year}{2011}\natexlab{}.
\newblock \showarticletitle{Iterative quantization: {A} procrustean approach to
  learning binary codes}. In \bibinfo{booktitle}{\emph{The 24th {IEEE}
  Conference on Computer Vision and Pattern Recognition, {CVPR} 2011, Colorado
  Springs, CO, USA, 20-25 June 2011}}. \bibinfo{publisher}{{IEEE} Computer
  Society}, \bibinfo{pages}{817--824}.
\newblock


\bibitem[\protect\citeauthoryear{Grill, Strub, Altch{\'{e}}, Tallec, Richemond,
  Buchatskaya, Doersch, Pires, Guo, Azar, Piot, Kavukcuoglu, Munos, and
  Valko}{Grill et~al\mbox{.}}{2020}]%
        {BYOL2020NIPS}
\bibfield{author}{\bibinfo{person}{Jean{-}Bastien Grill},
  \bibinfo{person}{Florian Strub}, \bibinfo{person}{Florent Altch{\'{e}}},
  \bibinfo{person}{Corentin Tallec}, \bibinfo{person}{Pierre~H. Richemond},
  \bibinfo{person}{Elena Buchatskaya}, \bibinfo{person}{Carl Doersch},
  \bibinfo{person}{Bernardo~{\'{A}}vila Pires}, \bibinfo{person}{Zhaohan Guo},
  \bibinfo{person}{Mohammad~Gheshlaghi Azar}, \bibinfo{person}{Bilal Piot},
  \bibinfo{person}{Koray Kavukcuoglu}, \bibinfo{person}{R{\'{e}}mi Munos},
  {and} \bibinfo{person}{Michal Valko}.} \bibinfo{year}{2020}\natexlab{}.
\newblock \showarticletitle{Bootstrap Your Own Latent - {A} New Approach to
  Self-Supervised Learning}.
\newblock  (\bibinfo{year}{2020}).
\newblock


\bibitem[\protect\citeauthoryear{Hao, Duan, Zhang, Zhu, Zhou, and He}{Hao
  et~al\mbox{.}}{2022}]%
        {MCMSH2022MM}
\bibfield{author}{\bibinfo{person}{Yanbin Hao}, \bibinfo{person}{Jingru Duan},
  \bibinfo{person}{Hao Zhang}, \bibinfo{person}{Bin Zhu},
  \bibinfo{person}{Pengyuan Zhou}, {and} \bibinfo{person}{Xiangnan He}.}
  \bibinfo{year}{2022}\natexlab{}.
\newblock \showarticletitle{Unsupervised Video Hashing with Multi-granularity
  Contextualization and Multi-structure Preservation}. In
  \bibinfo{booktitle}{\emph{{MM} '22: The 30th {ACM} International Conference
  on Multimedia, Lisboa, Portugal, October 10 - 14, 2022}}.
  \bibinfo{publisher}{{ACM}}, \bibinfo{pages}{3754--3763}.
\newblock


\bibitem[\protect\citeauthoryear{He, Fan, Wu, Xie, and Girshick}{He
  et~al\mbox{.}}{2020}]%
        {MoCo2020CVPR}
\bibfield{author}{\bibinfo{person}{Kaiming He}, \bibinfo{person}{Haoqi Fan},
  \bibinfo{person}{Yuxin Wu}, \bibinfo{person}{Saining Xie}, {and}
  \bibinfo{person}{Ross~B. Girshick}.} \bibinfo{year}{2020}\natexlab{}.
\newblock \showarticletitle{Momentum Contrast for Unsupervised Visual
  Representation Learning}. In \bibinfo{booktitle}{\emph{2020 {IEEE/CVF}
  Conference on Computer Vision and Pattern Recognition, {CVPR} 2020, Seattle,
  WA, USA, June 13-19, 2020}}. \bibinfo{publisher}{Computer Vision Foundation /
  {IEEE}}, \bibinfo{pages}{9726--9735}.
\newblock


\bibitem[\protect\citeauthoryear{He, Zhang, Ren, and Sun}{He
  et~al\mbox{.}}{2016}]%
        {ResNet2016CVPR}
\bibfield{author}{\bibinfo{person}{Kaiming He}, \bibinfo{person}{Xiangyu
  Zhang}, \bibinfo{person}{Shaoqing Ren}, {and} \bibinfo{person}{Jian Sun}.}
  \bibinfo{year}{2016}\natexlab{}.
\newblock \showarticletitle{Deep Residual Learning for Image Recognition}. In
  \bibinfo{booktitle}{\emph{2016 {IEEE} Conference on Computer Vision and
  Pattern Recognition, {CVPR} 2016, Las Vegas, NV, USA, June 27-30, 2016}}.
  \bibinfo{publisher}{{IEEE} Computer Society}, \bibinfo{pages}{770--778}.
\newblock


\bibitem[\protect\citeauthoryear{Heilbron, Escorcia, Ghanem, and
  Niebles}{Heilbron et~al\mbox{.}}{2015}]%
        {ActivityNet2015CVPR}
\bibfield{author}{\bibinfo{person}{Fabian~Caba Heilbron},
  \bibinfo{person}{Victor Escorcia}, \bibinfo{person}{Bernard Ghanem}, {and}
  \bibinfo{person}{Juan~Carlos Niebles}.} \bibinfo{year}{2015}\natexlab{}.
\newblock \showarticletitle{ActivityNet: {A} large-scale video benchmark for
  human activity understanding}. In \bibinfo{booktitle}{\emph{{IEEE} Conference
  on Computer Vision and Pattern Recognition, {CVPR} 2015, Boston, MA, USA,
  June 7-12, 2015}}. \bibinfo{publisher}{{IEEE} Computer Society},
  \bibinfo{pages}{961--970}.
\newblock


\bibitem[\protect\citeauthoryear{Hochreiter and Schmidhuber}{Hochreiter and
  Schmidhuber}{1997a}]%
        {LSTM1997NECO}
\bibfield{author}{\bibinfo{person}{Sepp Hochreiter} {and}
  \bibinfo{person}{J{\"{u}}rgen Schmidhuber}.}
  \bibinfo{year}{1997}\natexlab{a}.
\newblock \showarticletitle{Long Short-Term Memory}.
\newblock \bibinfo{journal}{\emph{Neural Comput.}} \bibinfo{volume}{9},
  \bibinfo{number}{8} (\bibinfo{year}{1997}), \bibinfo{pages}{1735--1780}.
\newblock
\urldef\tempurl%
\url{https://doi.org/10.1162/neco.1997.9.8.1735}
\showDOI{\tempurl}


\bibitem[\protect\citeauthoryear{Hochreiter and Schmidhuber}{Hochreiter and
  Schmidhuber}{1997b}]%
        {LSTM1997NatCommun}
\bibfield{author}{\bibinfo{person}{Sepp Hochreiter} {and}
  \bibinfo{person}{J{\"{u}}rgen Schmidhuber}.}
  \bibinfo{year}{1997}\natexlab{b}.
\newblock \showarticletitle{Long Short-Term Memory}.
\newblock \bibinfo{journal}{\emph{Neural Comput.}} \bibinfo{volume}{9},
  \bibinfo{number}{8} (\bibinfo{year}{1997}), \bibinfo{pages}{1735--1780}.
\newblock


\bibitem[\protect\citeauthoryear{Huang, Yi, Zhao, and Jiang}{Huang
  et~al\mbox{.}}{2023}]%
        {Generalization2023ICLR}
\bibfield{author}{\bibinfo{person}{Weiran Huang}, \bibinfo{person}{Mingyang
  Yi}, \bibinfo{person}{Xuyang Zhao}, {and} \bibinfo{person}{Zihao Jiang}.}
  \bibinfo{year}{2023}\natexlab{}.
\newblock \showarticletitle{Towards the Generalization of Contrastive
  Self-Supervised Learning}. In \bibinfo{booktitle}{\emph{The Eleventh
  International Conference on Learning Representations, {ICLR} 2023}}.
\newblock


\bibitem[\protect\citeauthoryear{Jia, Tang, Chen, Cardie, Belongie, Hariharan,
  and Lim}{Jia et~al\mbox{.}}{2022}]%
        {Visual_prompt2022ECCV}
\bibfield{author}{\bibinfo{person}{Menglin Jia}, \bibinfo{person}{Luming Tang},
  \bibinfo{person}{Bor{-}Chun Chen}, \bibinfo{person}{Claire Cardie},
  \bibinfo{person}{Serge~J. Belongie}, \bibinfo{person}{Bharath Hariharan},
  {and} \bibinfo{person}{Ser{-}Nam Lim}.} \bibinfo{year}{2022}\natexlab{}.
\newblock \showarticletitle{Visual Prompt Tuning}. In
  \bibinfo{booktitle}{\emph{Computer Vision - {ECCV} 2022 - 17th European
  Conference, Tel Aviv, Israel, October 23-27, 2022, Proceedings, Part
  {XXXIII}}} \emph{(\bibinfo{series}{Lecture Notes in Computer Science},
  Vol.~\bibinfo{volume}{13693})}, \bibfield{editor}{\bibinfo{person}{Shai
  Avidan}, \bibinfo{person}{Gabriel~J. Brostow}, \bibinfo{person}{Moustapha
  Ciss{\'{e}}}, \bibinfo{person}{Giovanni~Maria Farinella}, {and}
  \bibinfo{person}{Tal Hassner}} (Eds.). \bibinfo{publisher}{Springer},
  \bibinfo{pages}{709--727}.
\newblock


\bibitem[\protect\citeauthoryear{Jiang, Wu, Wang, Xue, and Chang}{Jiang
  et~al\mbox{.}}{2018}]%
        {FCVID2018TPAMI}
\bibfield{author}{\bibinfo{person}{Yu-Gang Jiang}, \bibinfo{person}{Zuxuan Wu},
  \bibinfo{person}{Jun Wang}, \bibinfo{person}{Xiangyang Xue}, {and}
  \bibinfo{person}{Shih-Fu Chang}.} \bibinfo{year}{2018}\natexlab{}.
\newblock \showarticletitle{Exploiting Feature and Class Relationships in Video
  Categorization with Regularized Deep Neural Networks}.
\newblock \bibinfo{journal}{\emph{{IEEE} Transactions on Pattern Analysis and
  Machine Intelligence}} \bibinfo{volume}{40}, \bibinfo{number}{2}
  (\bibinfo{year}{2018}), \bibinfo{pages}{352--364}.
\newblock


\bibitem[\protect\citeauthoryear{Kingma and Ba}{Kingma and Ba}{2015}]%
        {Adam2015ICLR}
\bibfield{author}{\bibinfo{person}{Diederik~P. Kingma} {and}
  \bibinfo{person}{Jimmy Ba}.} \bibinfo{year}{2015}\natexlab{}.
\newblock \showarticletitle{Adam: {A} Method for Stochastic Optimization}. In
  \bibinfo{booktitle}{\emph{3rd International Conference on Learning
  Representations, {ICLR} 2015, San Diego, CA, USA, May 7-9, 2015, Conference
  Track Proceedings}}, \bibfield{editor}{\bibinfo{person}{Yoshua Bengio} {and}
  \bibinfo{person}{Yann LeCun}} (Eds.).
\newblock


\bibitem[\protect\citeauthoryear{Kuang, Zhu, Zhang, Li, Tighe, Schwertfeger,
  Stachniss, and Li}{Kuang et~al\mbox{.}}{2021}]%
        {VCLR2021ICCVW}
\bibfield{author}{\bibinfo{person}{Haofei Kuang}, \bibinfo{person}{Yi Zhu},
  \bibinfo{person}{Zhi Zhang}, \bibinfo{person}{Xinyu Li},
  \bibinfo{person}{Joseph Tighe}, \bibinfo{person}{S{\"{o}}ren Schwertfeger},
  \bibinfo{person}{Cyrill Stachniss}, {and} \bibinfo{person}{Mu Li}.}
  \bibinfo{year}{2021}\natexlab{}.
\newblock \showarticletitle{Video Contrastive Learning with Global Context}. In
  \bibinfo{booktitle}{\emph{{IEEE/CVF} International Conference on Computer
  Vision Workshops, {ICCVW} 2021, Montreal, BC, Canada, October 11-17, 2021}}.
  \bibinfo{publisher}{{IEEE}}, \bibinfo{pages}{3188}.
\newblock


\bibitem[\protect\citeauthoryear{Kuehne, Jhuang, Garrote, Poggio, and
  Serre}{Kuehne et~al\mbox{.}}{2011}]%
        {HMDB2011ICCV}
\bibfield{author}{\bibinfo{person}{Hildegard Kuehne}, \bibinfo{person}{Hueihan
  Jhuang}, \bibinfo{person}{Est{\'{\i}}baliz Garrote},
  \bibinfo{person}{Tomaso~A. Poggio}, {and} \bibinfo{person}{Thomas Serre}.}
  \bibinfo{year}{2011}\natexlab{}.
\newblock \showarticletitle{{HMDB:} {A} large video database for human motion
  recognition}. In \bibinfo{booktitle}{\emph{{IEEE} International Conference on
  Computer Vision, {ICCV} 2011, Barcelona, Spain, November 6-13, 2011}},
  \bibfield{editor}{\bibinfo{person}{Dimitris~N. Metaxas},
  \bibinfo{person}{Long Quan}, \bibinfo{person}{Alberto Sanfeliu}, {and}
  \bibinfo{person}{Luc~Van Gool}} (Eds.). \bibinfo{publisher}{{IEEE} Computer
  Society}, \bibinfo{pages}{2556--2563}.
\newblock


\bibitem[\protect\citeauthoryear{Kwon and Ye}{Kwon and Ye}{2022}]%
        {CLIP_tran2022CVPR}
\bibfield{author}{\bibinfo{person}{Gihyun Kwon} {and}
  \bibinfo{person}{Jong~Chul Ye}.} \bibinfo{year}{2022}\natexlab{}.
\newblock \showarticletitle{CLIPstyler: Image Style Transfer with a Single Text
  Condition}. In \bibinfo{booktitle}{\emph{{IEEE/CVF} Conference on Computer
  Vision and Pattern Recognition, {CVPR} 2022, New Orleans, LA, USA, June
  18-24, 2022}}. \bibinfo{publisher}{{IEEE}}, \bibinfo{pages}{18041--18050}.
\newblock


\bibitem[\protect\citeauthoryear{Lee, Huang, Singh, and Yang}{Lee
  et~al\mbox{.}}{2017}]%
        {VideoOrder2017ICCV}
\bibfield{author}{\bibinfo{person}{Hsin{-}Ying Lee}, \bibinfo{person}{Jia{-}Bin
  Huang}, \bibinfo{person}{Maneesh Singh}, {and} \bibinfo{person}{Ming{-}Hsuan
  Yang}.} \bibinfo{year}{2017}\natexlab{}.
\newblock \showarticletitle{Unsupervised Representation Learning by Sorting
  Sequences}. In \bibinfo{booktitle}{\emph{{IEEE} International Conference on
  Computer Vision, {ICCV} 2017, Venice, Italy, October 22-29, 2017}}.
  \bibinfo{publisher}{{IEEE} Computer Society}, \bibinfo{pages}{667--676}.
\newblock


\bibitem[\protect\citeauthoryear{Lee, Park, Kim, Kim, Boboev, and Baek}{Lee
  et~al\mbox{.}}{2023}]%
        {CLIP_3D2023WACV}
\bibfield{author}{\bibinfo{person}{Seongyeong Lee}, \bibinfo{person}{Hansoo
  Park}, \bibinfo{person}{Dong~Uk Kim}, \bibinfo{person}{Jihyeon Kim},
  \bibinfo{person}{Muhammadjon Boboev}, {and} \bibinfo{person}{Seungryul
  Baek}.} \bibinfo{year}{2023}\natexlab{}.
\newblock \showarticletitle{Image-free Domain Generalization via {CLIP} for 3D
  Hand Pose Estimation}. In \bibinfo{booktitle}{\emph{{IEEE/CVF} Winter
  Conference on Applications of Computer Vision, {WACV} 2023, Waikoloa, HI,
  USA, January 2-7, 2023}}. \bibinfo{publisher}{{IEEE}},
  \bibinfo{pages}{2933--2943}.
\newblock


\bibitem[\protect\citeauthoryear{Li, Yang, Cao, and Huang}{Li
  et~al\mbox{.}}{2017}]%
        {JTAE2017CIKM}
\bibfield{author}{\bibinfo{person}{Chao Li}, \bibinfo{person}{Yang Yang},
  \bibinfo{person}{Jiewei Cao}, {and} \bibinfo{person}{Zi Huang}.}
  \bibinfo{year}{2017}\natexlab{}.
\newblock \showarticletitle{Jointly Modeling Static Visual Appearance and
  Temporal Pattern for Unsupervised Video Hashing}. In
  \bibinfo{booktitle}{\emph{Proceedings of the 2017 {ACM} on Conference on
  Information and Knowledge Management, {CIKM} 2017, Singapore, November 06 -
  10, 2017}}. \bibinfo{publisher}{{ACM}}, \bibinfo{pages}{9--17}.
\newblock


\bibitem[\protect\citeauthoryear{Li, Zhou, Xiong, and Hoi}{Li
  et~al\mbox{.}}{2021b}]%
        {PCL2021ICLR}
\bibfield{author}{\bibinfo{person}{Junnan Li}, \bibinfo{person}{Pan Zhou},
  \bibinfo{person}{Caiming Xiong}, {and} \bibinfo{person}{Steven Hoi}.}
  \bibinfo{year}{2021}\natexlab{b}.
\newblock \showarticletitle{Prototypical Contrastive Learning of Unsupervised
  Representations}. In \bibinfo{booktitle}{\emph{International Conference on
  Learning Representations, {ICLR} 2021}}.
\newblock


\bibitem[\protect\citeauthoryear{Li, Zheng, and Sun}{Li et~al\mbox{.}}{2022}]%
        {ASSPH2022MM}
\bibfield{author}{\bibinfo{person}{Liang Li}, \bibinfo{person}{Baihua Zheng},
  {and} \bibinfo{person}{Weiwei Sun}.} \bibinfo{year}{2022}\natexlab{}.
\newblock \showarticletitle{Adaptive Structural Similarity Preserving for
  Unsupervised Cross Modal Hashing}. In \bibinfo{booktitle}{\emph{Proceedings
  of the 30th ACM International Conference on Multimedia}} (Lisboa, Portugal)
  \emph{(\bibinfo{series}{MM '22})}. \bibinfo{publisher}{Association for
  Computing Machinery}, \bibinfo{address}{New York, NY, USA},
  \bibinfo{pages}{3712–3721}.
\newblock
\showISBNx{9781450392037}


\bibitem[\protect\citeauthoryear{Li, Chen, Lu, Li, and Zhou}{Li
  et~al\mbox{.}}{2019}]%
        {NPH2019ICCV}
\bibfield{author}{\bibinfo{person}{Shuyan Li}, \bibinfo{person}{Zhixiang Chen},
  \bibinfo{person}{Jiwen Lu}, \bibinfo{person}{Xiu Li}, {and}
  \bibinfo{person}{Jie Zhou}.} \bibinfo{year}{2019}\natexlab{}.
\newblock \showarticletitle{Neighborhood Preserving Hashing for Scalable Video
  Retrieval}. In \bibinfo{booktitle}{\emph{2019 {IEEE/CVF} International
  Conference on Computer Vision, {ICCV} 2019, Seoul, Korea (South), October 27
  - November 2, 2019}}. \bibinfo{publisher}{{IEEE}},
  \bibinfo{pages}{8211--8220}.
\newblock


\bibitem[\protect\citeauthoryear{Li, Li, Lu, and Zhou}{Li
  et~al\mbox{.}}{2021a}]%
        {BTH2021CVPR}
\bibfield{author}{\bibinfo{person}{Shuyan Li}, \bibinfo{person}{Xiu Li},
  \bibinfo{person}{Jiwen Lu}, {and} \bibinfo{person}{Jie Zhou}.}
  \bibinfo{year}{2021}\natexlab{a}.
\newblock \showarticletitle{Self-Supervised Video Hashing via Bidirectional
  Transformers}. In \bibinfo{booktitle}{\emph{{IEEE} Conference on Computer
  Vision and Pattern Recognition, {CVPR} 2021, virtual, June 19-25, 2021}}.
  \bibinfo{publisher}{Computer Vision Foundation / {IEEE}},
  \bibinfo{pages}{13549--13558}.
\newblock


\bibitem[\protect\citeauthoryear{Li and van Gemert}{Li and van Gemert}{2021}]%
        {BitEntropy}
\bibfield{author}{\bibinfo{person}{Yunqiang Li} {and} \bibinfo{person}{Jan van
  Gemert}.} \bibinfo{year}{2021}\natexlab{}.
\newblock \showarticletitle{Deep Unsupervised Image Hashing by Maximizing Bit
  Entropy}. In \bibinfo{booktitle}{\emph{Thirty-Fifth {AAAI} Conference on
  Artificial Intelligence, {AAAI} 2021, Thirty-Third Conference on Innovative
  Applications of Artificial Intelligence, {IAAI} 2021, The Eleventh Symposium
  on Educational Advances in Artificial Intelligence, {EAAI} 2021, Virtual
  Event, February 2-9, 2021}}. \bibinfo{publisher}{{AAAI} Press},
  \bibinfo{pages}{2002--2010}.
\newblock


\bibitem[\protect\citeauthoryear{Liong, Lu, Wang, Moulin, and Zhou}{Liong
  et~al\mbox{.}}{2015}]%
        {DH2015CVPR}
\bibfield{author}{\bibinfo{person}{Venice~Erin Liong}, \bibinfo{person}{Jiwen
  Lu}, \bibinfo{person}{Gang Wang}, \bibinfo{person}{Pierre Moulin}, {and}
  \bibinfo{person}{Jie Zhou}.} \bibinfo{year}{2015}\natexlab{}.
\newblock \showarticletitle{Deep hashing for compact binary codes learning}. In
  \bibinfo{booktitle}{\emph{{IEEE} Conference on Computer Vision and Pattern
  Recognition, {CVPR} 2015, Boston, MA, USA, June 7-12, 2015}}.
  \bibinfo{publisher}{{IEEE} Computer Society}, \bibinfo{pages}{2475--2483}.
\newblock


\bibitem[\protect\citeauthoryear{Luo, Huang, Gong, Jin, and Liu}{Luo
  et~al\mbox{.}}{2023}]%
        {CVPR2023Video}
\bibfield{author}{\bibinfo{person}{Dezhao Luo}, \bibinfo{person}{Jiabo Huang},
  \bibinfo{person}{Shaogang Gong}, \bibinfo{person}{Hailin Jin}, {and}
  \bibinfo{person}{Yang Liu}.} \bibinfo{year}{2023}\natexlab{}.
\newblock \showarticletitle{Towards Generalisable Video Moment Retrieval:
  Visual-Dynamic Injection to Image-Text Pre-Training}.
\newblock \bibinfo{journal}{\emph{CoRR}}  \bibinfo{volume}{abs/2303.00040}
  (\bibinfo{year}{2023}).
\newblock


\bibitem[\protect\citeauthoryear{Luo, Wu, Ma, Chen, Deng, Huang, and Hua}{Luo
  et~al\mbox{.}}{2021}]%
        {DATE2021MM}
\bibfield{author}{\bibinfo{person}{Xiao Luo}, \bibinfo{person}{Daqing Wu},
  \bibinfo{person}{Zeyu Ma}, \bibinfo{person}{Chong Chen},
  \bibinfo{person}{Minghua Deng}, \bibinfo{person}{Jianqiang Huang}, {and}
  \bibinfo{person}{Xian{-}Sheng Hua}.} \bibinfo{year}{2021}\natexlab{}.
\newblock \showarticletitle{A Statistical Approach to Mining Semantic
  Similarity for Deep Unsupervised Hashing}. In \bibinfo{booktitle}{\emph{{MM}
  '21: {ACM} Multimedia Conference, Virtual Event, China, October 20 - 24,
  2021}}. \bibinfo{publisher}{{ACM}}, \bibinfo{pages}{4306--4314}.
\newblock


\bibitem[\protect\citeauthoryear{Misra, Zitnick, and Hebert}{Misra
  et~al\mbox{.}}{2016}]%
        {VideoOrder2016ECCV}
\bibfield{author}{\bibinfo{person}{Ishan Misra}, \bibinfo{person}{C.~Lawrence
  Zitnick}, {and} \bibinfo{person}{Martial Hebert}.}
  \bibinfo{year}{2016}\natexlab{}.
\newblock \showarticletitle{Shuffle and Learn: Unsupervised Learning Using
  Temporal Order Verification}. In \bibinfo{booktitle}{\emph{Computer Vision -
  {ECCV} 2016 - 14th European Conference, Amsterdam, The Netherlands, October
  11-14, 2016, Proceedings, Part {I}}} \emph{(\bibinfo{series}{Lecture Notes in
  Computer Science}, Vol.~\bibinfo{volume}{9905})},
  \bibfield{editor}{\bibinfo{person}{Bastian Leibe}, \bibinfo{person}{Jiri
  Matas}, \bibinfo{person}{Nicu Sebe}, {and} \bibinfo{person}{Max Welling}}
  (Eds.). \bibinfo{publisher}{Springer}, \bibinfo{pages}{527--544}.
\newblock


\bibitem[\protect\citeauthoryear{Pan, Song, Yang, Jiang, and Liu}{Pan
  et~al\mbox{.}}{2021}]%
        {VideoMoCo2021CVPR}
\bibfield{author}{\bibinfo{person}{Tian Pan}, \bibinfo{person}{Yibing Song},
  \bibinfo{person}{Tianyu Yang}, \bibinfo{person}{Wenhao Jiang}, {and}
  \bibinfo{person}{Wei Liu}.} \bibinfo{year}{2021}\natexlab{}.
\newblock \showarticletitle{VideoMoCo: Contrastive Video Representation
  Learning With Temporally Adversarial Examples}. In
  \bibinfo{booktitle}{\emph{{IEEE} Conference on Computer Vision and Pattern
  Recognition, {CVPR} 2021, virtual, June 19-25, 2021}}.
  \bibinfo{publisher}{Computer Vision Foundation / {IEEE}},
  \bibinfo{pages}{11205--11214}.
\newblock


\bibitem[\protect\citeauthoryear{Piergiovanni, Kuo, and Angelova}{Piergiovanni
  et~al\mbox{.}}{2022}]%
        {TubeViT2023arxiv}
\bibfield{author}{\bibinfo{person}{AJ Piergiovanni}, \bibinfo{person}{Weicheng
  Kuo}, {and} \bibinfo{person}{Anelia Angelova}.}
  \bibinfo{year}{2022}\natexlab{}.
\newblock \showarticletitle{Rethinking Video ViTs: Sparse Video Tubes for Joint
  Image and Video Learning}.
\newblock \bibinfo{journal}{\emph{CoRR}}  \bibinfo{volume}{abs/2212.03229}
  (\bibinfo{year}{2022}).
\newblock


\bibitem[\protect\citeauthoryear{Qian, Meng, Gong, Yang, Wang, Belongie, and
  Cui}{Qian et~al\mbox{.}}{2021}]%
        {CVRL2021CVPR}
\bibfield{author}{\bibinfo{person}{Rui Qian}, \bibinfo{person}{Tianjian Meng},
  \bibinfo{person}{Boqing Gong}, \bibinfo{person}{Ming{-}Hsuan Yang},
  \bibinfo{person}{Huisheng Wang}, \bibinfo{person}{Serge~J. Belongie}, {and}
  \bibinfo{person}{Yin Cui}.} \bibinfo{year}{2021}\natexlab{}.
\newblock \showarticletitle{Spatiotemporal Contrastive Video Representation
  Learning}. In \bibinfo{booktitle}{\emph{{IEEE} Conference on Computer Vision
  and Pattern Recognition, {CVPR} 2021, virtual, June 19-25, 2021}}.
  \bibinfo{publisher}{Computer Vision Foundation / {IEEE}},
  \bibinfo{pages}{6964--6974}.
\newblock


\bibitem[\protect\citeauthoryear{Qiu, Su, Ou, Yu, and Chen}{Qiu
  et~al\mbox{.}}{2021}]%
        {CIBHash2021IJCAI}
\bibfield{author}{\bibinfo{person}{Zexuan Qiu}, \bibinfo{person}{Qinliang Su},
  \bibinfo{person}{Zijing Ou}, \bibinfo{person}{Jianxing Yu}, {and}
  \bibinfo{person}{Changyou Chen}.} \bibinfo{year}{2021}\natexlab{}.
\newblock \showarticletitle{Unsupervised Hashing with Contrastive Information
  Bottleneck}. In \bibinfo{booktitle}{\emph{Proceedings of the Thirtieth
  International Joint Conference on Artificial Intelligence, {IJCAI} 2021,
  Virtual Event / Montreal, Canada, 19-27 August 2021}},
  \bibfield{editor}{\bibinfo{person}{Zhi{-}Hua Zhou}} (Ed.).
  \bibinfo{publisher}{ijcai.org}, \bibinfo{pages}{959--965}.
\newblock


\bibitem[\protect\citeauthoryear{Radford, Kim, Hallacy, Ramesh, Goh, Agarwal,
  Sastry, Askell, Mishkin, Clark, Krueger, and Sutskever}{Radford
  et~al\mbox{.}}{2021}]%
        {CLIP2021ICML}
\bibfield{author}{\bibinfo{person}{Alec Radford}, \bibinfo{person}{Jong~Wook
  Kim}, \bibinfo{person}{Chris Hallacy}, \bibinfo{person}{Aditya Ramesh},
  \bibinfo{person}{Gabriel Goh}, \bibinfo{person}{Sandhini Agarwal},
  \bibinfo{person}{Girish Sastry}, \bibinfo{person}{Amanda Askell},
  \bibinfo{person}{Pamela Mishkin}, \bibinfo{person}{Jack Clark},
  \bibinfo{person}{Gretchen Krueger}, {and} \bibinfo{person}{Ilya Sutskever}.}
  \bibinfo{year}{2021}\natexlab{}.
\newblock \showarticletitle{Learning Transferable Visual Models From Natural
  Language Supervision}. In \bibinfo{booktitle}{\emph{Proceedings of the 38th
  International Conference on Machine Learning, {ICML} 2021, 18-24 July 2021,
  Virtual Event}} \emph{(\bibinfo{series}{Proceedings of Machine Learning
  Research}, Vol.~\bibinfo{volume}{139})},
  \bibfield{editor}{\bibinfo{person}{Marina Meila} {and} \bibinfo{person}{Tong
  Zhang}} (Eds.). \bibinfo{publisher}{{PMLR}}, \bibinfo{pages}{8748--8763}.
\newblock


\bibitem[\protect\citeauthoryear{Ranasinghe, Naseer, Khan, Khan, and
  Ryoo}{Ranasinghe et~al\mbox{.}}{2022}]%
        {SVT2022CVPR}
\bibfield{author}{\bibinfo{person}{Kanchana Ranasinghe},
  \bibinfo{person}{Muzammal Naseer}, \bibinfo{person}{Salman Khan},
  \bibinfo{person}{Fahad~Shahbaz Khan}, {and} \bibinfo{person}{Michael~S.
  Ryoo}.} \bibinfo{year}{2022}\natexlab{}.
\newblock \showarticletitle{Self-supervised Video Transformer}. In
  \bibinfo{booktitle}{\emph{{IEEE/CVF} Conference on Computer Vision and
  Pattern Recognition, {CVPR} 2022, New Orleans, LA, USA, June 18-24, 2022}}.
  \bibinfo{publisher}{{IEEE}}, \bibinfo{pages}{2864--2874}.
\newblock


\bibitem[\protect\citeauthoryear{Simonyan and Zisserman}{Simonyan and
  Zisserman}{2015}]%
        {vgg2014ICLR}
\bibfield{author}{\bibinfo{person}{Karen Simonyan} {and}
  \bibinfo{person}{Andrew Zisserman}.} \bibinfo{year}{2015}\natexlab{}.
\newblock \showarticletitle{Very Deep Convolutional Networks for Large-Scale
  Image Recognition}.
\newblock  (\bibinfo{year}{2015}).
\newblock


\bibitem[\protect\citeauthoryear{Song, Yang, Huang, Shen, and Hong}{Song
  et~al\mbox{.}}{2011a}]%
        {MPH2011MM}
\bibfield{author}{\bibinfo{person}{Jingkuan Song}, \bibinfo{person}{Yi Yang},
  \bibinfo{person}{Zi Huang}, \bibinfo{person}{Heng~Tao Shen}, {and}
  \bibinfo{person}{Richang Hong}.} \bibinfo{year}{2011}\natexlab{a}.
\newblock \showarticletitle{Multiple Feature Hashing for Real-Time Large Scale
  near-Duplicate Video Retrieval}. In \bibinfo{booktitle}{\emph{Proceedings of
  the 19th ACM International Conference on Multimedia}} (Scottsdale, Arizona,
  USA) \emph{(\bibinfo{series}{MM '11})}. \bibinfo{publisher}{Association for
  Computing Machinery}, \bibinfo{address}{New York, NY, USA},
  \bibinfo{pages}{423–432}.
\newblock
\showISBNx{9781450306164}
\urldef\tempurl%
\url{https://doi.org/10.1145/2072298.2072354}
\showDOI{\tempurl}


\bibitem[\protect\citeauthoryear{Song, Yang, Huang, Shen, and Hong}{Song
  et~al\mbox{.}}{2011b}]%
        {MFH2011MM}
\bibfield{author}{\bibinfo{person}{Jingkuan Song}, \bibinfo{person}{Yi Yang},
  \bibinfo{person}{Zi Huang}, \bibinfo{person}{Heng~Tao Shen}, {and}
  \bibinfo{person}{Richang Hong}.} \bibinfo{year}{2011}\natexlab{b}.
\newblock \showarticletitle{Multiple feature hashing for real-time large scale
  near-duplicate video retrieval}. In \bibinfo{booktitle}{\emph{Proceedings of
  the 19th International Conference on Multimedia 2011, Scottsdale, AZ, USA,
  November 28 - December 1, 2011}}. \bibinfo{publisher}{{ACM}},
  \bibinfo{pages}{423--432}.
\newblock


\bibitem[\protect\citeauthoryear{Song, Zhang, Li, Gao, Wang, and Hong}{Song
  et~al\mbox{.}}{2018}]%
        {SSVH2018TIP}
\bibfield{author}{\bibinfo{person}{Jingkuan Song}, \bibinfo{person}{Hanwang
  Zhang}, \bibinfo{person}{Xiangpeng Li}, \bibinfo{person}{Lianli Gao},
  \bibinfo{person}{Meng Wang}, {and} \bibinfo{person}{Richang Hong}.}
  \bibinfo{year}{2018}\natexlab{}.
\newblock \showarticletitle{Self-Supervised Video Hashing With Hierarchical
  Binary Auto-Encoder}.
\newblock \bibinfo{journal}{\emph{{IEEE} Trans. Image Process.}}
  \bibinfo{volume}{27}, \bibinfo{number}{7} (\bibinfo{year}{2018}),
  \bibinfo{pages}{3210--3221}.
\newblock


\bibitem[\protect\citeauthoryear{Soomro, Zamir, and Shah}{Soomro
  et~al\mbox{.}}{2012}]%
        {UCF-1012012arxiv}
\bibfield{author}{\bibinfo{person}{Khurram Soomro},
  \bibinfo{person}{Amir~Roshan Zamir}, {and} \bibinfo{person}{Mubarak Shah}.}
  \bibinfo{year}{2012}\natexlab{}.
\newblock \showarticletitle{{UCF101:} {A} Dataset of 101 Human Actions Classes
  From Videos in The Wild}.
\newblock \bibinfo{journal}{\emph{CoRR}}  \bibinfo{volume}{abs/1212.0402}
  (\bibinfo{year}{2012}).
\newblock


\bibitem[\protect\citeauthoryear{Sun, Peng, Huang, and Ren}{Sun
  et~al\mbox{.}}{2022}]%
        {FSVCH2022MM}
\bibfield{author}{\bibinfo{person}{Yuan Sun}, \bibinfo{person}{Dezhong Peng},
  \bibinfo{person}{Haixiao Huang}, {and} \bibinfo{person}{Zhenwen Ren}.}
  \bibinfo{year}{2022}\natexlab{}.
\newblock \showarticletitle{Feature and Semantic Views Consensus Hashing for
  Image Set Classification}. In \bibinfo{booktitle}{\emph{Proceedings of the
  30th ACM International Conference on Multimedia}} (Lisboa, Portugal)
  \emph{(\bibinfo{series}{MM '22})}. \bibinfo{publisher}{Association for
  Computing Machinery}, \bibinfo{address}{New York, NY, USA},
  \bibinfo{pages}{2097–2105}.
\newblock
\showISBNx{9781450392037}


\bibitem[\protect\citeauthoryear{Tolstikhin, Houlsby, Kolesnikov, Beyer, Zhai,
  Unterthiner, Yung, Steiner, Keysers, Uszkoreit, Lucic, and
  Dosovitskiy}{Tolstikhin et~al\mbox{.}}{2021}]%
        {MLP-Mixer2021NIPS}
\bibfield{author}{\bibinfo{person}{Ilya~O. Tolstikhin}, \bibinfo{person}{Neil
  Houlsby}, \bibinfo{person}{Alexander Kolesnikov}, \bibinfo{person}{Lucas
  Beyer}, \bibinfo{person}{Xiaohua Zhai}, \bibinfo{person}{Thomas Unterthiner},
  \bibinfo{person}{Jessica Yung}, \bibinfo{person}{Andreas Steiner},
  \bibinfo{person}{Daniel Keysers}, \bibinfo{person}{Jakob Uszkoreit},
  \bibinfo{person}{Mario Lucic}, {and} \bibinfo{person}{Alexey Dosovitskiy}.}
  \bibinfo{year}{2021}\natexlab{}.
\newblock \showarticletitle{MLP-Mixer: An all-MLP Architecture for Vision}. In
  \bibinfo{booktitle}{\emph{Advances in Neural Information Processing Systems
  34: Annual Conference on Neural Information Processing Systems 2021, NeurIPS
  2021, December 6-14, 2021, virtual}}. \bibinfo{pages}{24261--24272}.
\newblock


\bibitem[\protect\citeauthoryear{van~der Maaten and Hinton}{van~der Maaten and
  Hinton}{2008}]%
        {tsne2008JMLR}
\bibfield{author}{\bibinfo{person}{Laurens van~der Maaten} {and}
  \bibinfo{person}{Geoffrey~E. Hinton}.} \bibinfo{year}{2008}\natexlab{}.
\newblock \showarticletitle{Visualizing Data using t-SNE}.
\newblock \bibinfo{journal}{\emph{Journal of Machine Learning Research}}
  \bibinfo{volume}{9} (\bibinfo{year}{2008}), \bibinfo{pages}{2579--2605}.
\newblock


\bibitem[\protect\citeauthoryear{Vaswani, Shazeer, Parmar, Uszkoreit, Jones,
  Gomez, Kaiser, and Polosukhin}{Vaswani et~al\mbox{.}}{2017}]%
        {Transformer2017NIPS}
\bibfield{author}{\bibinfo{person}{Ashish Vaswani}, \bibinfo{person}{Noam
  Shazeer}, \bibinfo{person}{Niki Parmar}, \bibinfo{person}{Jakob Uszkoreit},
  \bibinfo{person}{Llion Jones}, \bibinfo{person}{Aidan~N. Gomez},
  \bibinfo{person}{Lukasz Kaiser}, {and} \bibinfo{person}{Illia Polosukhin}.}
  \bibinfo{year}{2017}\natexlab{}.
\newblock \showarticletitle{Attention is All you Need}. In
  \bibinfo{booktitle}{\emph{Advances in Neural Information Processing Systems
  30: Annual Conference on Neural Information Processing Systems 2017, December
  4-9, 2017, Long Beach, CA, {USA}}},
  \bibfield{editor}{\bibinfo{person}{Isabelle Guyon}, \bibinfo{person}{Ulrike
  von Luxburg}, \bibinfo{person}{Samy Bengio}, \bibinfo{person}{Hanna~M.
  Wallach}, \bibinfo{person}{Rob Fergus}, \bibinfo{person}{S.~V.~N.
  Vishwanathan}, {and} \bibinfo{person}{Roman Garnett}} (Eds.).
  \bibinfo{pages}{5998--6008}.
\newblock


\bibitem[\protect\citeauthoryear{Wang, Wang, Chen, Zeng, and Xia}{Wang
  et~al\mbox{.}}{2023}]%
        {ConMH2023AAAI}
\bibfield{author}{\bibinfo{person}{Yuting Wang}, \bibinfo{person}{Jinpeng
  Wang}, \bibinfo{person}{Bin Chen}, \bibinfo{person}{Ziyun Zeng}, {and}
  \bibinfo{person}{Shu-Tao Xia}.} \bibinfo{year}{2023}\natexlab{}.
\newblock \showarticletitle{Contrastive Masked Autoencoders for Self-Supervised
  Video Hashing}. In \bibinfo{booktitle}{\emph{Proceedings of the AAAI
  Conference on Artificial Intelligence}}.
\newblock


\bibitem[\protect\citeauthoryear{Wang, Lu, Li, Tao, Guo, Gong, and Liu}{Wang
  et~al\mbox{.}}{2022}]%
        {CLIP_seg2022CVPR}
\bibfield{author}{\bibinfo{person}{Zhaoqing Wang}, \bibinfo{person}{Yu Lu},
  \bibinfo{person}{Qiang Li}, \bibinfo{person}{Xunqiang Tao},
  \bibinfo{person}{Yandong Guo}, \bibinfo{person}{Mingming Gong}, {and}
  \bibinfo{person}{Tongliang Liu}.} \bibinfo{year}{2022}\natexlab{}.
\newblock \showarticletitle{{CRIS:} CLIP-Driven Referring Image Segmentation}.
  In \bibinfo{booktitle}{\emph{{IEEE/CVF} Conference on Computer Vision and
  Pattern Recognition, {CVPR} 2022, New Orleans, LA, USA, June 18-24, 2022}}.
  \bibinfo{publisher}{{IEEE}}, \bibinfo{pages}{11676--11685}.
\newblock


\bibitem[\protect\citeauthoryear{Weiss, Torralba, and Fergus}{Weiss
  et~al\mbox{.}}{2008}]%
        {SpectralHashing2008NIPS}
\bibfield{author}{\bibinfo{person}{Yair Weiss}, \bibinfo{person}{Antonio
  Torralba}, {and} \bibinfo{person}{Rob Fergus}.}
  \bibinfo{year}{2008}\natexlab{}.
\newblock \showarticletitle{Spectral Hashing}. In
  \bibinfo{booktitle}{\emph{Proceedings of the 21st International Conference on
  Neural Information Processing Systems}} (Vancouver, British Columbia, Canada)
  \emph{(\bibinfo{series}{NIPS'08})}. \bibinfo{publisher}{Curran Associates
  Inc.}, \bibinfo{address}{Red Hook, NY, USA}, \bibinfo{pages}{1753–1760}.
\newblock
\showISBNx{9781605609492}


\bibitem[\protect\citeauthoryear{Yao, Zhang, Qiu, Pan, and Mei}{Yao
  et~al\mbox{.}}{2021}]%
        {SeCo2021AAAI}
\bibfield{author}{\bibinfo{person}{Ting Yao}, \bibinfo{person}{Yiheng Zhang},
  \bibinfo{person}{Zhaofan Qiu}, \bibinfo{person}{Yingwei Pan}, {and}
  \bibinfo{person}{Tao Mei}.} \bibinfo{year}{2021}\natexlab{}.
\newblock \showarticletitle{SeCo: Exploring Sequence Supervision for
  Unsupervised Representation Learning}. In
  \bibinfo{booktitle}{\emph{Thirty-Fifth {AAAI} Conference on Artificial
  Intelligence, {AAAI} 2021, Thirty-Third Conference on Innovative Applications
  of Artificial Intelligence, {IAAI} 2021, The Eleventh Symposium on
  Educational Advances in Artificial Intelligence, {EAAI} 2021, Virtual Event,
  February 2-9, 2021}}. \bibinfo{publisher}{{AAAI} Press},
  \bibinfo{pages}{10656--10664}.
\newblock


\bibitem[\protect\citeauthoryear{Yu, Zhan, Wu, Zhang, Lu, Cui, Xie, Hua, and
  Miao}{Yu et~al\mbox{.}}{2022}]%
        {CLIP_MAN2022MM}
\bibfield{author}{\bibinfo{person}{Yingchen Yu}, \bibinfo{person}{Fangneng
  Zhan}, \bibinfo{person}{Rongliang Wu}, \bibinfo{person}{Jiahui Zhang},
  \bibinfo{person}{Shijian Lu}, \bibinfo{person}{Miaomiao Cui},
  \bibinfo{person}{Xuansong Xie}, \bibinfo{person}{Xian{-}Sheng Hua}, {and}
  \bibinfo{person}{Chunyan Miao}.} \bibinfo{year}{2022}\natexlab{}.
\newblock \showarticletitle{Towards Counterfactual Image Manipulation via
  {CLIP}}. In \bibinfo{booktitle}{\emph{{MM} '22: The 30th {ACM} International
  Conference on Multimedia, Lisboa, Portugal, October 10 - 14, 2022}},
  \bibfield{editor}{\bibinfo{person}{Jo{\~{a}}o Magalh{\~{a}}es},
  \bibinfo{person}{Alberto~Del Bimbo}, \bibinfo{person}{Shin'ichi Satoh},
  \bibinfo{person}{Nicu Sebe}, \bibinfo{person}{Xavier Alameda{-}Pineda},
  \bibinfo{person}{Qin Jin}, \bibinfo{person}{Vincent Oria}, {and}
  \bibinfo{person}{Laura Toni}} (Eds.). \bibinfo{publisher}{{ACM}},
  \bibinfo{pages}{3637--3645}.
\newblock


\bibitem[\protect\citeauthoryear{Yuan, Wang, Zhang, Tay, Jie, Liu, and
  Feng}{Yuan et~al\mbox{.}}{2020}]%
        {CSQ2020CVPR}
\bibfield{author}{\bibinfo{person}{Li Yuan}, \bibinfo{person}{Tao Wang},
  \bibinfo{person}{Xiaopeng Zhang}, \bibinfo{person}{Francis E.~H. Tay},
  \bibinfo{person}{Zequn Jie}, \bibinfo{person}{Wei Liu}, {and}
  \bibinfo{person}{Jiashi Feng}.} \bibinfo{year}{2020}\natexlab{}.
\newblock \showarticletitle{Central Similarity Quantization for Efficient Image
  and Video Retrieval}. In \bibinfo{booktitle}{\emph{2020 {IEEE/CVF} Conference
  on Computer Vision and Pattern Recognition, {CVPR} 2020, Seattle, WA, USA,
  June 13-19, 2020}}. \bibinfo{publisher}{Computer Vision Foundation / {IEEE}},
  \bibinfo{pages}{3080--3089}.
\newblock


\bibitem[\protect\citeauthoryear{Yun, Kim, Han, Song, Ha, and Shin}{Yun
  et~al\mbox{.}}{2022}]%
        {TimeMatter2022ICML}
\bibfield{author}{\bibinfo{person}{Sukmin Yun}, \bibinfo{person}{Jaehyung Kim},
  \bibinfo{person}{Dongyoon Han}, \bibinfo{person}{Hwanjun Song},
  \bibinfo{person}{Jung{-}Woo Ha}, {and} \bibinfo{person}{Jinwoo Shin}.}
  \bibinfo{year}{2022}\natexlab{}.
\newblock \showarticletitle{Time Is MattEr: Temporal Self-supervision for Video
  Transformers}. In \bibinfo{booktitle}{\emph{International Conference on
  Machine Learning, {ICML} 2022, 17-23 July 2022, Baltimore, Maryland, {USA}}}
  \emph{(\bibinfo{series}{Proceedings of Machine Learning Research},
  Vol.~\bibinfo{volume}{162})}, \bibfield{editor}{\bibinfo{person}{Kamalika
  Chaudhuri}, \bibinfo{person}{Stefanie Jegelka}, \bibinfo{person}{Le~Song},
  \bibinfo{person}{Csaba Szepesv{\'{a}}ri}, \bibinfo{person}{Gang Niu}, {and}
  \bibinfo{person}{Sivan Sabato}} (Eds.). \bibinfo{publisher}{{PMLR}},
  \bibinfo{pages}{25804--25816}.
\newblock


\bibitem[\protect\citeauthoryear{Zaremba, Sutskever, and Vinyals}{Zaremba
  et~al\mbox{.}}{2014}]%
        {RNN2014arxiv}
\bibfield{author}{\bibinfo{person}{Wojciech Zaremba}, \bibinfo{person}{Ilya
  Sutskever}, {and} \bibinfo{person}{Oriol Vinyals}.}
  \bibinfo{year}{2014}\natexlab{}.
\newblock \showarticletitle{Recurrent Neural Network Regularization}.
\newblock \bibinfo{journal}{\emph{CoRR}}  \bibinfo{volume}{abs/1409.2329}
  (\bibinfo{year}{2014}).
\newblock
\showeprint[arXiv]{1409.2329}


\bibitem[\protect\citeauthoryear{Zhang, Wang, Hong, and Chua}{Zhang
  et~al\mbox{.}}{2016}]%
        {SSTH2016MM}
\bibfield{author}{\bibinfo{person}{Hanwang Zhang}, \bibinfo{person}{Meng Wang},
  \bibinfo{person}{Richang Hong}, {and} \bibinfo{person}{Tat{-}Seng Chua}.}
  \bibinfo{year}{2016}\natexlab{}.
\newblock \showarticletitle{Play and Rewind: Optimizing Binary Representations
  of Videos by Self-Supervised Temporal Hashing}. In
  \bibinfo{booktitle}{\emph{Proceedings of the 2016 {ACM} Conference on
  Multimedia Conference, {MM} 2016, Amsterdam, The Netherlands, October 15-19,
  2016}}, \bibfield{editor}{\bibinfo{person}{Alan Hanjalic},
  \bibinfo{person}{Cees Snoek}, \bibinfo{person}{Marcel Worring},
  \bibinfo{person}{Dick C.~A. Bulterman}, \bibinfo{person}{Benoit Huet},
  \bibinfo{person}{Aisling Kelliher}, \bibinfo{person}{Yiannis Kompatsiaris},
  {and} \bibinfo{person}{Jin Li}} (Eds.). \bibinfo{publisher}{{ACM}},
  \bibinfo{pages}{781--790}.
\newblock


\bibitem[\protect\citeauthoryear{Zhu, Long, Wang, and Cao}{Zhu
  et~al\mbox{.}}{2016}]%
        {DHN2016AAAI}
\bibfield{author}{\bibinfo{person}{Han Zhu}, \bibinfo{person}{Mingsheng Long},
  \bibinfo{person}{Jianmin Wang}, {and} \bibinfo{person}{Yue Cao}.}
  \bibinfo{year}{2016}\natexlab{}.
\newblock \showarticletitle{Deep Hashing Network for Efficient Similarity
  Retrieval}. In \bibinfo{booktitle}{\emph{Proceedings of the Thirtieth {AAAI}
  Conference on Artificial Intelligence, February 12-17, 2016, Phoenix,
  Arizona, {USA}}}, \bibfield{editor}{\bibinfo{person}{Dale Schuurmans} {and}
  \bibinfo{person}{Michael~P. Wellman}} (Eds.). \bibinfo{publisher}{{AAAI}
  Press}, \bibinfo{pages}{2415--2421}.
\newblock


\end{thebibliography}

\clearpage
\appendix
\section{Retrieval Performance with CLIP}
\begin{figure*}
\centering
\subfloat[ActivityNet 16 bits]{\includegraphics[width=0.33\textwidth]{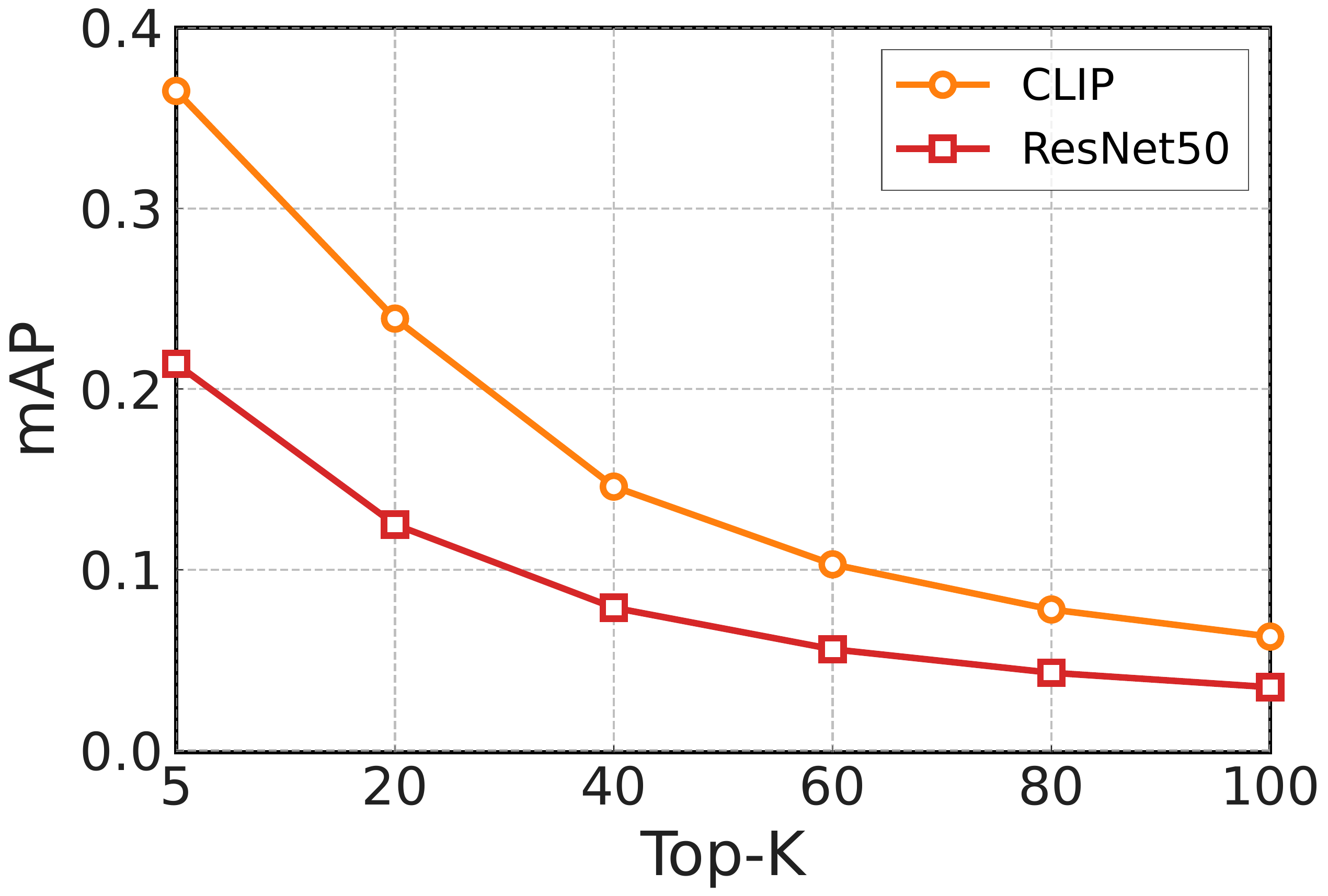}}
\hfill
\subfloat[ActivityNet 32 bits]{\includegraphics[width=0.33\textwidth]{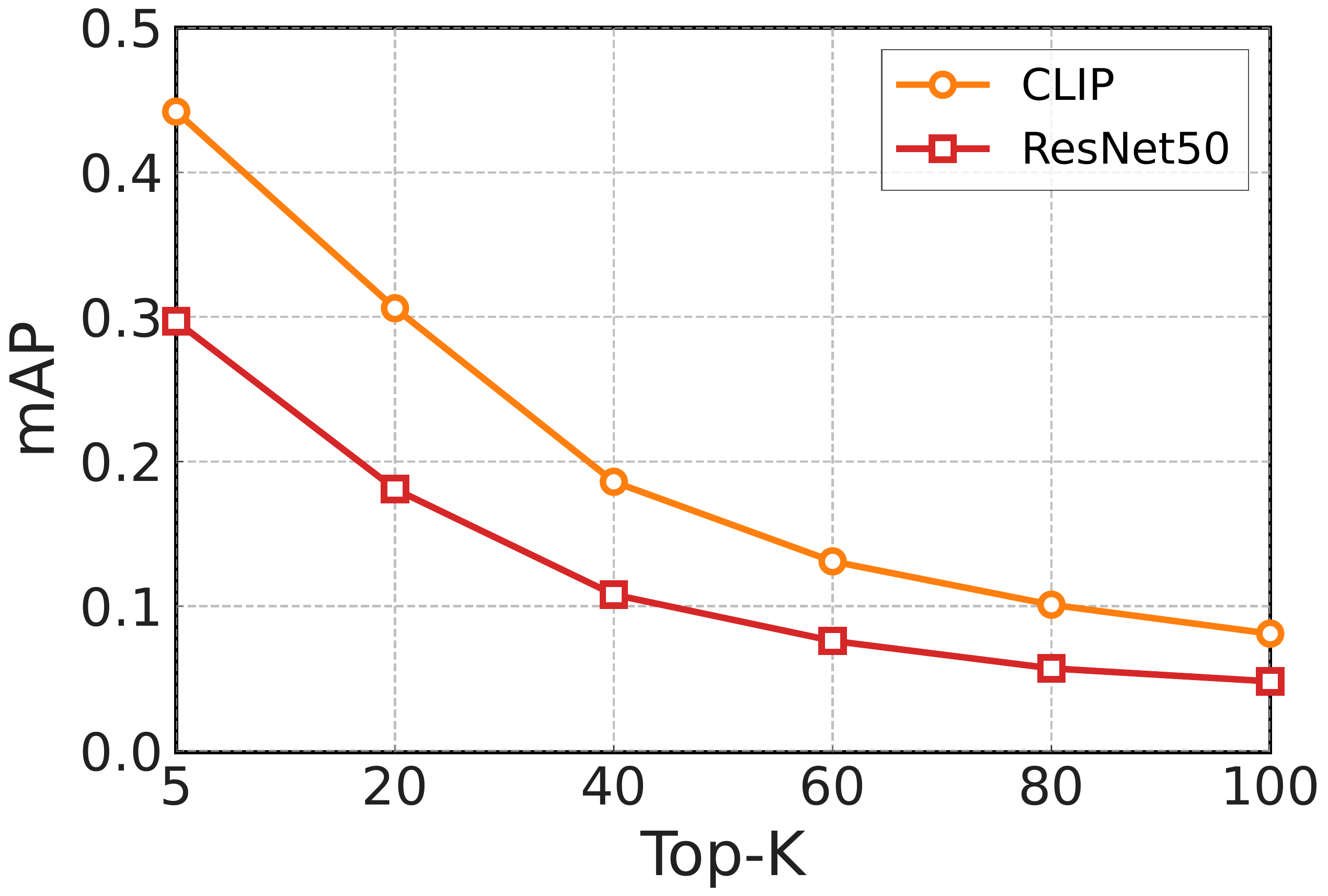}}
\hfill
\subfloat[ActivityNet 64 bits]{\includegraphics[width=0.33\textwidth]{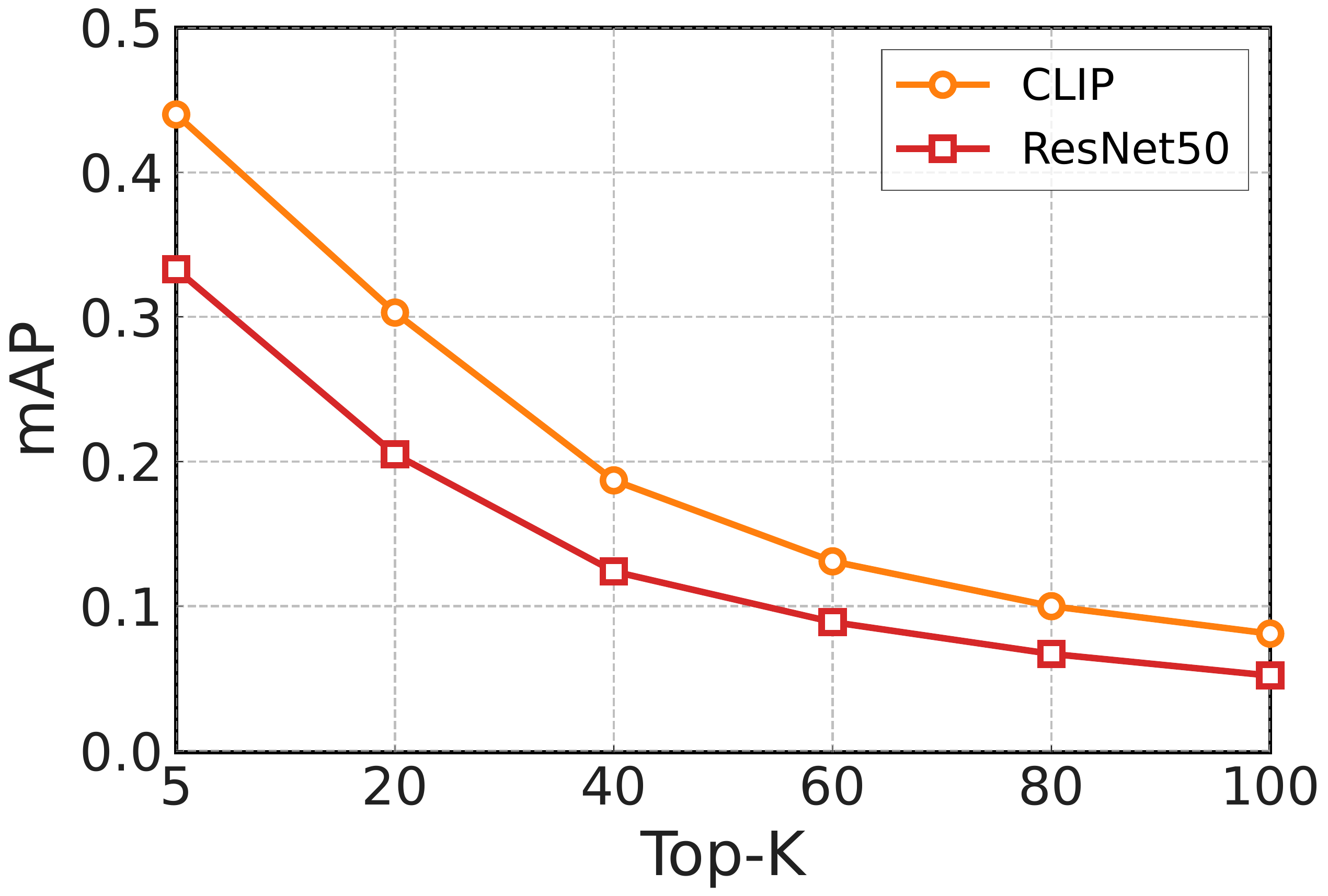}}
\\
\subfloat[UCF101 16 bits]{\includegraphics[width=0.33\textwidth]{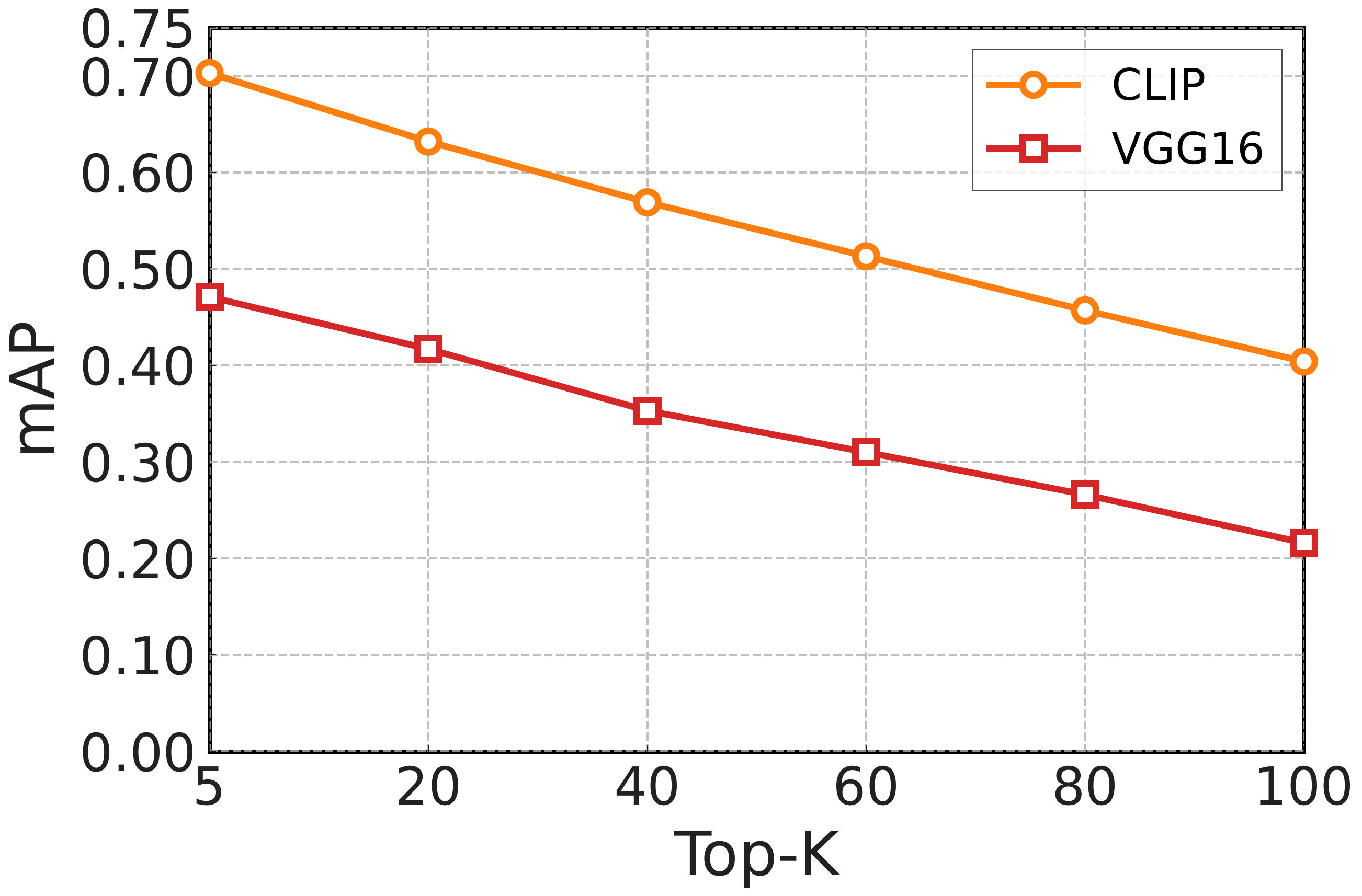}}
\hfill
\subfloat[UCF101 32 bits]{\includegraphics[width=0.33\textwidth]{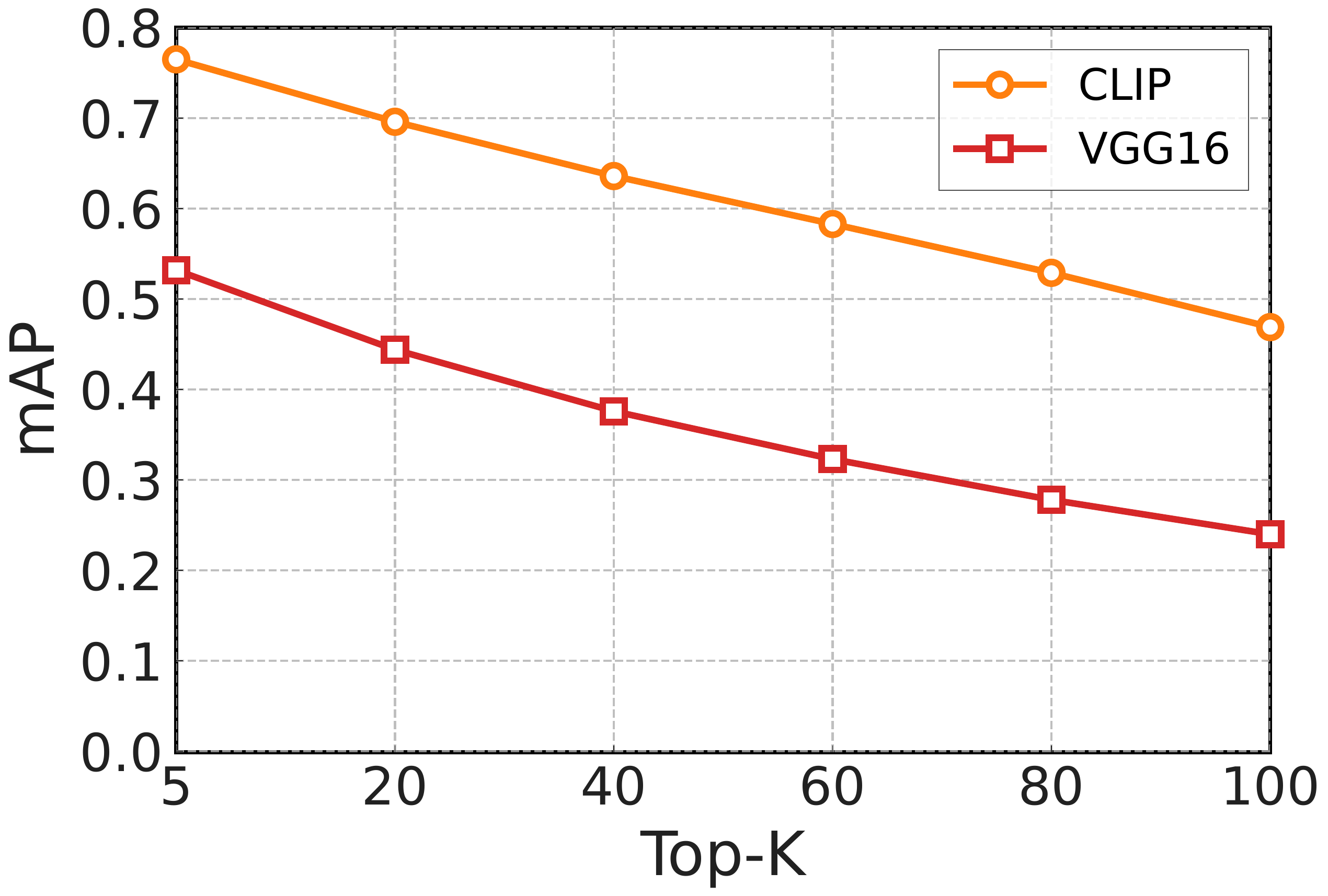}}
\hfill
\subfloat[UCF101 64 bits]{\includegraphics[width=0.33\textwidth]{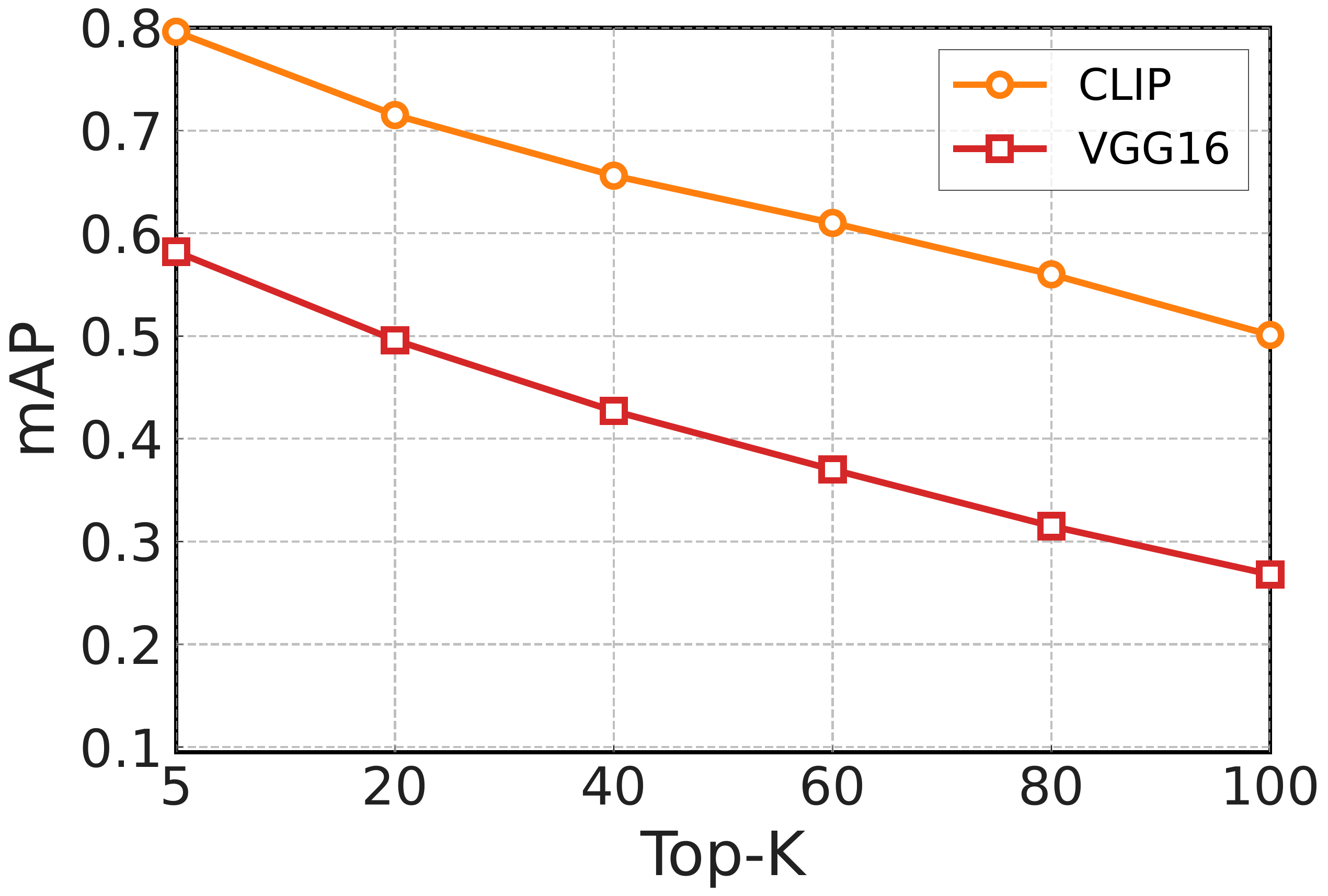}}
\caption{Comparison of retrieval performance using different 2D encoders in terms of mAP@K on two datasets.}
\label{fig:performance_clip}
\end{figure*}

Large-scale pre-trained models are known to be strong vision learners, and in particular, the CLIP model\cite{CLIP2021ICML} has demonstrated remarkable capabilities in various tasks\cite{CLIP_seg2022CVPR,CLIP_tran2022CVPR,CLIP_retrieval2022CVPR,CLIP_3D2023WACV,CLIP_MAN2022MM}. Compared to the CNN network used in our baseline model, the CLIP model has proven to excel in image feature extraction, owing to its pre-training on large-scale image-text datasets and experssive model design. To investigate how \ourmethod can benefit from a stronger feature extraction backbone, we use the CLIP visual branch (\ie, ViT-B/16~\cite{ViT2021ICLR}) to replace the CNN backbones. We evaluate the performance of \ourmethod in different datasets under such a stronger setting, and the results are shown in Figure~\ref{fig:performance_clip}.

We can observe that CLIP can provide significant performance improvements. For instance, on the ActivityNet and UCF101 datasets, \ourmethod+CLIP achieves mAP@20 values that are $47.8\%$ and $44.1\%$ higher than the baseline models that use VGG16~\cite{vgg2014ICLR} and ResNet50~\cite{ResNet2016CVPR}, respectively, when using 64-bit hash codes. Even with low-bit hash codes, CLIP brings remarkable enhancements, improving performance by up to $91.2\%$ and $51.5\%$ on the two datasets at 16 bits.

The CLIP model has great potential in video hashing tasks, as demonstrated by our initial results. However, our current approach only utilizes the pre-trained vision branch as our 2D frame feature extractor, without fully exploring its capabilities, particularly in zero-shot scenarios. To address this, we aim to investigate how to leverage visual prompt~\cite{Visual_prompt2022ECCV} learning for video hashing and explore ways to incorporate the text branch to further enhance the model's understanding of videos.

\section{Performance varying with number of sampled frames}

\begin{figure*}
    \centering
    \subfloat[ActivityNet 16 bits]{\includegraphics[width=0.33\textwidth]{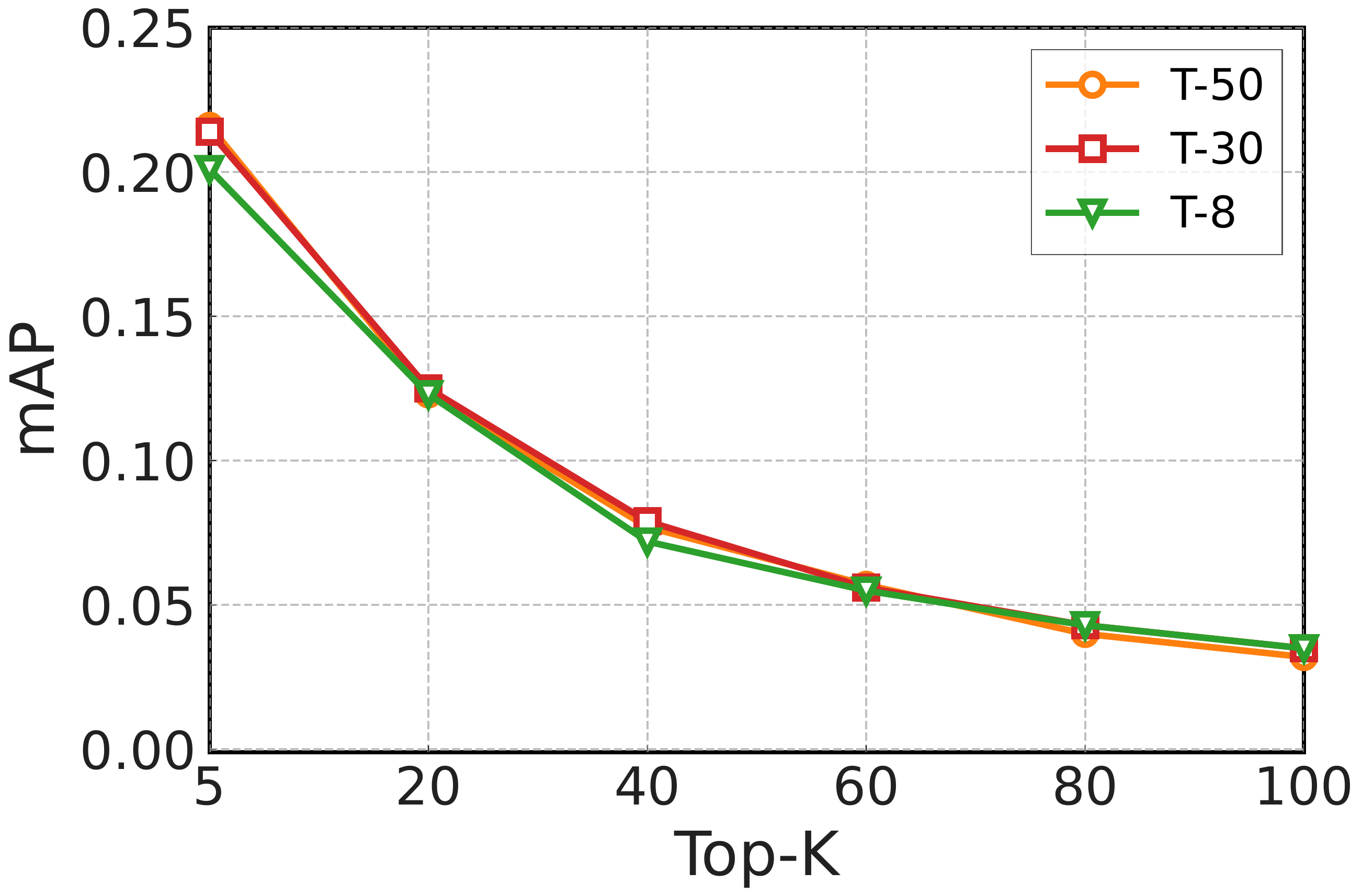}}
    \hfill
    \subfloat[ActivityNet 32 bits]{\includegraphics[width=0.33\textwidth]{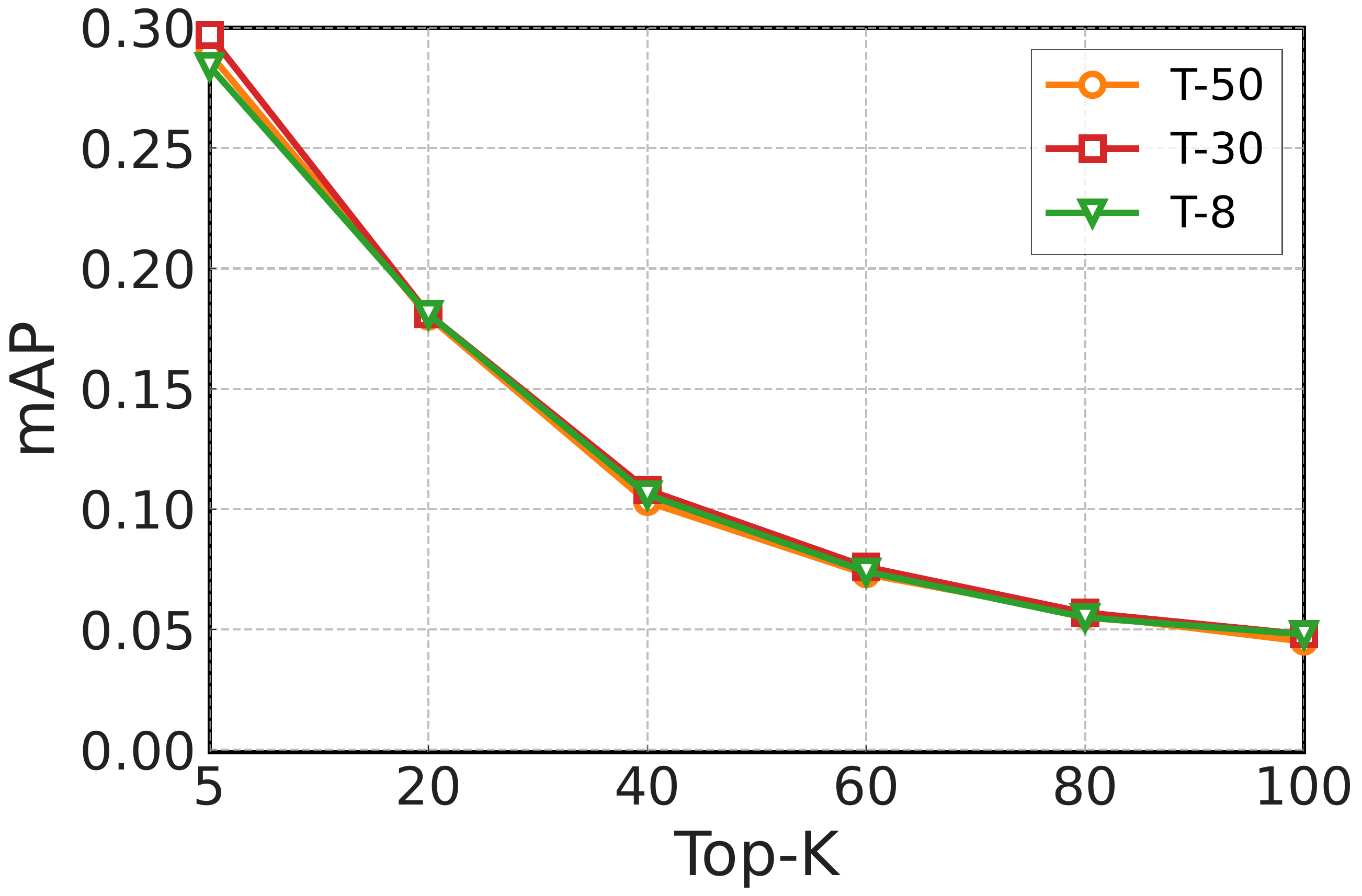}}
    \hfill
    \subfloat[ActivityNet 64 bits]{\includegraphics[width=0.33\textwidth]{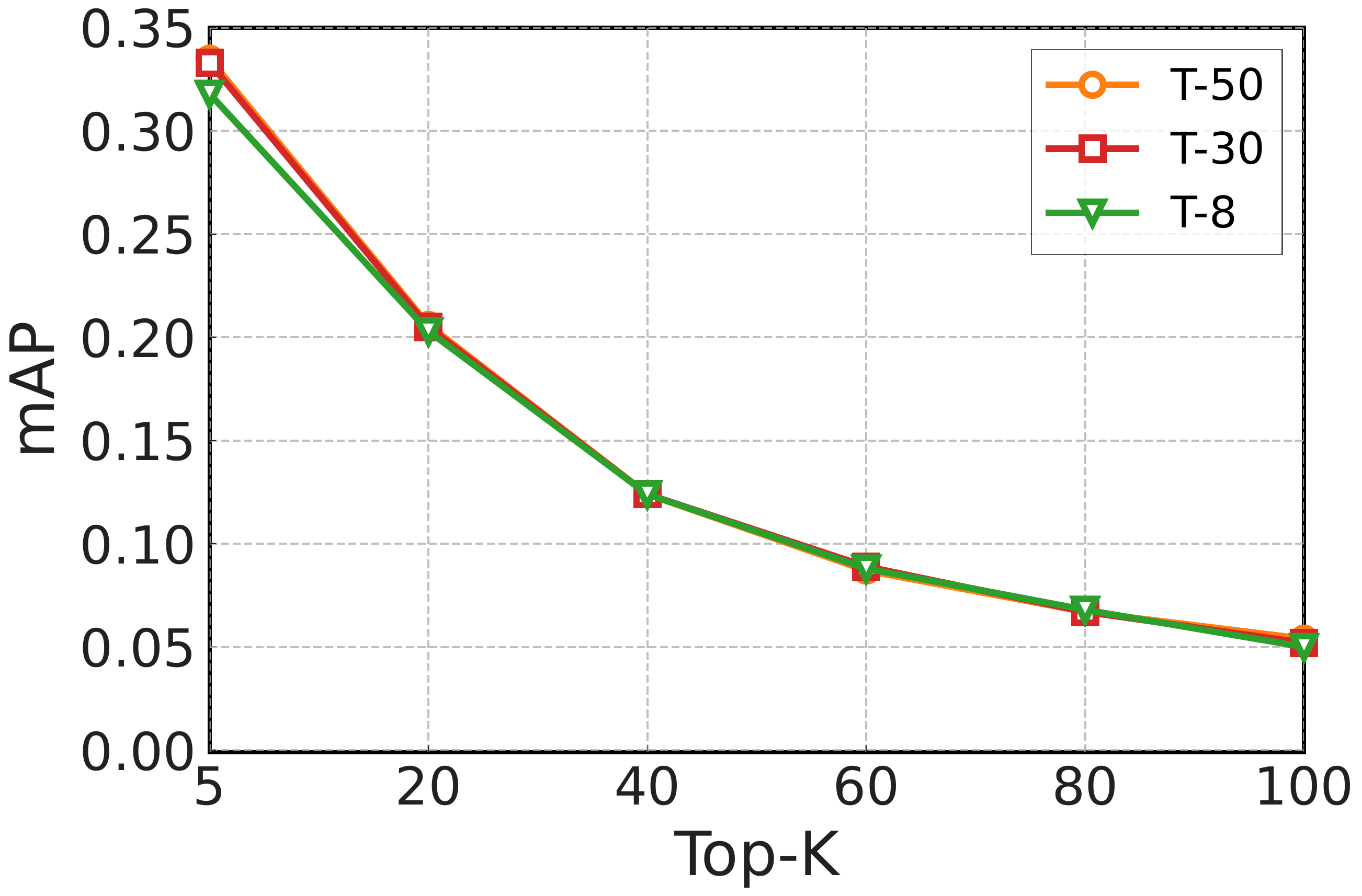}}
    \\
    \subfloat[UCF101 16 bits]{\includegraphics[width=0.33\textwidth]{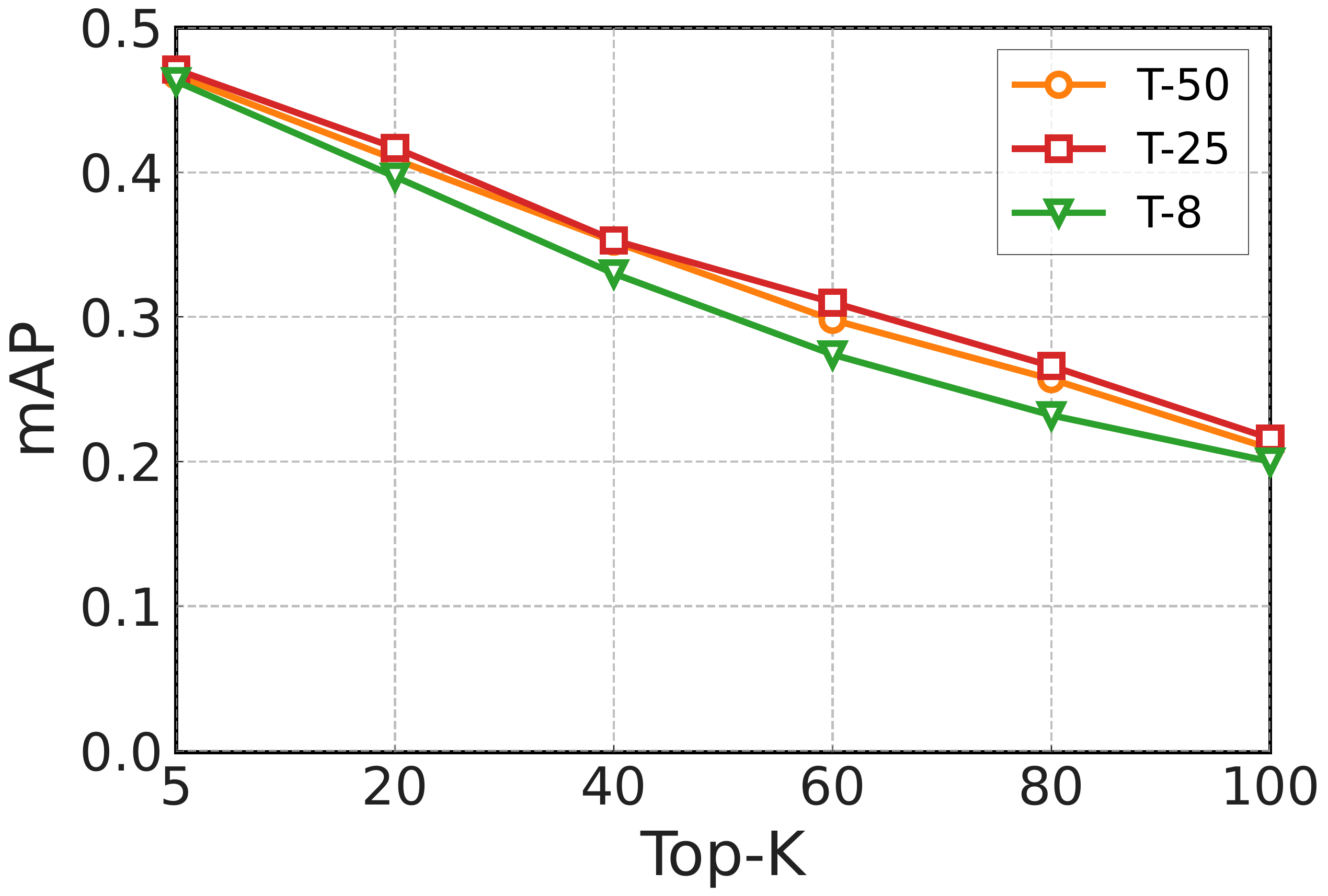}}
    \hfill
    \subfloat[UCF101 32 bits]{\includegraphics[width=0.33\textwidth]{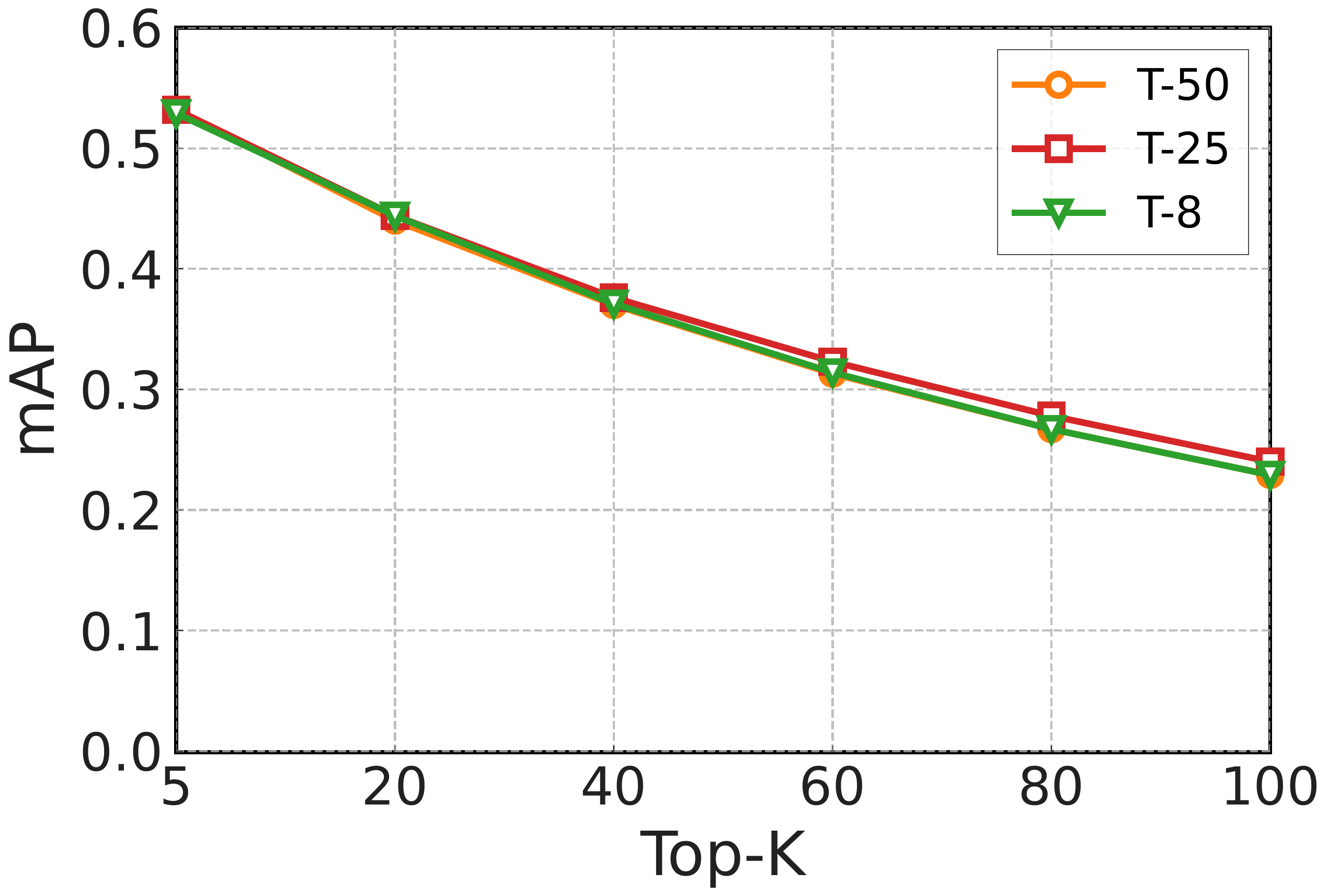}}
    \hfill
    \subfloat[UCF101 64 bits]{\includegraphics[width=0.33\textwidth]{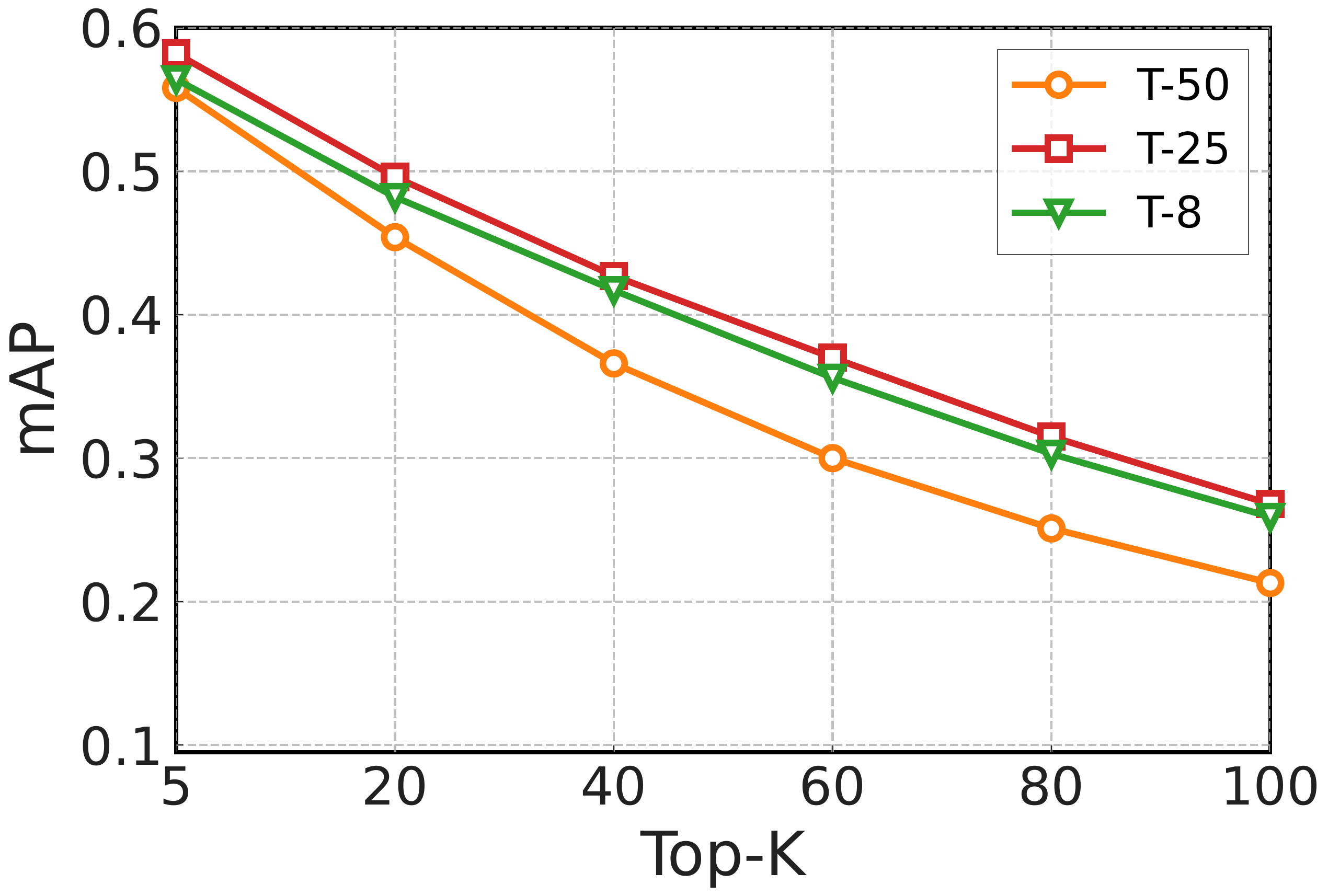}}
    \hfil
        \caption{Retrieval performance varying with different number of frames.}
\label{fig:sensitive_T}
\end{figure*}

For fair comparison, we set the number of sampled frames $T=25$ for UCF101 and $T=30$ for ActivityNet in our experiments. We investigate the influence of $T$ on retrieval performance, and results are illustrated in Figure~\ref{fig:sensitive_T}. Our findings indicate that sampling 50 frames (\ie, $T=50$) does not offer significant performance improvements, and may even result in degradation in certain datasets, such as UCF101 at 64 bits. The reason for this is that the videos in these datasets are relatively short and contain relatively simple content, and sampling too many frames introduces unnecessary redundancy. However, we observed a slight decrease in retrieval performance when the number of frames was reduced ($T=8$ v.s $T=25\text{ or }30$).

\section{Performance with different temporal encoders}

Note that the temporal encoder in our method is implementation-agnostic, allowing the integration of the latest models in this area. As a result, we also investigate the retrieval performance of \ourmethod when using a more distinct temporal encoder, such as MC-MLP~\cite{MCMSH2022MM}. In addition to the transformer encoder-based model mentioned above, we also evaluate other temporal encoders, namely LSTM~\cite{LSTM1997NatCommun} from SSTH~\cite{SSTH2016MM} and MC-MLP from MCMSH~\cite{MCMSH2022MM}. The results are presented in Table~\ref{tab:temporal_encoder}. These results demonstrate that the adoption of MC-MLP and transformer encoders can significantly enhance the performance of \ourmethod.

\section{Experimental Environment}
The experiments are conducted on a machine with Intel(R) Xeon(R) Gold 6226R CPU @ 2.90GHz, and a single NVIDIA RTX 3090 GPU with 24GB GPU memory. The operating system of the machine is Ubuntu 20.04.5 LTS. For software versions, we use Python 3.9.16, Pytorch 1.12.1, and CUDA 11.7.

\section{Limitations and Future Work}
Although our proposed \ourmethod achieves remarkable retrieval performance, there are still some limitations that need to be solved in our future work.
\begin{itemize}
    \item \textbf{Under-exploration of more advanced backbones.} Although we use VGG/ResNet as the spatial encoder and a transformer encoder to model temporal dynamics for fair comparisons, it falls short in dealing with real-world long videos. We plan to extend our method to more recent backbones such as TimeSformer~\cite{TimesFormer2021ICML} and TubeViT~\cite{TubeViT2023arxiv}.

    \item \textbf{High-cost in model deployment.} Although the time complexity of \ourmethod is in line with the baselines, its deployment in real-time applications remains challenging. We aim to explore computationally efficient models and compress the current model to better meet the needs of scenarios with high real-time requirements.
    
    \item \textbf{Expensive model updating.} The over-time changes in the distribution of real-world video data will degrade the retrieval performance. To address these issues, we aim to explore more effective updating strategies that can reduce the cost of re-training the model and refreshing hashing indexes.
\end{itemize}

\begin{table}[htb!]
    \centering
    \caption{Comparison of mAP@K using different temporal encoders on the ActivityNet dataset with 64-bit hash codes.}
    \scalebox{0.9}{
    \begin{tabular}{ccccccc}
    \toprule
         temporal encoder&K=5&K=20&K=40&K=60&K=80&K=100  \\
         \midrule
         \ourmethod+LSTM& 0.294&0.169&0.109&0.079&0.060&0.047 \\
         \ourmethod+Transformer& 0.333&0.205&0.124&0.089&0.067&0.052 \\
         \ourmethod+MC-MLP& 0.339&0.209&0.128&0.092&0.069&0.051\\
    \bottomrule
    \end{tabular}   
    }
    \label{tab:temporal_encoder}
\end{table}

\end{document}